\documentclass[10pt,journal,compsoc]{IEEEtran}
\ifCLASSOPTIONcompsoc
  \usepackage[nocompress]{cite}
\else
  \usepackage{cite}
\fi
\ifCLASSINFOpdf
\else
\fi
\usepackage{times}
\usepackage{url}
\usepackage{graphics}
\usepackage{color}
\usepackage[pdftex]{hyperref}
\usepackage{subfigure}
\usepackage{graphicx}
\usepackage{array}
\usepackage{placeins}
\usepackage{multirow}
\usepackage{epstopdf}

\usepackage{amssymb}
\usepackage{algorithm, algpseudocode}
\usepackage{caption}
\usepackage{flushend}
\usepackage{amsmath}
\usepackage{rotating}
\usepackage{indentfirst}

\begin{document}
%
\title{Semi-automated Signal Surveying Using Smartphones and Floorplans}
%
%
%
%

\author{Chao Gao, Robert Harle
\IEEEcompsocitemizethanks{\IEEEcompsocthanksitem C. Gao and R. Harle are with the Computer Laboratory, University of Cambridge, UK, CB3 0FD.\protect\\
E-mail: cg500@cam.ac.uk and rkh23@cam.ac.uk}
}

\IEEEtitleabstractindextext{%
  \begin{abstract}
Location fingerprinting locates devices based on pattern matching signal observations to a pre-defined signal map. This
paper introduces a technique to enable fast signal map creation given a
dedicated surveyor with a smartphone and floorplan. Our technique
(PFSurvey) uses accelerometer, gyroscope and magnetometer data to
estimate the surveyor's trajectory post-hoc using Simultaneous
Localisation and Mapping and particle filtering to incorporate
a building floorplan. We demonstrate conventional methods can fail to
recover the survey path robustly and determine the room unambiguously.
To counter this we use a novel loop closure detection method based on
magnetic field signals and propose to incorporate the magnetic loop
closures and straight-line constraints into the filtering process to
ensure robust trajectory recovery. We show this allows room
ambiguities to be resolved.

An entire building can be surveyed by the proposed system in minutes
rather than days. We evaluate in a large office space and compare to
state-of-the-art approaches. We achieve trajectories within 1.1~m of
the ground truth 90\% of the time. Output signal maps well approximate
those built from conventional, laborious manual survey. We also
demonstrate that the signal maps built by PFSurvey provide similar or
even better positioning performance than the manual signal maps.
  
\end{abstract}

}

\maketitle

\IEEEdisplaynontitleabstractindextext

%
\IEEEpeerreviewmaketitle

\IEEEraisesectionheading{\section{Introduction}\label{sec:introduction}}

%
%
%
%
\IEEEPARstart{T}{he} advent of personal computing devices has brought
with it an increased demand for location-based services. These
services have been powered primarily by Global Navigation Satellite
Systems such as GPS due to their wide availability and high
accuracy. However, these systems can only provide location
outdoors. Within buildings---where people spend the majority of their
time---ubiquitous location remains elusive. Many research prototypes
have been developed but wide deployments are typically hindered by the
need for custom infrastructure.

\begin{figure}
	\centering
	\begin{tabular}{c}
		\subfigure[An example grid-based manual
		survey. \label{fig:example-manual-survey}]{\includegraphics[height=7cm,angle=90]{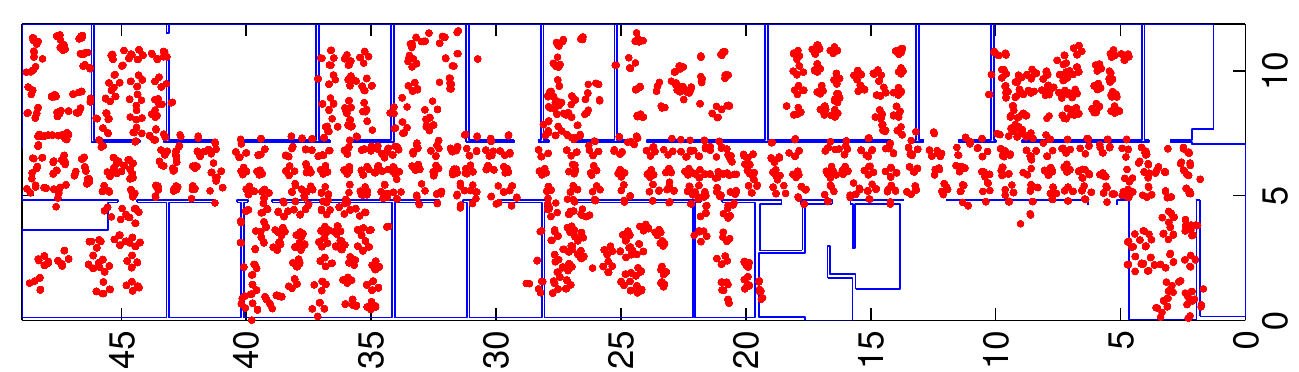}}
		\\ \subfigure[An example path
		survey \label{fig:example-path-survey}]{\includegraphics[height=7cm,angle=90]{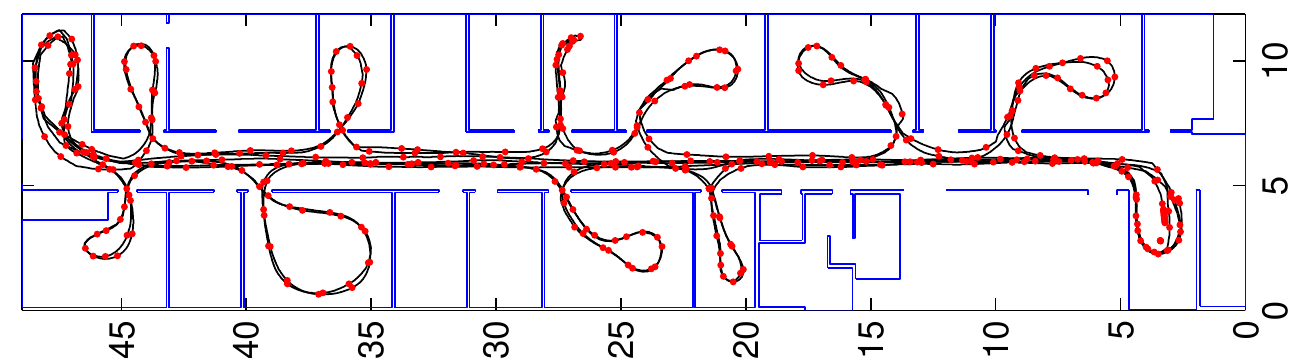}}
	\end{tabular}
	\caption{Manual survey vs path survey}
	\label{fig:example-manual-path-survey}  
\end{figure}

The most successful systems in terms of adoption are based on
repurposing existing infrastructure, often exploiting pervasive WiFi
signals. Due to the complex propagation of radio indoors, the
empirical \emph{fingerprinting} location technique has dominated. This
involves two stages: an offline \emph{signal survey} at regular points
throughout the space of interest (e.g. Figure
\ref{fig:example-manual-survey}) to collect signal strength samples
labelled with location information; then an online positioning
phase where the labelled samples are used to create a radio map, and
radio measurements observed at the mobile device are pattern-matched
to this map. Although this can achieve good positioning results the
offline survey is typically laborious. Coupled with the need for
regular resurveys due to environment changes, this has limited the
scalability (and general availability) of fingerprinting. To
illustrate this point, mapping 450,000m$^2$ of the COEX complex in
Seoul, Korea is reported to have required 15 surveyors and taken two
weeks~\cite{han2014building}. 

Some researchers have advocated crowdsourcing the signal survey
process. In some sense, the survey is then `free': a survey point is
created by any device that knows its location (or can be subsequently
located) and is willing to report its current observations. This
is a highly scalable concept, but there are a number of practical
issues that have prevented its wide use to date:

\begin{enumerate}
  
  \item The crowdsourced data collection is battery-intensive. It
    usually requires all of the inertial sensors to be on alongside
    continuous WiFi scanning and other sensing. Typical users are
    reluctant to run sensors they are not directly using since they
    reduce battery life; heat up the phone; and interfere with normal
    usage (e.g. repeated WiFi scanning adversely impacts the WiFi
    performance).

  \item Map quality can vary dramatically according to the volume and
    quality of the crowdsourced data in a specific space. This results
    in inconsistent location accuracy, which is difficult for
    location-aware applications to handle. Furthermore, the places that are less interesting to most people (so the volume and quality of the crowdsourced data are low) could be important for others. A good example is the areas such as toilets for disabled people. This inconsistency harms the user experience of these people who needs special care.

  \item Security and privacy are potentially at risk. Malicious users
      could contrive to adapt the map to their advantage, and privacy
      is at risk unless the data are carefully anonymised (which may
      be difficult given that devices must be individually profiled
      for best results).
  
  \item Device heterogeneity makes it difficult to combine crowdsourced
  measurements. A number of previous works have reported significant
  differences in the signal strength measurements made on one phone
  model to those made on another in the same context~\cite{kjaergaard2011indoor}.

  \item Even the same device can record a different fingerprint at the
  same location according to its context. For example, being carried
  in-hand vs in-pocket vs in-bag vs within a dense crowd.
  
  \end{enumerate}

This paper investigates a different approach for the \emph{offline
  signal survey}. We retain the notion of a \emph{dedicated surveyor}
(i.e. a user who is willing to follow specific guidelines to
explicitly collect the fingerprint data needed), and focus instead on
how to make their job much easier. The core idea is to collect survey
data along trajectories rather than at discrete grid points. We
advocate that if the trajectory can be accurately embedded in a
building's frame of reference, any signals measured along the path are
survey points (i.e. a triple of \{time, signal strength, location\})
that contribute to a larger \emph{path survey} as illustrated in
Figure \ref{fig:example-path-survey}. The black line represents the
surveyor's trajectory; the red points are the measurement points. Then
the signal survey is done simply by finishing a trajectory (or
trajectories) that explore a given space comprehensively.

By removing the laborious manual survey and advocating the use of a
dedicated surveyor, we enable efficient signal surveying which has the
following advantages over crowdsourcing:

\begin{enumerate}
	
	\item Power consumption is no longer a major concern because
          we do not require the user to contribute to the signal
          survey. A survey walk finishes much quicker than a
          manual survey (minutes vs. hours). 
	
	\item A dedicated surveyor can guarantee the space has
          been surveyed comprehensively to achieve consistent positioning
          accuracy.
	
	\item Security and privacy are protected since no
          user data transmission is needed in the signal survey.
	
	\item The problem of device heterogeneity could be alleviated
          because we can at least guarantee that only a single device
          is used to conduct one signal survey. To generalise the
          signal samples collected by one device to different kinds of
          device, some techniques
          like~\cite{journals/tmc/HossainJSV13} could be used.
	
	\item For the problem of different phone positions, the
          dedicated surveyor can keep the phone position constant.
          This has the additional benefit that the grip can be chosen
          to make the step detection problem simpler.
          ~\cite{brajdic2013walk}.
\end{enumerate}

In terms of cost, our proposed survey method requires a dedicated
surveyor to do the survey task. But our method enables the survey to
be done on a regular basis in extremely low cost because we are able
to simplify the survey to a simple walk that passes within a few
metres of anywhere that positioning is required. Surveying a typical
office space takes only minutes and may even be carried out by
security personnel or cleaning staff (both of which are expected to
visit all of the building regularly).

However, this approach is challenging in practice: high quality
trajectory estimation is difficult and accurate trajectory embedding
is rarely possible. To solve this, we present \emph{PFSurvey} and make
the following contributions in this paper:
\begin{itemize}
	\item We show how traditional particle filtering techniques
          can fail to recover the survey path robustly.
	
	\item We propose the \emph{PFSurvey} system that uses 
          smartphone accelerometer, gyroscope and magnetometer data to
          estimate a dedicated surveyor's trajectory post-hoc using
          Simultaneous Localisation and Mapping (SLAM) techniques and
          particle filtering to incorporate a building floorplan.

	\item We propose a novel loop closure detection method based
          on magnetic field signals and show how magnetic loop
          closures and straight-line constraints can be incorporated
          into the filtering process to ensure robust trajectory
          recovery.
\end{itemize}
  
Our focus is on the offline signal survey, the purpose of which is to
collect signal strength samples and label them with location
information. Our goal is to reduce the efforts of a signal survey by
allowing them to simply walk around the environment (a \emph{path
  survey}). The subsequent online positioning phase, which uses the
data to estimate position, is out of the scope of this work, although
we use a straw-man implementation to illustrate our results. A more
rigorous evaluation can be found in~\cite{gao2016easing}.

\subsection{A Dedicated Surveyor\label{sec:dedicated-surveyor}}
We emphasise that we assume a dedicated surveyor in this work. This is
motivated by our belief that Pedestrian Dead reckoning (PDR)
algorithms are not sufficiently mature to robustly estimate the
trajectories of arbitrary multi-purpose devices. Assuming a dedicated surveyor affords us a number of advantages:

\begin{itemize}
	\item the surveyor will carry the smartphone consistently; We require the surveyor holds the phone flat in front of the human body as if navigation. This increases the success rate of PDR algorithms, which is the very first step of the proposed system.
	\item the surveyor will
	cover the area comprehensively following best-practice guidelines. This ensures the quality of the signal maps being built.
	\item a start position can be manually specified. This is not
	mandatory, but significantly reduces the computational complexity.
\end{itemize}

We have previously described some survey guidelines to give good
results~\cite{gao2016easing}, summarised here for convenience:

\begin{itemize}
\item The survey path should visit each room and pass within 1--2~m of
  wherever positioning is required.

  \item The surveyor should repeat some parts of the path to increase
    the signal sampling density (particularly important for WiFi).

    \item Each path should be traversed in both directions wherever
      possible.

\end{itemize}

 Many public buildings have security or building management
personnel who regularly walk through them. These people offer an ideal
opportunity to perform regular surveys. Note that because our approach
does not require a `live' location estimate, we can post-process the
data using greater computational resources.

By quantitative evaluation, we demonstrate that our system achieves
good accuracy and efficiency in trajectory recovery. We also
demonstrate that the signal maps built by our system well approximate
the signal maps built by a more costly manual survey and provide similar
or even better positioning performance.

\section{Related Work}

{\bf Indoor location.} Indoor location is an active research
field---comprehensive surveys of the many prototypes and techniques
can be found in~\cite{liu2007survey, Al2011,Koyuncu,Gu2009}. In this
work we seek to produce radio maps from a (possibly jointly) estimated
trajectory.

{\bf Radio maps.} The idea of a radio map for location stems from the
RADAR system~\cite{bahl2000radar}, which was extended by the Horus
system~\cite{YoussefHorus} and many others \cite{Honkavirta09}. Most
of these systems assume an offline manual survey where a surveyor
measures the signal at each point on a fine grid covering the indoor
area. This is a difficult task, partly due to the lengthy and
laborious nature of the work and partly because it is difficult to
ensure the measuring device is precisely at a given grid point.

{\bf Inertial sensors and Pedestrian Dead Reckoning}.  Estimating the
trajectory of a pedestrian without dedicated infrastructure has been
achieved in a number of ways. The starting point is some form of step
detection based on the inertial sensors, where a step is characterised
with a length and (possibly change of) heading. This can be achieved
very accurately using foot-mounted sensors~\cite{woodman2008}, or more
coarsely with unconstrained
devices~\cite{brajdic2013walk}. Accumulating the step vectors leads to
a raw PDR trajectory that is prone to drift.

This drift can be constrained using extrinsic information,
particularly floorplans. A particle filter is typically used to ensure
the trajectory remains consistent with the floorplan. Systems such as
\cite{woodman2008,Krach2008a} provide \emph{instantaneous} location
estimates---i.e. at time $t=T$ they sample the probability
distribution for the current position based on all measurements and
state for $0<t<T$. The position estimate is usually taken as the
weighted mean of the samples.

{\bf Post-hoc trajectory estimation}. In this work we do not require
instantaneous location estimation as the data arrive but rather a best
estimate of the trajectory post-hoc. Thus at time $t=T$ we want to
estimate the probability distribution given the events at all epochs,
even the `future' ones ($t>T$). Essentially, knowing where we are now
may allow a better estimate of where we were, particularly if we were
uncertain at the time (e.g. a multi-modal distribution). Particle
\emph{smoother} algorithms are commonly used to provide the best
post-hoc estimates. For example Fixed Lag Smoothing
(FL)~\cite{Klepal2008}, Forward Filter Backward Smoothing
(FFBS)~\cite{doucet2000sequential} and Forward Filter Backward
Simulation (FFBSi)~\cite{Godsill2004aa}. These involve retaining the
particle distributions at each epoch (which can be costly in terms of
space) and reprocessing at various stages (which can be
computationally costly)

A simpler but less formally correct approach was described anonymously
in \cite{Eliazar1} (DP-SLAM)---we refer to it as particle pruning.  An
ancestor tree for each particle is retained as before. However, when a
particle is not resampled we walk up its ancestor branch, removing any
parent particles in previous epochs that have no other child
(`pruning'). At the end of the filter, the position at each epoch is
computed as the weighted mean of the remaining particles for that
epoch. The pruning approach is less resource-intensive but needs a
large number of particles to ensure older epochs do not suffer
particle depletion.

{\bf SLAM.} An alternative post-hoc approach is to use external
spatially-variant signals (possibly even those we wish to map) to
enable Simultaneous Localisation and Mapping (SLAM). The core idea is
to search the external observations for evidence of loops in the
trajectory (when the external observations return to values recorded
earlier in the trajectory). These form \emph{loop closures} that are
used to constrain the post-hoc trajectory estimate. The SLAM
algorithms are either graph-based (e.g. GraphSLAM~\cite{GraphSLAM}) or
use particle filters (e.g DP-SLAM~\cite{Eliazar1} and
FastSLAM~\cite{montemerlo2002fastslam}). They have been used when a
floorplan was unavailable, giving \emph{unanchored}
trajectories. Section \ref{sec:path-survey} shows that these
unanchored trajectories actually have limited use since they cannot
easily be mapped back to meaningful features (e.g. specific rooms). So
this SLAM method is optimal only when a floor plan is not available.

{\bf Crowdsourcing.} Crowdsourced radio map systems often make use of
a variety of the techniques just listed. Zee~\cite{rai2012zee} fuses
user traces to a floor plan using a particle filter and WiFi to help
locate later traces. UnLoc~\cite{wang2012no} locates traces to a
floorplan via landmarks. WILL~\cite{wu2013will} clusters
fingerprints collected along traces into ``virtual rooms'', and then
maps virtual rooms to physical rooms on the floor
plan automatically by the proposed \emph{subsection mapping method} (SSMM). LiFS~\cite{wu2015smartphones} maps a high
dimensional fingerprint space formed by RSS fingerprints and traces to
physical space (floor plan). Chai et al.~\cite{chai2007reducing}
collect fingerprints labelled with location information on sparse
sample points to generate an initial map, and then locates traces to
the floor plan using this map and then iteratively improves the
map. EZ~\cite{chintalapudi2010indoor} and LARM~\cite{pan2012tracking}
try to model the signal strength distribution using learning-based
methods, but they still need some fingerprints labelled with
location information. UCMA~\cite{jung2016unsupervised} is also a
learning-based crowdsourcing system but requires no labelled
fingerprints and no inertial data. It proposes a method that
integrates a memetic algorithm and a segmental k-means algorithm in a
hybrid global-local optimisation scheme to locate unlabelled
fingerprints to the floor plan. WiFi-SLAM~\cite{Ferris1} models the
signal strength distribution and simultaneously recovers the
trajectories. It is based on the assumption of dense AP deployment and
still needs some labelled data to train the initial values for some
model parameters. WiFi GraphSLAM~\cite{GraphSLAM} also recovers and
corrects user traces using WiFi signals but it relaxes the assumption
of dense AP deployment by incorporating inertial sensors.

Clearly many of these crowdsourcing systems also require a
trajectory-recovery component like the one used in this work. However,
trajectory recovery is rarely straight-forward and/or accurate enough
in these systems. This is why they need to work with other components
together to map the signal environment. More complicated survey
trajectories like those shown in Figure \ref{fig:pdr-w1},
\ref{fig:pdr-2015_05_30__11_04_50} and
\ref{fig:pdr-2015_05_30__08_51_29} are very challenging for such
systems. This is especially true for those based on signal propagation
models like WiFi-SLAM~\cite{Ferris1} and WiFi
GraphSLAM~\cite{GraphSLAM}. Furthermore, they suffer from the general
problems outlined in the Introduction above.

In contrast our proposed system recovers the trajectory
explicitly and accurately, and is much simpler and more practical to
use. It does not rely on any signal propagation model  and reduces the
efforts of the signal survey to a simple walk around the
environment.

\section{Path Surveys \label{sec:path-survey}}

\begin{figure*}
	\centering
	
	\begin{tabular}{cc}
		Use \emph{PDR-traj} as input&Use \emph{SLAM-traj} as input\\
		
		\begin{tabular}{c|c}
			\hline
			\rotatebox{90}{Input}&
			\subfigure[\textit{PDR-traj} \label{fig:pdr-w1}]{\includegraphics[width=2.5cm,height=6cm,angle=90]{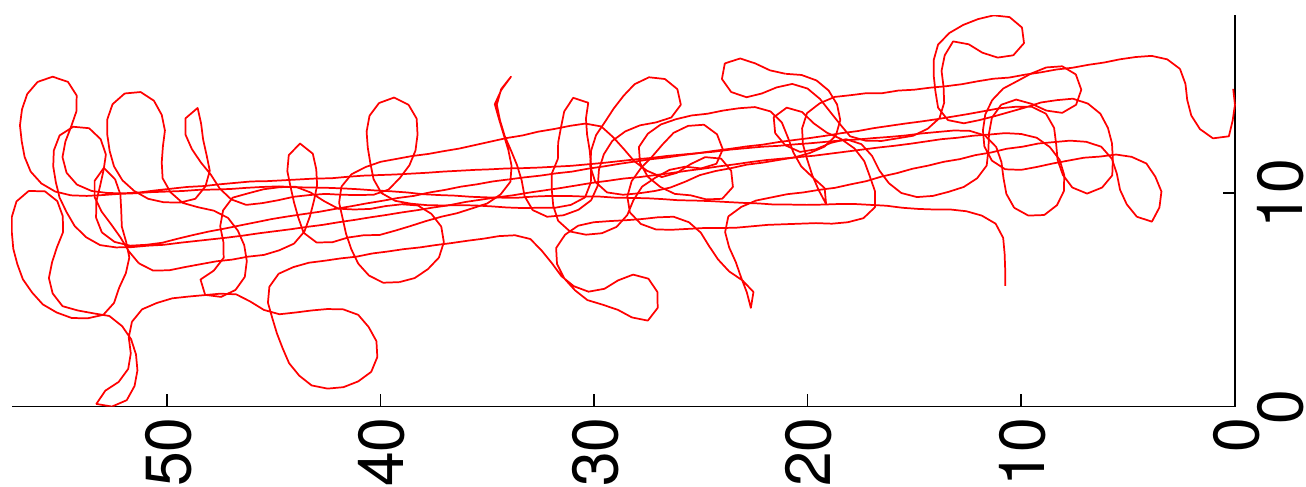}} \\
			\hline
			\rotatebox{90}{No smoothing}&
			\subfigure[\label{fig:pdr-non-smoothed}]{\includegraphics[width=2.5cm,height=6cm,angle=90]{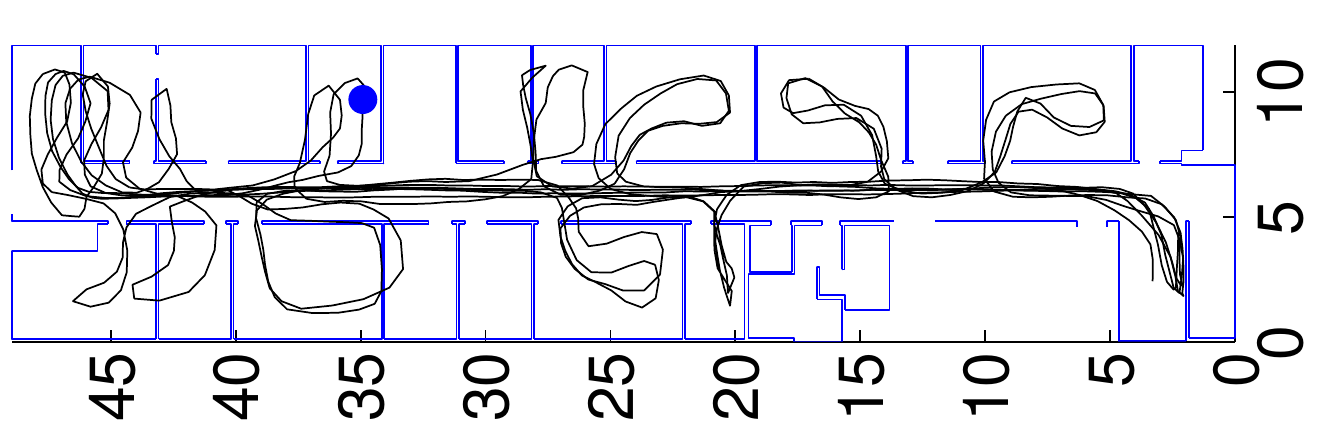}}\\
			\hline
			\rotatebox{90}{Pruning}&
			\subfigure[\label{fig:pdr-prune}]{\includegraphics[width=2.5cm,height=6cm,angle=90]{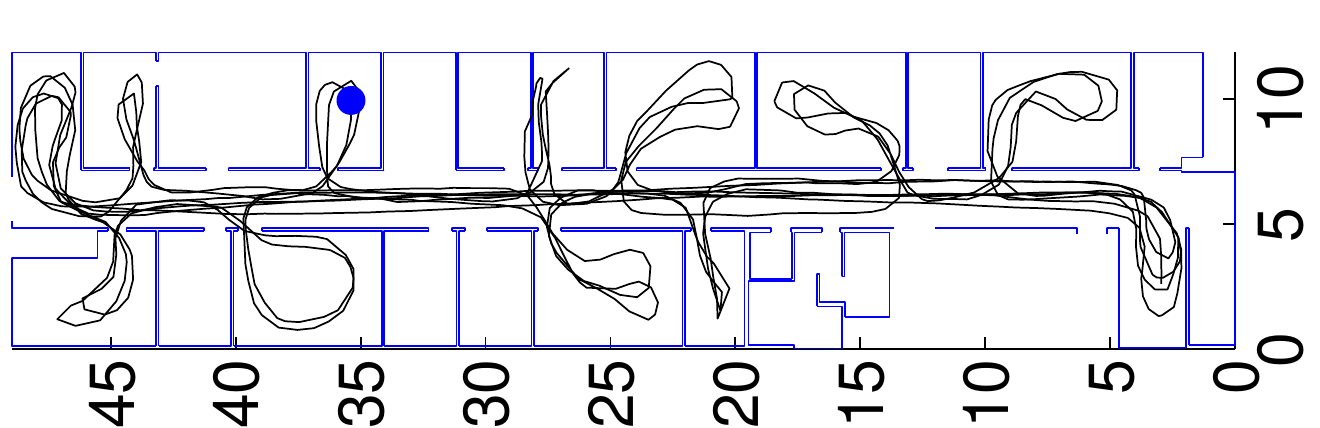}} \\
			\hline 		
			\rotatebox{90}{FL}&
			\subfigure[\label{fig:normal-pfs-on-w1-fl}]{\includegraphics[width=2.5cm,height=6cm,angle=90]{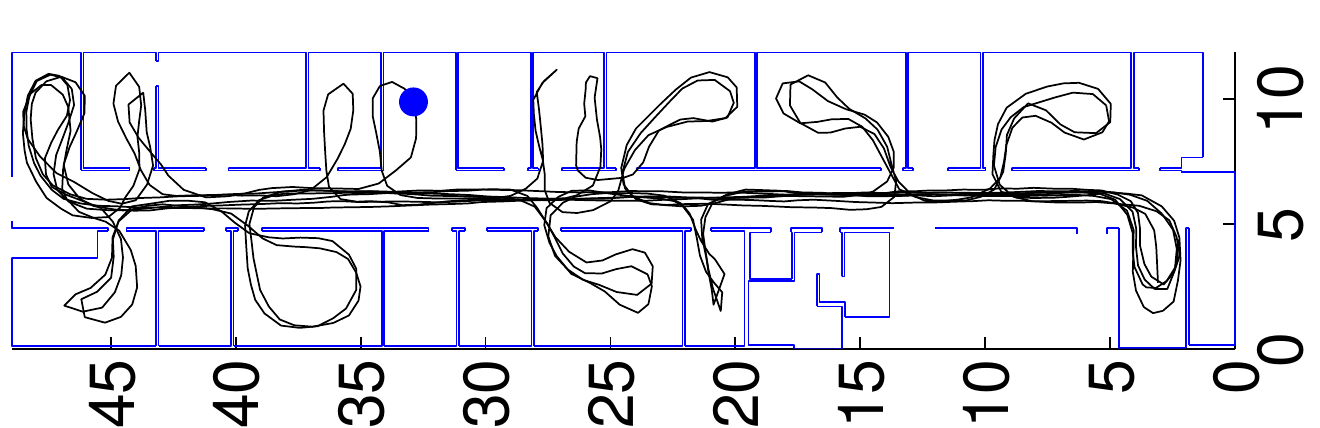}}\\
			\hline
			\rotatebox{90}{FFBSm}&
			\subfigure[]{\includegraphics[width=2.5cm,height=6cm,angle=90]{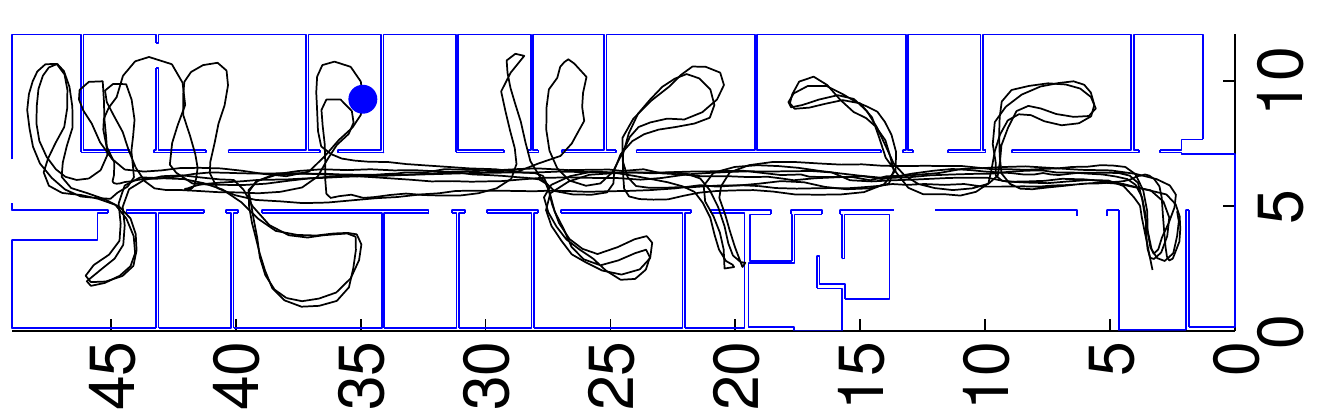}}\\
			\hline
			\rotatebox{90}{FFBSi}&
			\subfigure[\label{fig:pdr-ffbsi}]{\includegraphics[width=2.5cm,height=6cm,angle=90]{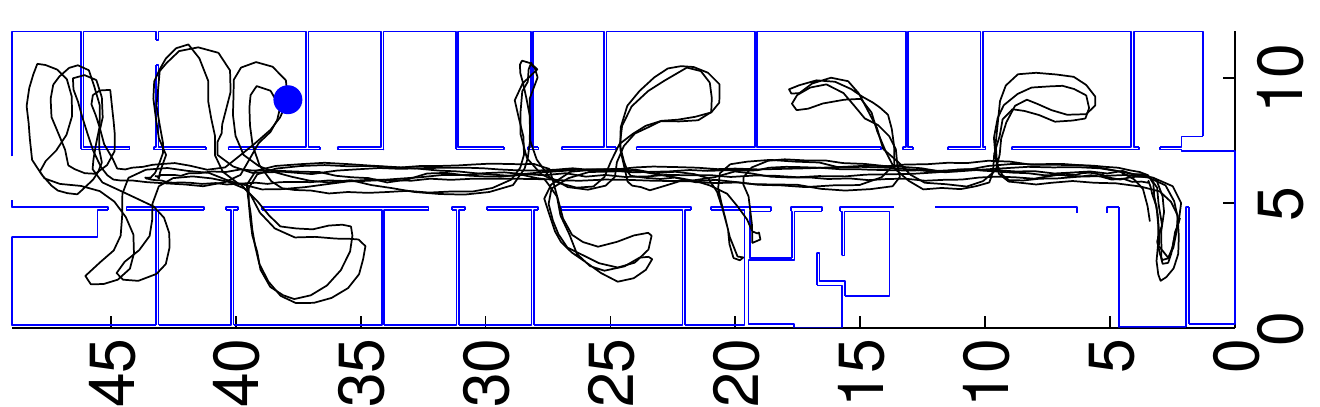}}\\
			\hline
			\rotatebox{90}{Particle Cloud}&
			\subfigure[\label{fig:pcloud-w1-normal-pdr}]{\includegraphics[width=2.5cm,height=6cm,angle=90]{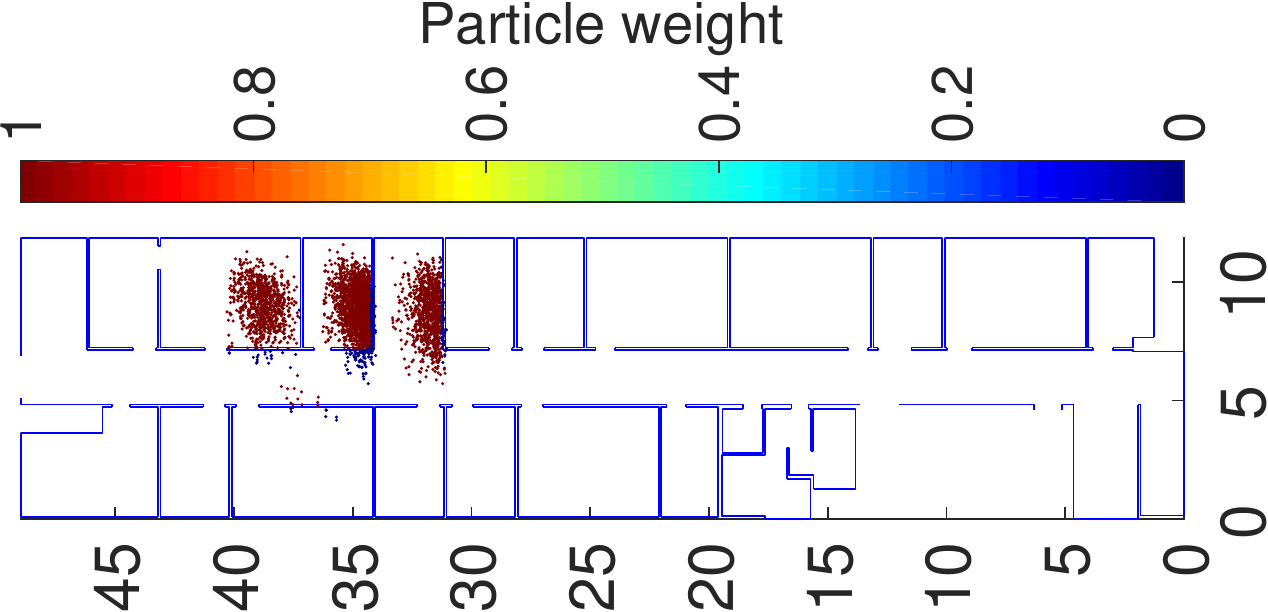}}\\
			\hline
		\end{tabular}
		
		&
		
		\begin{tabular}{c|c}
			\hline
			\rotatebox{90}{Input}&
			\subfigure[\textit{SLAM-traj}\label{fig:slam-traj-w1}]{\includegraphics[width=2.5cm,height=6cm,angle=90]{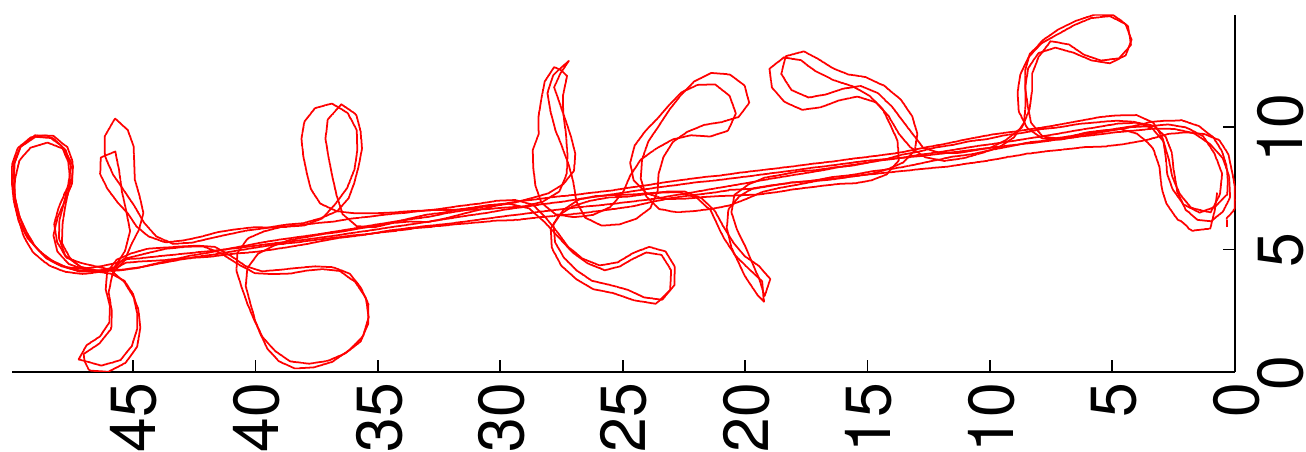}} \\
			
			\hline
			\rotatebox{90}{Non-Smoothed}&
			\subfigure[]{\includegraphics[width=2.5cm,height=6cm,angle=90]{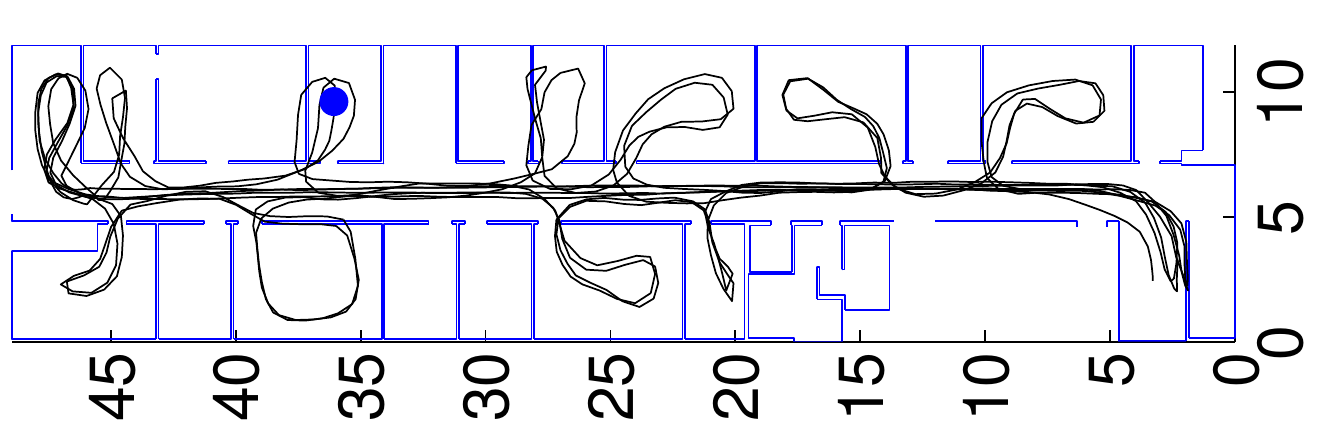}}\\
			\hline
			\rotatebox{90}{Pruning}&
			\subfigure[]{\includegraphics[width=2.5cm,height=6cm,angle=90]{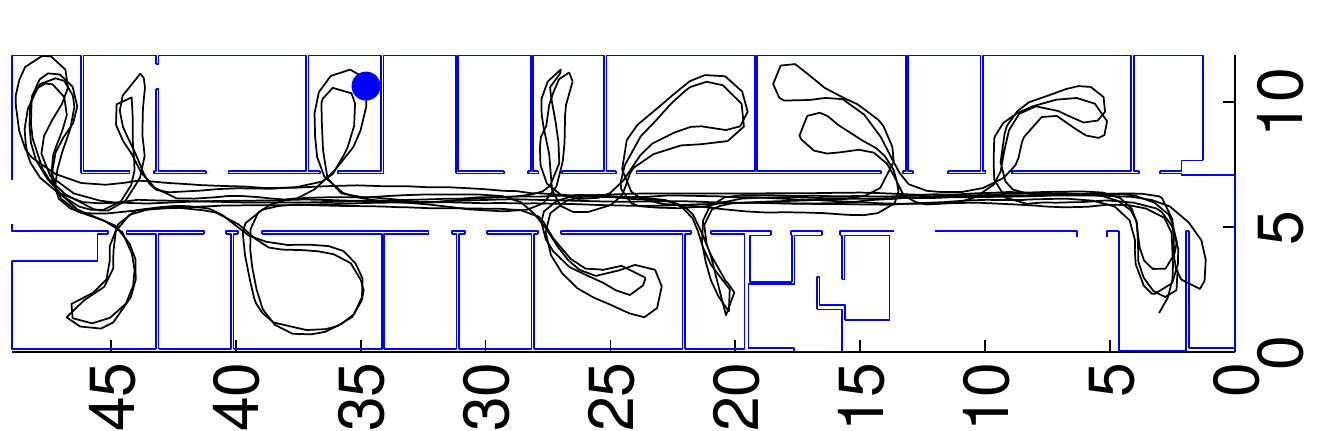}}\\
			
			\hline 		
			\rotatebox{90}{FL}&
			\subfigure[]{\includegraphics[width=2.5cm,height=6cm,angle=90]{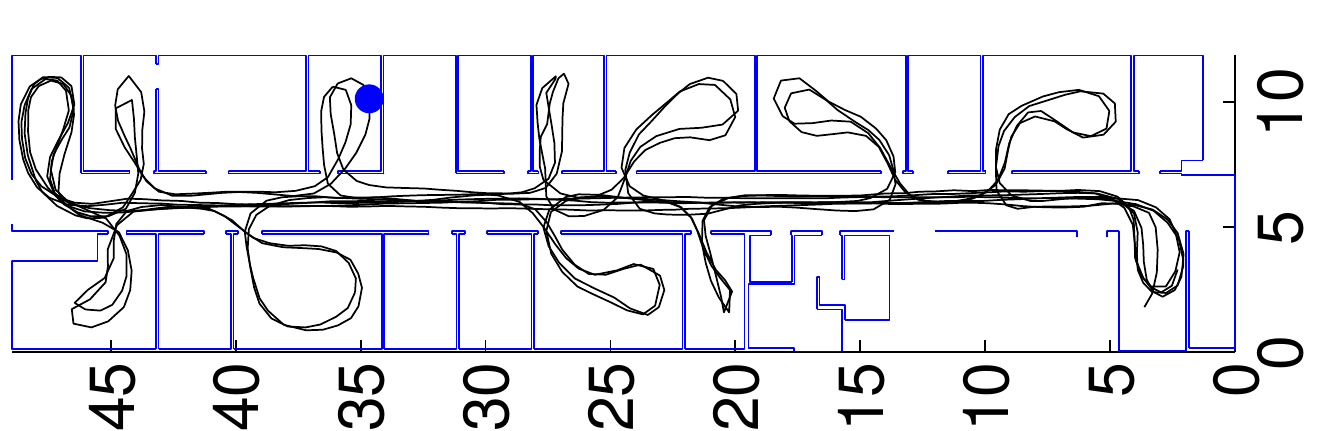}}\\
			\hline
			\rotatebox{90}{FFBSm}&
			\subfigure[]{\includegraphics[width=2.5cm,height=6cm,angle=90]{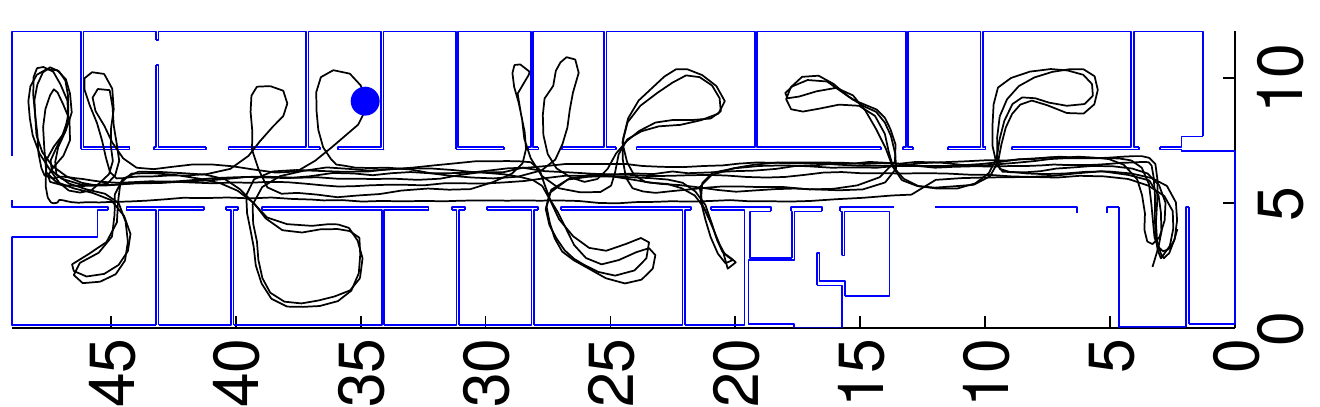}}\\
			\hline
			\rotatebox{90}{FFBSi}&
			\subfigure[]{\includegraphics[width=2.5cm,height=6cm,angle=90]{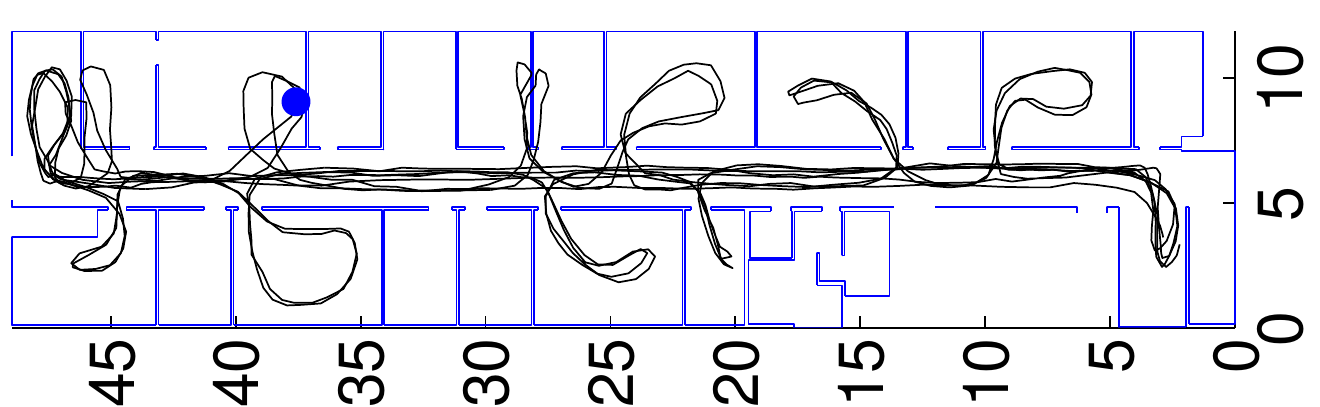}}\\
			\hline
			\rotatebox{90}{Particle Cloud}&
			\subfigure[\label{fig:pcloud-w1-normal-slam}]{\includegraphics[width=2.5cm,height=6cm,angle=90]{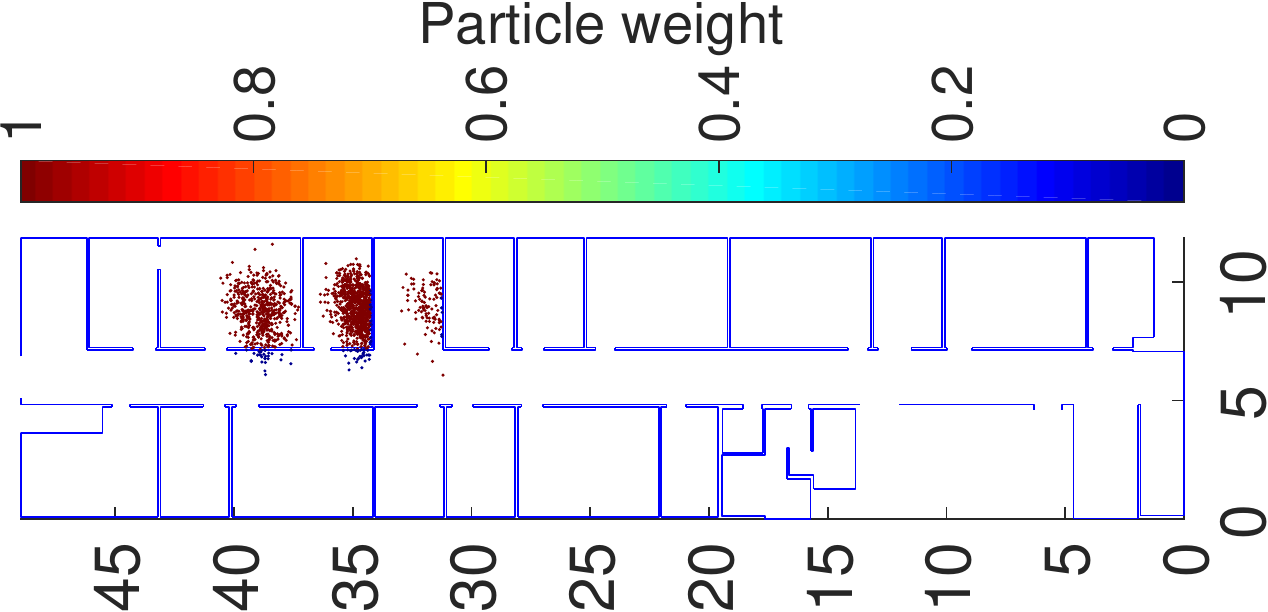}}\\
			\hline
		\end{tabular}
		
	\end{tabular}
	
	\caption{Example outputs of the conventional particle filter
		plus smoother approach. The ground truth trajectory is shown
		in Figure~\ref{fig:example-path-survey}. These results show that state-of-the-art techniques failed in giving robust trajectory estimation. }
	\label{fig:normal-pfs-on-w1}
\end{figure*}

We consider any PDR-based system that maps a spatially-variant
quantity using trajectories rather then grid point measurements to be
a \emph{path survey}. The primary goal of this work is to produce the
best trajectory estimate consistent with (and anchored to) a
floorplan. This section evaluates state-of-the-art techniques that
could potentially enable path survey. We try different combinations of
various techniques and show why they fail.

The simplest solution is to feed the raw PDR trajectory (herein
referred to as \emph{PDR-traj}) into a wall-sensitive particle filter
and then use a particle smoother to recover the optimal
trajectory. The left column of Figure \ref{fig:normal-pfs-on-w1}
illustrates this case. The raw PDR-traj path exhibits typical drift
issues seen when walking a path such as the ground truth (Figure
\ref{fig:example-path-survey}, obtained using a high accuracy
ultrasonic positioning system~\cite{Addlesee01}).  The error is sufficiently
high that various smoothers give significantly different trajectory
results (Figures
\ref{fig:pdr-non-smoothed}--\ref{fig:pdr-ffbsi}). Moreover, the
trajectories are not truly consistent with the floorplan, since they
cross walls (or enter wrong rooms) at various points. The explanation for this can be seen in
Figure~\ref{fig:pcloud-w1-normal-pdr}, which shows an instantaneous
particle distribution corresponding to the point marked with a blue
dot in the preceding images.  The high PDR uncertainty results in
multi-modal distributions spanning multiple rooms. The position
estimate is a weighted average of these particles and so can cross
walls. This \emph{room ambiguity} is very serious for a path survey: a
small perturbation to any of these systems could easily result in
signal data being assigned to the incorrect room and the subsequent
radio map containing serious errors.

\begin{figure*}
	\centering
	\begin{tabular}{ccc}
		\subfigure[Raw PDR result. \label{fig:slam-violate-fp-example-pdr}]{\includegraphics[height = 4cm,angle=90]{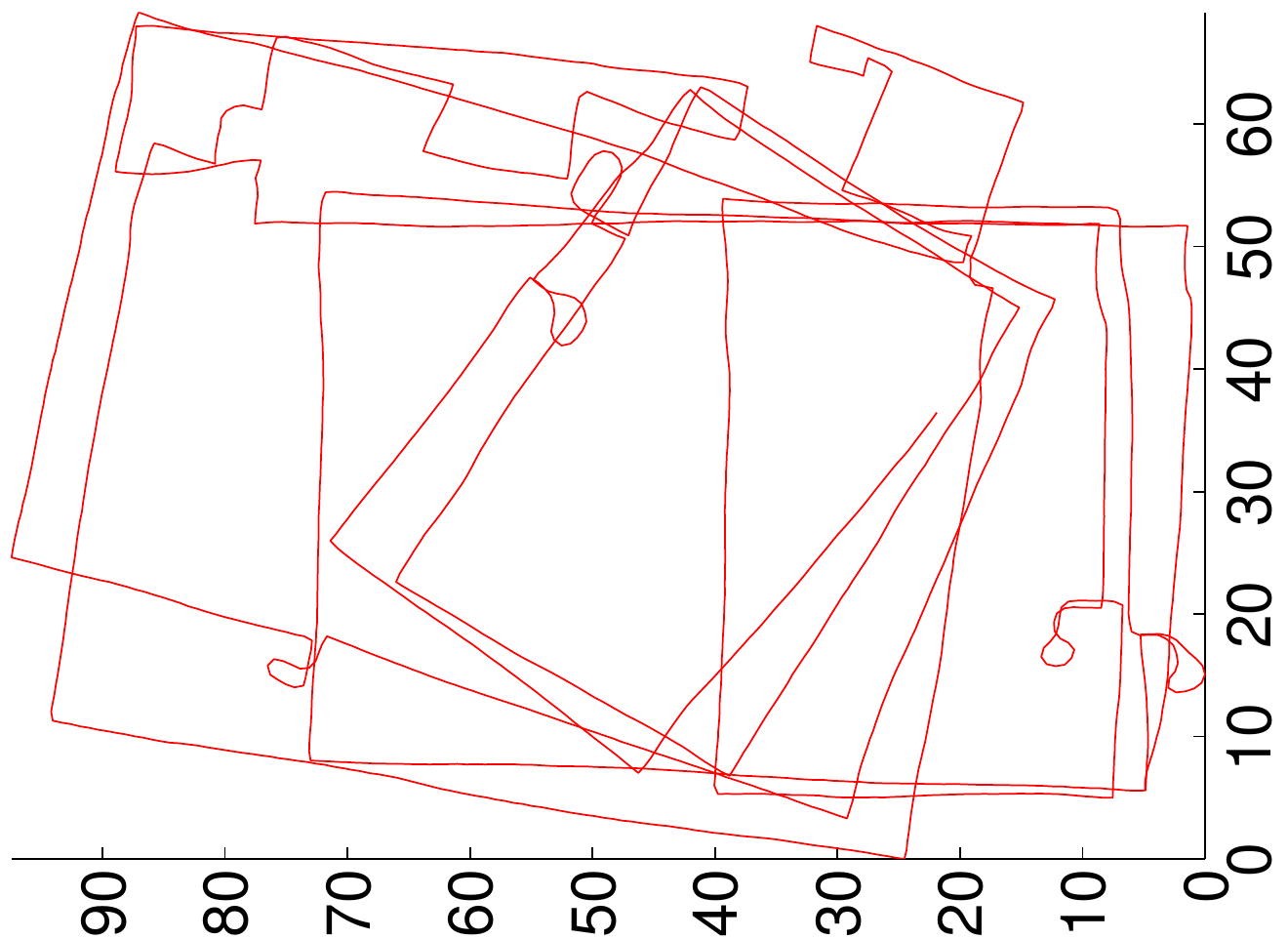}}
		&
		\subfigure[Pre-processed PDR result (black trajectory) with loop closures (dense red lines). \label{fig:slam-violate-fp-example-pdr-lp}]{\includegraphics[height = 4cm,angle=90]{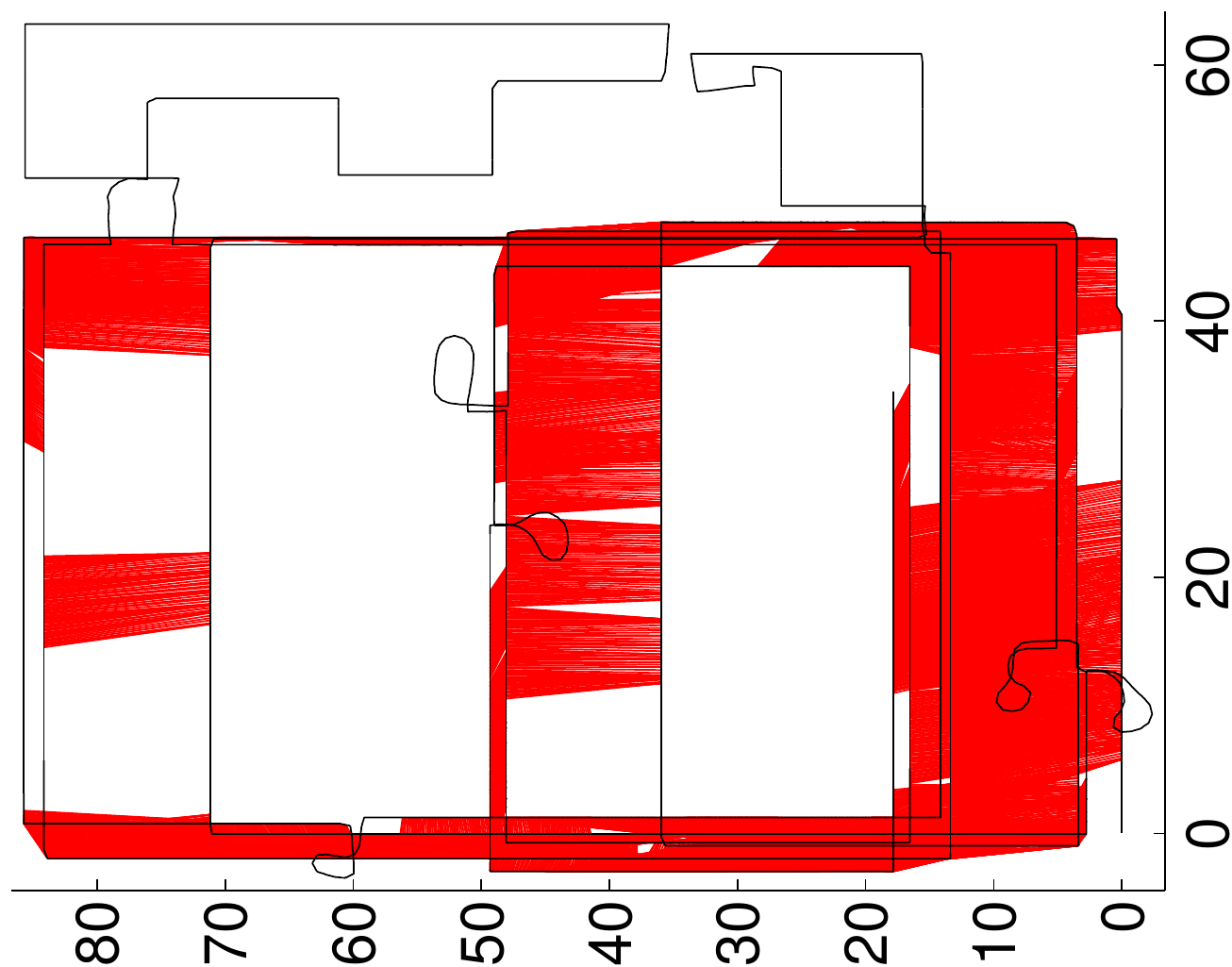}}
		&
		\subfigure[SLAM result plotted on floor plan. \label{fig:slam-violate-fp-example-slam}]{\includegraphics[width = 3.5cm]{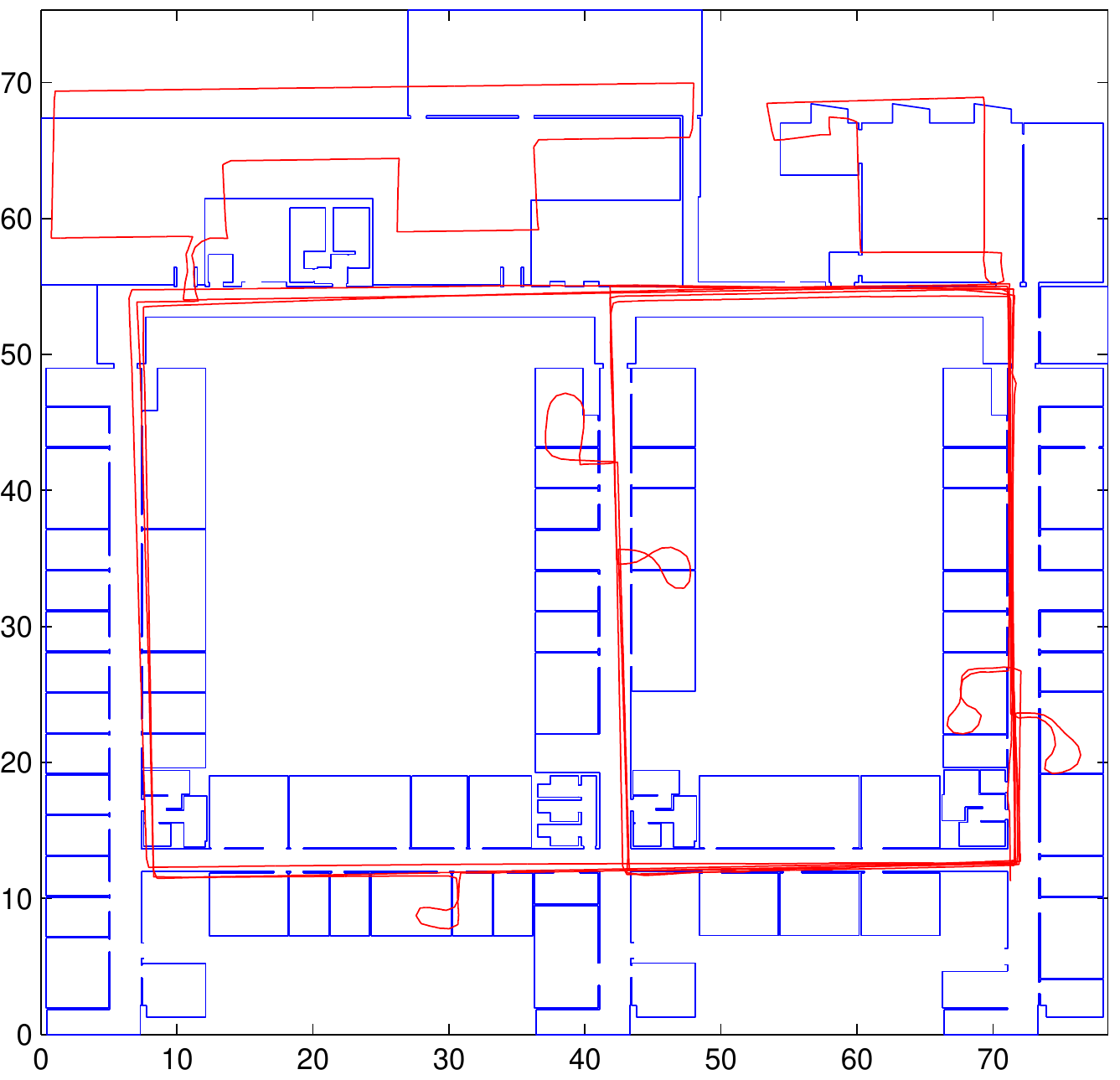}}\\
	\end{tabular}
	\caption{Example scaling errors from SLAM without a
		floorplan. (a) the input \emph{PDR-traj}. (a) the \emph{PDR-traj} was pre-processed by straight line filter and the loop closures were detected. (c) the SLAM result after manual alignment to the floorplan.}
	\label{fig:slam-violate-fp-example}  
\end{figure*}

To reduce the chance of multi-modal distributions we can preprocess
the PDR trajectory to correct the drift based on external
observations. For example we have previously demonstrated a
GraphSLAM-based system that does this by applying loop closure
constraints when a floorplan is
unavailable~\cite{gao2015sequence}. Figure~\ref{fig:slam-violate-fp-example-pdr-lp}
shows the loop closures on a typical (pre-processed) PDR
trajectory. These loop closures were detected by a window-based
searching scheme that matches similar magnetic sequences recorded
along the walking path. Based on these detected loop closures, the
GraphSLAM system could then correct the PDR drift to a large extent as
showed in Figure~\ref{fig:slam-violate-fp-example-slam} (the red
trajectory). However, the optimised trajectory may have incorrect
scale and might violate the environmental
constraints. Figure~\ref{fig:slam-violate-fp-example-slam} shows
the SLAM results on top of the floorplan (manually aligned). We
observe that the path crosses the walls and different parts of the
trajectory have different scaling errors. The reason for this is that
the loop closures only give information about the spatial
relationships between the surveyor's positions at different time
points/steps, but not any information about the environmental
constraints (floorplan). So a pure loop closure-based SLAM system is
not adequate to correctly recover the survey path.

\begin{figure*}
	\centering
	\begin{tabular}{c|c|c|c}
		\hline
		&Input (\textit{PDR-traj})&FL& Particle Cloud\\
		\hline
		\rotatebox{90}{\textit{Path-2}}&
		\subfigure[\label{fig:pdr-2015_05_30__11_04_50}]{\includegraphics[height=4cm]{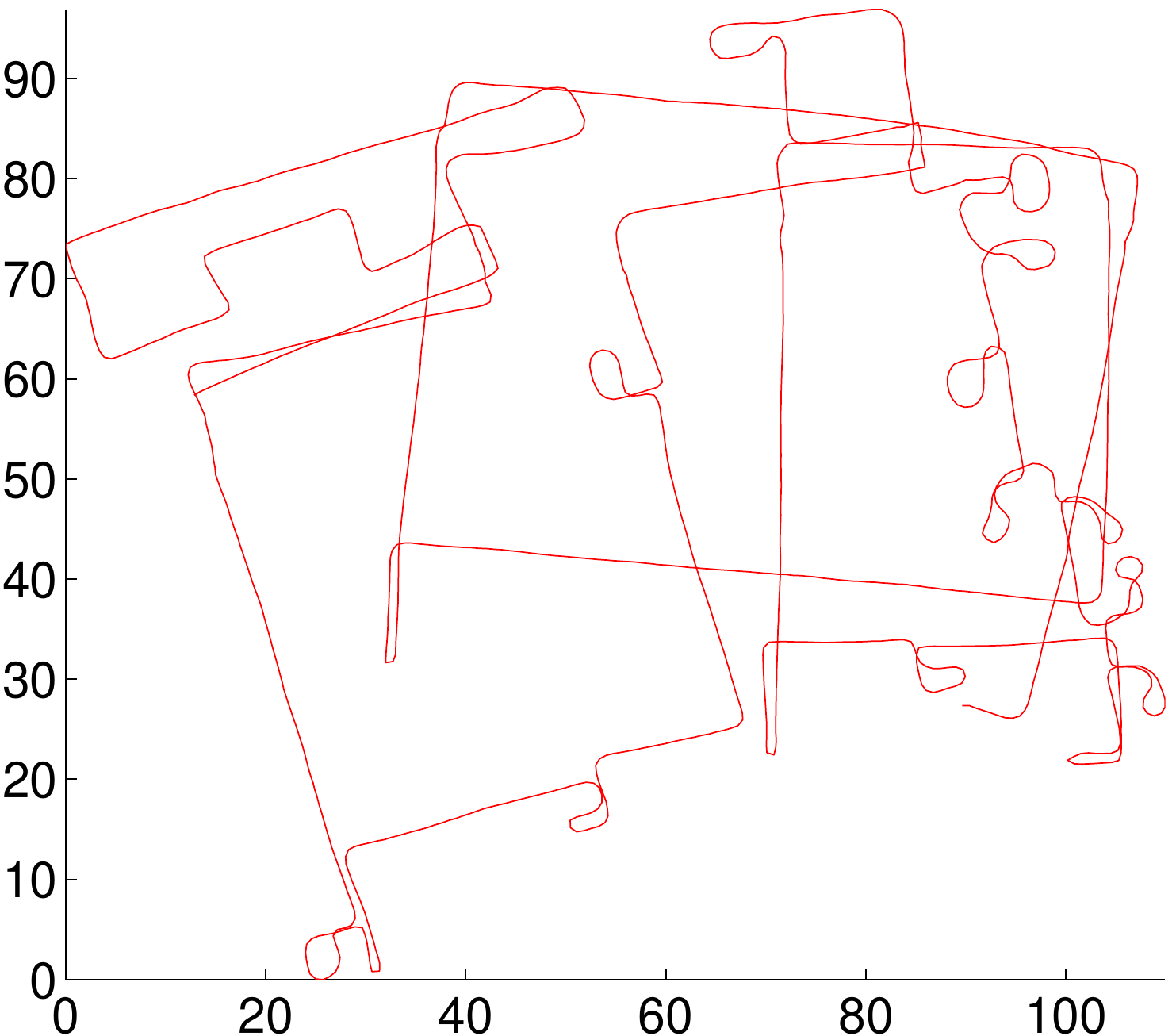}}
		&
		\subfigure[\label{fig:normal-pfs-on-2015_05_30__11_04_50-fl}]{\includegraphics[height=4cm]{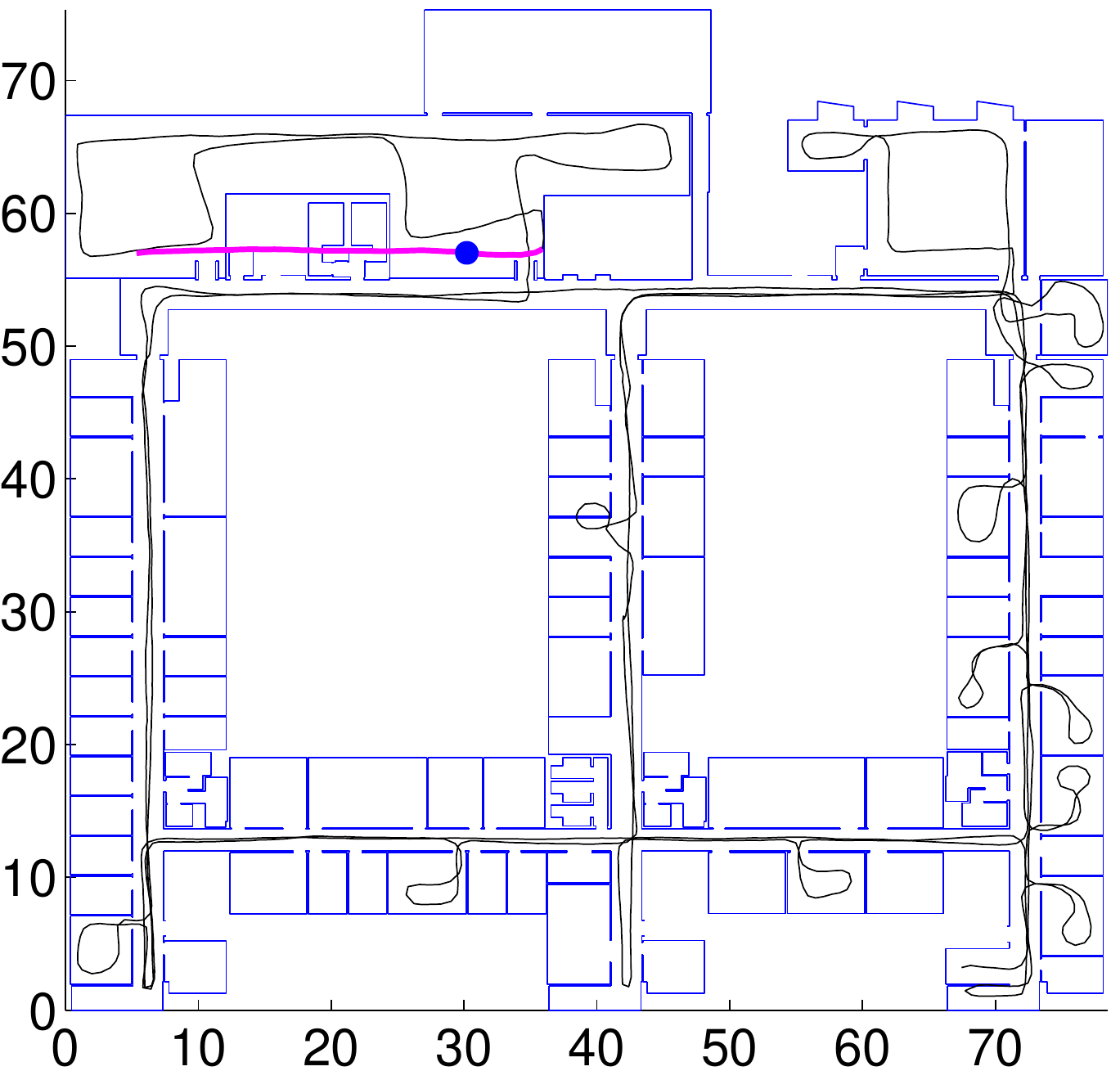}}
		&
		\subfigure[\label{fig:pcloud-path-2-normal-slam}]{\includegraphics[height=4cm]{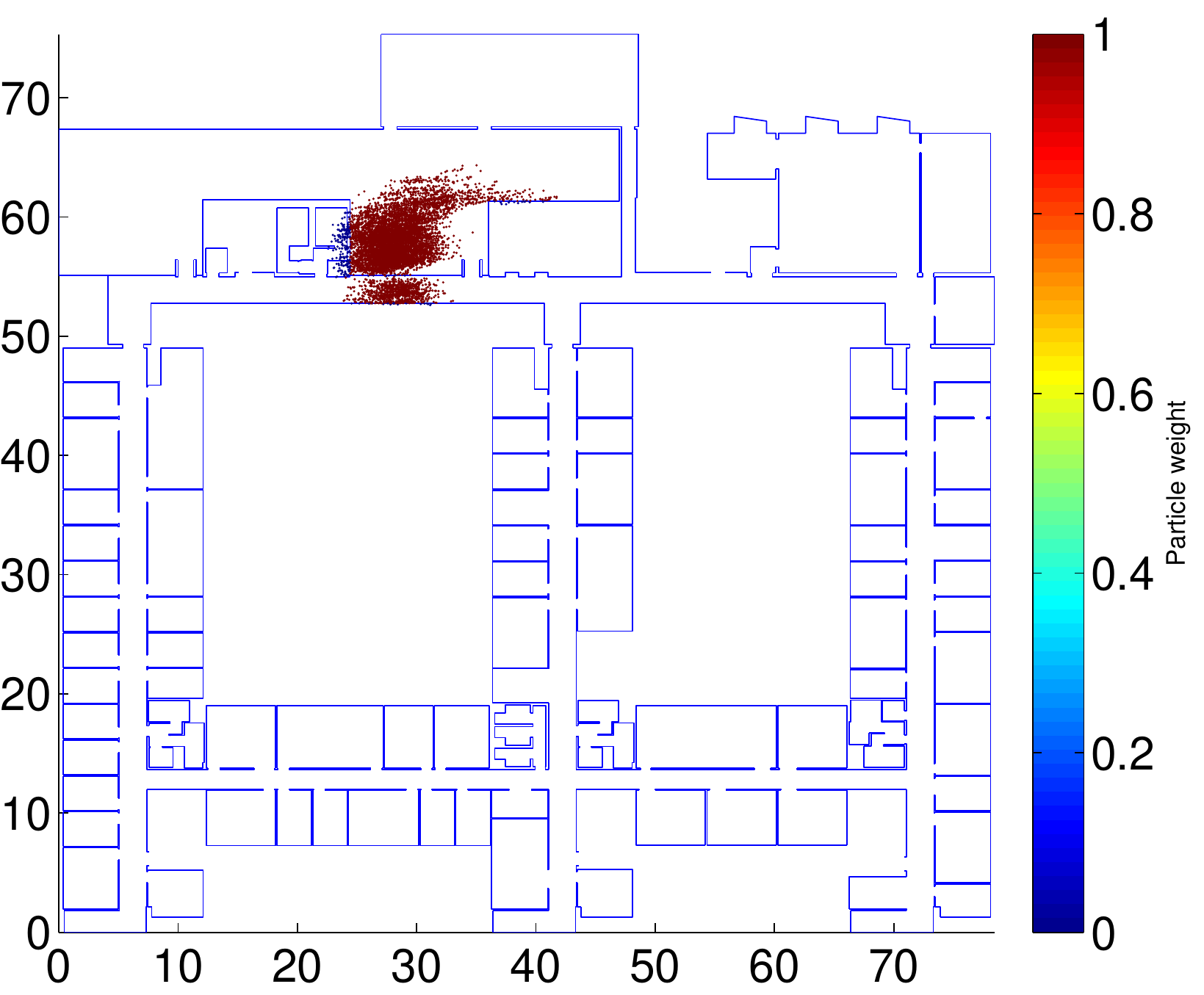}}
		\\
		\hline 
		\rotatebox{90}{\textit{Path-3}}&
		\subfigure[\label{fig:pdr-2015_05_30__08_51_29}]{\includegraphics[height=4cm]{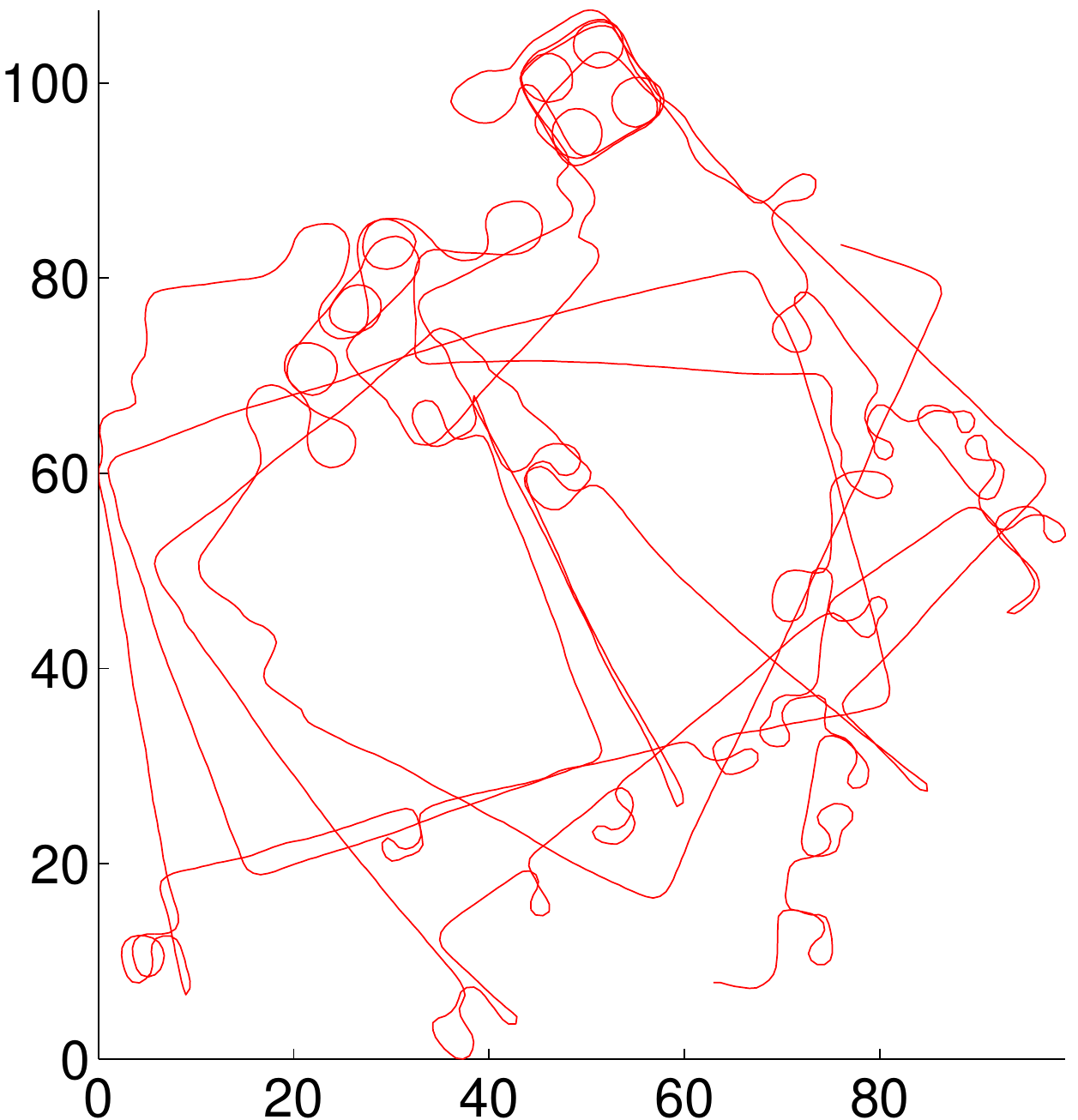}}
		&
		\subfigure[\label{fig:normal-pfs-on-2015_05_30__08_51_29-fl}]{\includegraphics[height=4cm]{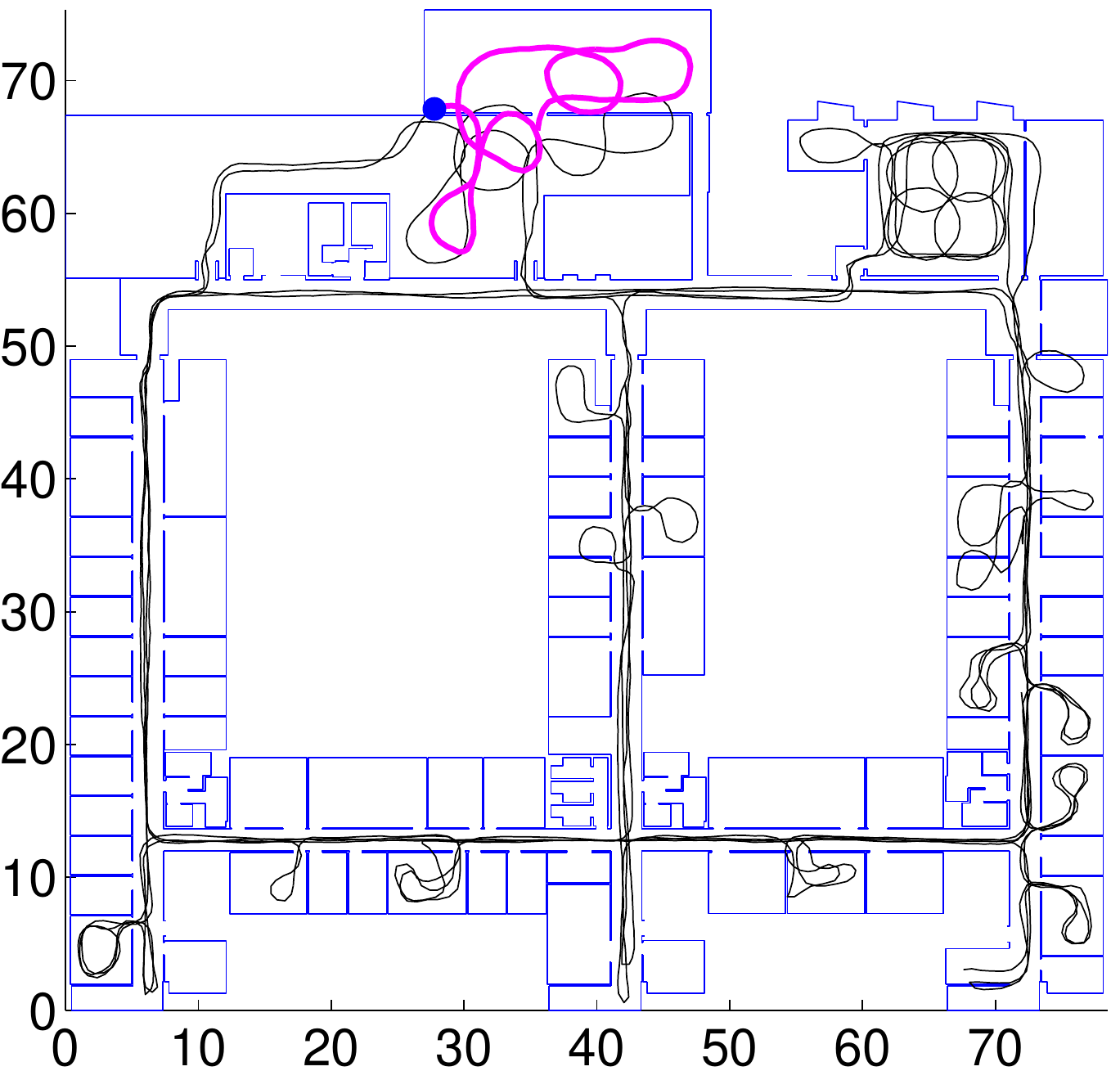}} 
		&
		\subfigure[\label{fig:pcloud-path-3-normal-slam}]{\includegraphics[height=4cm]{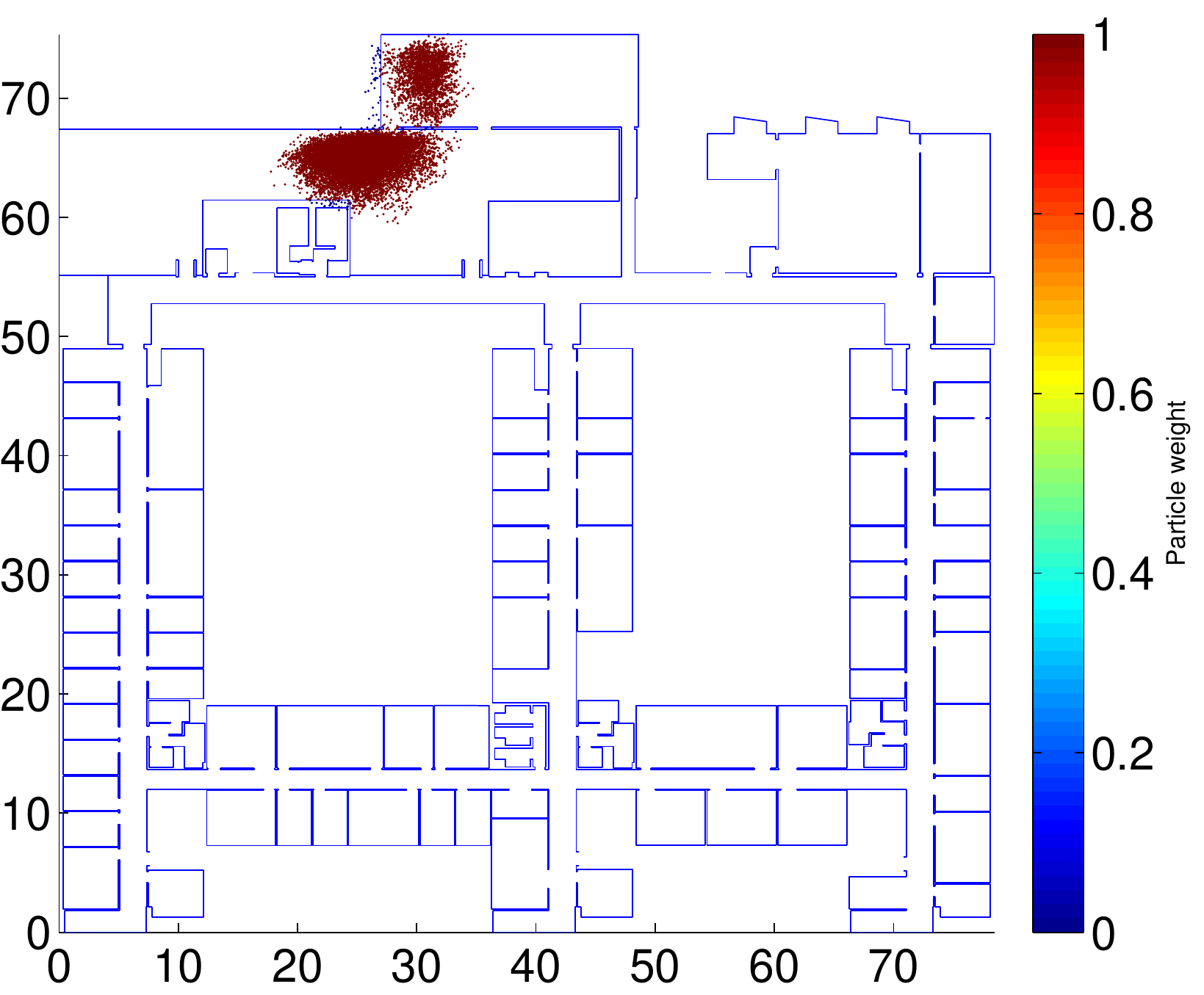}}
		\\
		\hline
	\end{tabular}
	\caption{Example outputs of the conventional particle filter
		plus smoother approach on longer walks.}
	\label{fig:normal-pfs-on-second-floor}  
\end{figure*}

A natural adaptation would be to take the low-drift (compared to raw
PDR trajectory) SLAM-corrected trajectory (herein referred to as
\emph{SLAM-traj}) and pass that through a floorplan-sensitive particle
filter and estimate the trajectory via a particle smoother or
pruning. The right column of Figure \ref{fig:normal-pfs-on-w1}
illustrates this idea. Compared to \emph{PDR-traj} (with both heading
and scaling errors) the \textit{SLAM-traj} has much lower heading
noise as can be seen in Figure~\ref{fig:slam-traj-w1}. However, the
scale errors persist and we typically find that the final result is
only marginally better. For these runs we see that only the FL
smoother was able to correctly recover the path (in fact part of the
trajectory still penetrates walls if observed carefully, but the error
is negligible). However, closer inspection of the particles at the
position marked with a blue dot in the smoother outputs shows that the
distribution was still multi-modal
(Figures~\ref{fig:pcloud-w1-normal-slam}).  This is more obvious for longer
walks such as those in Figure~\ref{fig:normal-pfs-on-second-floor},
where the FL smoother produced poor results (highlighted in magenta)
when faced with multi-modal distributions.

\section{PFSurvey Design}

We believe that a fundamental reason for the ambiguities in the
\textit{PDR-traj}-based filtering results discussed in Section
\ref{sec:path-survey} is the lack of sufficient constraints to
distribute weights reasonably over the particles. Please note that a particle represents a hypothesis about the 2D pose (coordinates x, y and orientation) of the surveyor. Each particle is associated with a weight represents the possibility of this hypothesis to be true. During the filtering process, particles with larger weights are more likely to be selected and survive than those with lower weights.
The wall-sensitive
particle filter and smoother kills all particles violating
environmental constraints and gives all the surviving particles the
same weight, e.g., a weight of $\frac{1}{N}$ for number of surviving
particles, $N$. Thus, all surviving particles have equal probability
of being re-sampled no matter how likely it is they hold the true
hypothesis of the system state. For the results in Figure
\ref{fig:normal-pfs-on-w1} and \ref{fig:normal-pfs-on-second-floor},
the particle clusters spread in the wrong locations have similar
weight sums with the particle clusters in the correct place. This
causes ambiguities and incorrectness in the final result. While loop
closures can assist by limiting the drift (i.e. keeping the particle
clouds small), the common techniques used to process them cannot
incorporate floorplan constraints.

PFSurvey attempts to use both the floorplan and loop closures
simultaneously to produce a more robust and accurate trajectory
estimate. The core fusion process is a particle filter, but a series
of pre-processing steps are necessary to create suitable loop
closures. The system architecture is illustrated in Figure
\ref{fig:system-with-floor-plan} and contains a series of components:

\begin{figure}[htbp]
	\begin{center}
		\includegraphics[height=6cm]{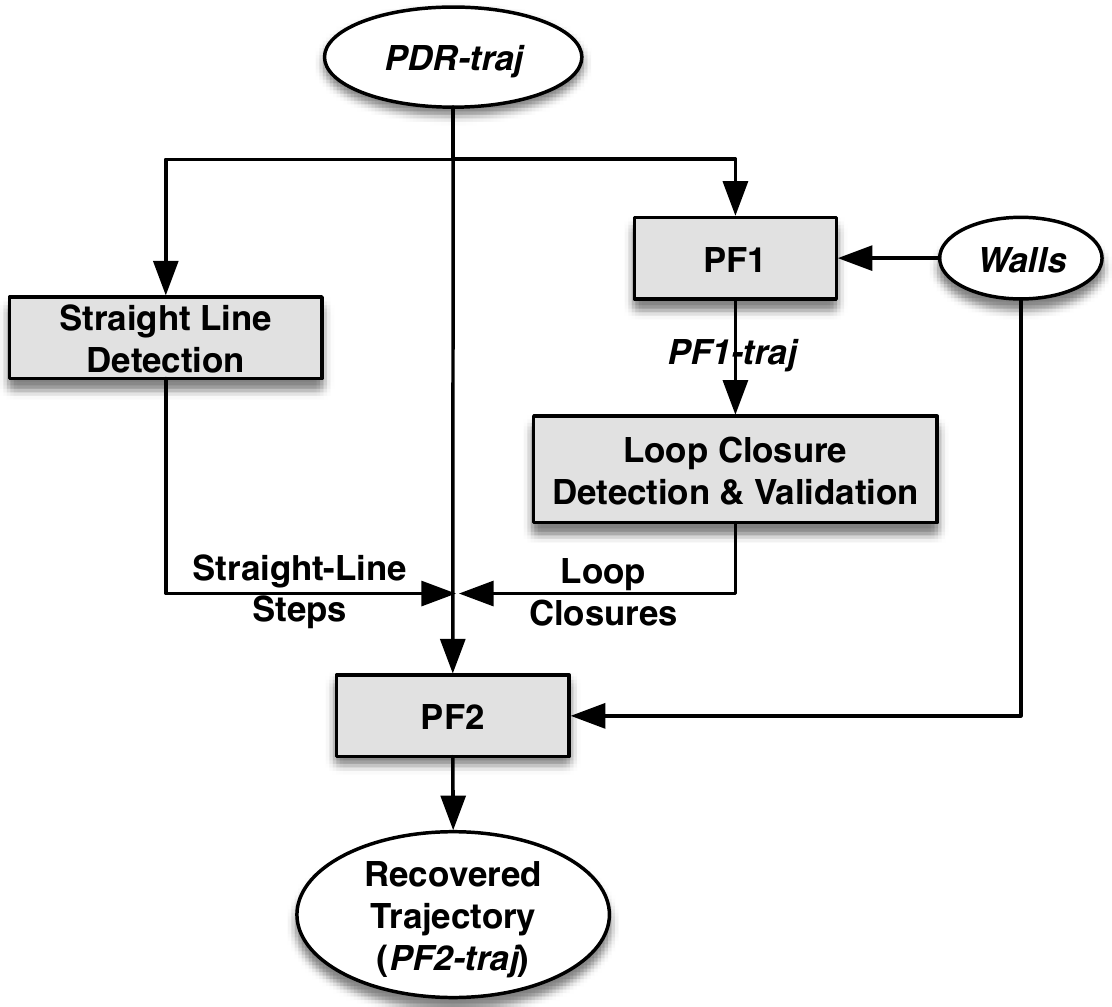} 
		\caption{The work flow of the proposed trajectory recovery system.}
		\label{fig:system-with-floor-plan}
	\end{center}
\end{figure}

	\textbf{Particle Filter 1 (PF1)}. This step
	takes the \textit{PDR-traj} as input, runs a
	wall-constraints-only particle filter to improve the topology
	of \textit{PDR-traj}. This is to bound the PDR drift using the
	floor plan to reduce heading errors. We denote the resultant
	trajectory of this step as \textit{PF1-traj}.

	\textbf{Straight line constraints}. We typically build our
	environments to be rectilinear and tend to move in
	straight lines. However, in the PDR results straight line
	steps often bend due to bias errors from the gyroscopes. We
	use a simple threshold-based method to identify
	probable straight-line steps in the PDR results. Then we weight
	particles in PF2 according to how well their
	headings match the orientations suggested by the
	environments (rooms or corridors).
	
	\textbf{Loop closure detection \& validation}.  This step
	detects and validates loop closures using the
	partially-corrected \textit{PF1-traj}. Without the initial
	PF1 pass, robustly identifying true loop closures is often all
	but impossible. We provide details of our detection and
	validation scheme in Section~\ref{sec:lc-detection-validation}.
	
	\textbf{Particle Filter 2 (PF2)}. This step adopts a
        customised particle filter to produce a
        survey trajectory consistent with the environment. It
        uses the wall, straight-line and loop closure constraints to
        weight the particles using \textit{PDR-traj} as input. We
        denote the resultant trajectory of this step as
        \textit{PF2-traj}, which is the final output. We now consider
        each of these components in more detail.
	

\subsection{PF1 \label{sec:pf1}}
PF1 takes the noisy \emph{PDR-traj} as input and aims to correct large
heading errors in the raw \emph{PDR-traj} using the floorplan. PF1 is
based on previous work using particle filters and a foot-mounted
IMU~\cite{woodman2008}. We adapt it to use 2D step vectors from
handheld smartphones, which are significantly more noisy than the
inputs used in the original work. In particular the step length is
unobservable. We represent a step event as $m_{i} = (l, \delta\theta_{i})$,
where $l$ is a fixed step length of 0.75~m and $\delta\theta_{i}$ is
the heading change of the surveyor during this step as estimated by
the gyroscope. The error models for these two components are assumed
to be independent and Gaussian:

\begin{equation}
\label{eq:step__event_error_models}
\begin{split}
e_{l} & \sim \mathcal{N}(0, \sigma_{l}^2) \\
e_{\delta\theta} & \sim \mathcal{N}(0, \sigma_{\delta\theta}^2)
\end{split}
\end{equation}

Considering the sensor characteristics of modern smartphone, we set
the uncertainty in the heading change $\sigma_{\delta\theta} =
0.5^\circ$ and the step length uncertainty to $\sigma_{l} = \lambda
l$\footnote {Our step variance is proportional to the step length,
  although that quantity is a constant here.}. We set $\lambda =
0.5$. The large uncertainties mean we are not sensitive to the
parameter values, at a small cost of additional computation resource.

PF1 (and indeed PF2) implement KLD adaptive
resampling~\cite{fox2001kld,woodman2008} to dynamically vary the
number of particles appropriately at each step and constrain the
computation requirements without impacting the result
quality. KLD-sampling adapts the particle number based on the
uncertainties of the system and some pre-defined parameters. Particles
from one generation are segmented into bins based on their location
and the number of occupied bins, $k$, used to determine the number of
particles in the new generation. In the discussion that follows we
briefly justify our other KLD parameters, full descriptions of which
can be found in~\cite{fox2001kld, woodman2008,WoodmanPhdThesis}. We
determined the KLD parameters as follows:

\begin{enumerate}
	\item
	\textbf{The bin sizes $\Delta_x$, $\Delta_y$ and
          $\Delta_\theta$.} Smaller bin sizes mean more bins occupied
        by the particle cloud hence cause more particles to be
        sampled, giving generally better performance. Since PF1 is
        solving the standard wall-constraints-only problem, we adopt
        parameters that have been shown to work well for that
        task~\cite{WoodmanPhdThesis}: $\Delta_x = \Delta_y  = 2.0~m$
        and $\Delta_\theta  = 30^\circ$.
	
	\item
	{\bf The bounding parameters $\delta$ and
          $\epsilon$}. KLD-sampling ensures that the error introduced
        by the sample-based approximation of the posterior
        distribution is below a specified threshold $\epsilon$ with
        probability $\delta$. We set $\delta = 0.01$ as
        in~\cite{WoodmanPhdThesis,fox2001kld,fox2003adapting}. We
        compute a value for $\epsilon$ such that the required number
        of particles is 300 for small $k$. Using the following
        standard approximation,

        \begin{equation}
        \label{eq:kld}
        n_{req} \approx \frac{k-1}{2\epsilon}(1-\frac{2}{9(k-1)}+\sqrt{\frac{2}{9(k-1)}}z_{1-\delta})^{3}, 
        \end{equation}

        where $z_{1-\delta}$ is the upper $1-\delta$ quantile
        of the standard normal distribution, we find $\epsilon = 0.0109238$ for
        $n_{req}=300$ and $k=2$.
	
	\item
	\textbf{The minimum number of particles $n_{min}$ required at
          each step}. KLD resamples particles until the number is
        greater than both $n_{req}$ (Equation \ref{eq:kld}) and
        $n_{min}$. The latter is important since small $n_{min}$ is
        risky: when the particle cloud has converged to just a few
        bins, the particles may all fall into a subset of bins that
        does not contain the true state simply by chance. This causes
        $n_{req}$ to be too small. In this case, if $n_{min}$
        is not large enough, insufficient particles will be sampled
        and the true state may be lost. To avoid this we set $n_{min}$
        to be the number required by KLD sampling when the particle
        cloud has converged to an area with size $\Delta_x*\Delta_y$
        in the $x-y$ plane and all the bins in the $\theta$-dimension
        are occupied, i.e.,
	\begin{equation}
	n_{min} = n_{req}\left( k = \frac{2\pi}{\Delta_\theta}\right)
	\end{equation}
	
	This gives $n_{min}=504$.
\end{enumerate}

With these parameters PF1 typically uses hundreds of particles and can
finish within 3$\sim$5 seconds for a 10-minute survey walk (about 950
steps), assuming the surveyor provides the initial room or corridor (a
reasonable expectation for a dedicated surveyor). To generate an
output path we apply the pruning smoother. Since we do not care about
room ambiguities in the output of PF1, it offers the least
resource-intensive solution. Other smoothers could be applied but
would come at additional complexity for no particular accuracy gain.

\subsection{Straight Line Filter}

This component first identifies candidates for \textit{straight-line
  steps} from the original PDR-traj. It uses a simple threshold-based
method: we identify consecutive steps where
the turning angle is less than t$^{\circ}$ to form candidate straight
line steps. Considering the gyro bias is relatively low and the surveyor is required to carry the smartphone consistently (holding the phone flat in front of the human body as if navigation), we set $t = 5$ and found this value works well in our case. Where there are $l$ or more candidates in a row, we assert that they are all straight line steps. The remaining candidates are
discarded. To ensure the robustness of the straight line filter, the value of $l$ should not be too small because this may cause many false positive straight-line movements being detected. And $l$ cannot be very large otherwise it would exceed the length of the longest straight corridor in the environment. By tests we found setting $l = 10$ is suitable for most indoor environments. The evaluation and discussion of this are given in Section~\ref{sec:sline-detect}.

\subsection{Loop Closure Detection and Validation \label{sec:lc-detection-validation}}

For optimal results, this component must detect a sufficient number of
true-positive loops in the trajectory without also detecting many
false-positive loops. In principle loops can be detected directly from
the PDR-traj (e.g. by looking for the same external signal values at
different times). This leads to a large number of false positive
closures since signals can reasonably adopt the same values at
different spatial locations. We seek instead to find loop closures
that link parts of the estimated trajectory that are already spatially
close.  This is the motivation for the PF1: it corrects the large
heading errors that result in even true loop closures being spatially
far apart (and hence difficult to distinguish from false positive
closures).

We have developed closure detection algorithms based on monitoring
\emph{sequences} of magnetic readings~\cite{gao2015sequence}. Magnetic signals are
ideally suited to the task: they have strong variance over space, which is mainly due to the steel shells of most modern buildings; they are temporally stable indoors (standard variance is typically within $0.001$ Earth field strength~\cite{li2012feasible}); the sensors are low power with frequent updates. We have not found
them to be a good signal to map for subsequent localisation because
they are very easily influenced by small changes to the environment
and hence transient in nature.  But during a single dedicated survey
walk lasting minutes, it is safely to assume that the disturbance like electronic devices is minimal, i.e. the signal is stable. The technique
proceeds as follows:

\begin{figure*}
	\centering
	\begin{tabular}{cc|cc|cc}
		\hline
		\multicolumn{2}{c|}{Segment Pair}&\multicolumn{2}{c|}{Maximum Segment Pair 1}&\multicolumn{2}{c}{Maximum Segment Pair 2}
		\\
		\hline
		\subfigure[Segment A. \label{fig:longest-sg-2-a}]{\includegraphics[width=2.1cm,height=6cm]{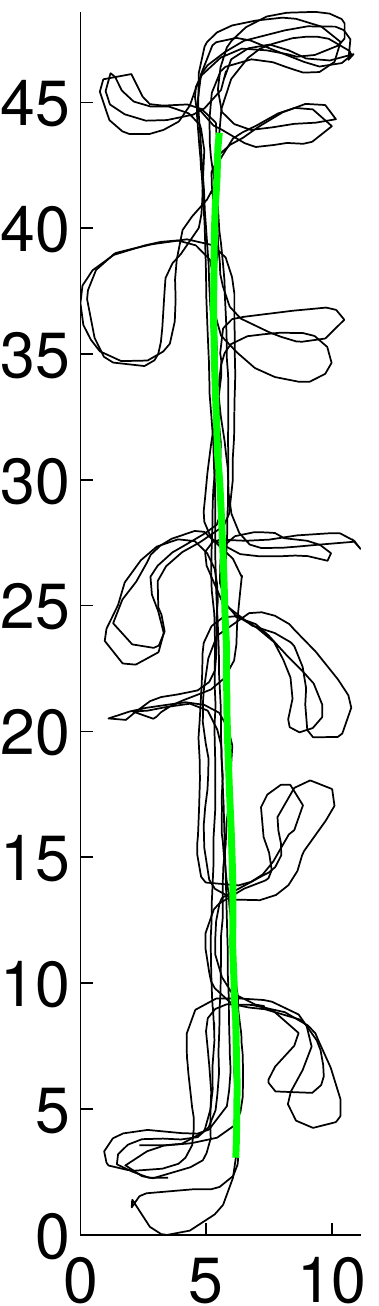}}
		& 
		\subfigure[Segment B. \label{fig:longest-sg-2-b}]{\includegraphics[width=2.1cm,height=6cm]{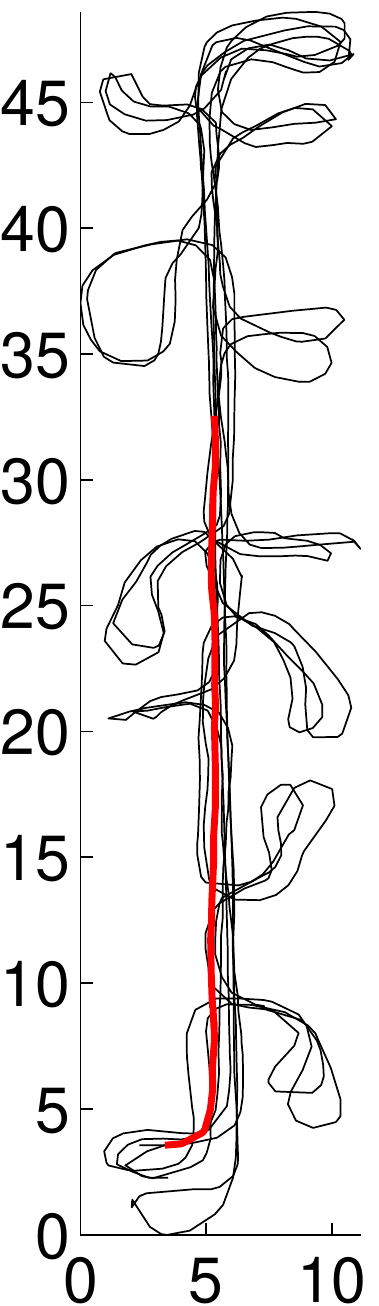}}
		&
		\subfigure[Segment A.]{\includegraphics[width=2.1cm,height=6cm]{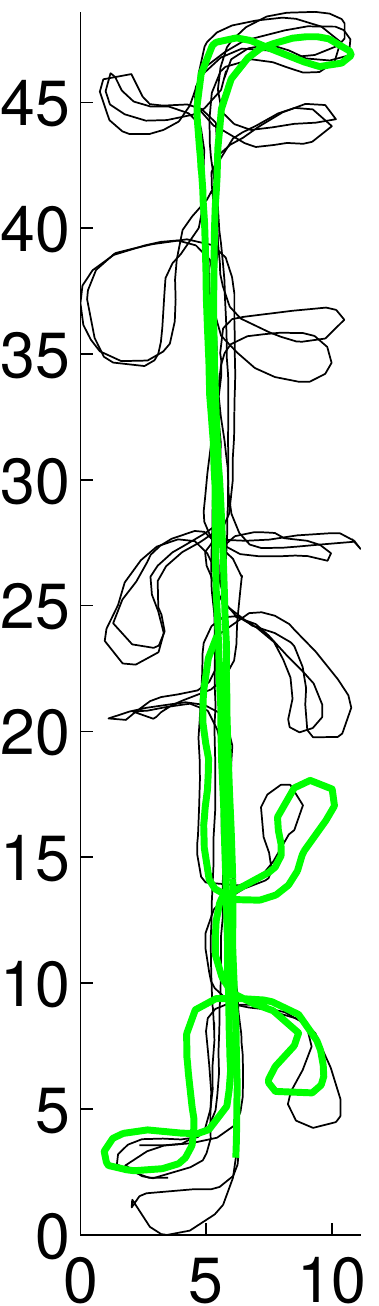}}
		&
		\subfigure[Segment B.]{\includegraphics[width=2.1cm,height=6cm]{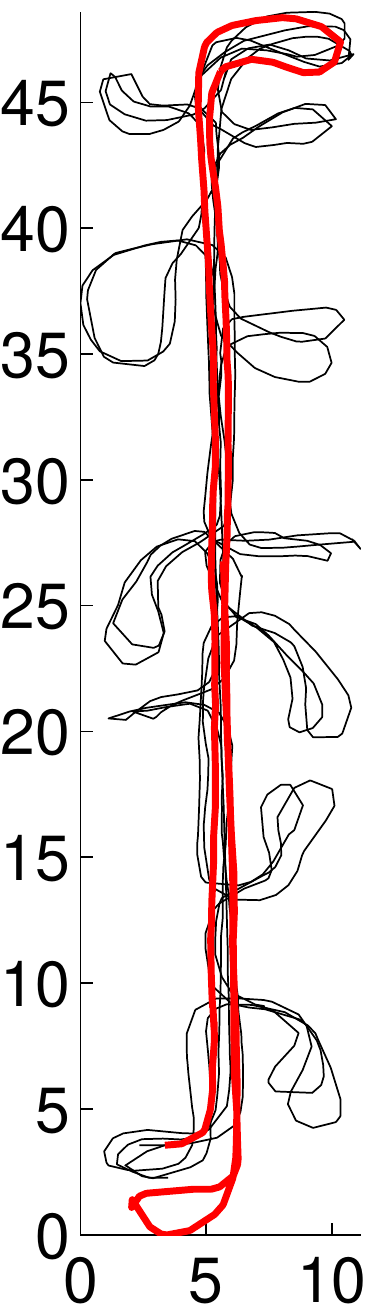}}
		&
		\subfigure[Segment A.]{\includegraphics[width=2.1cm,height=6cm]{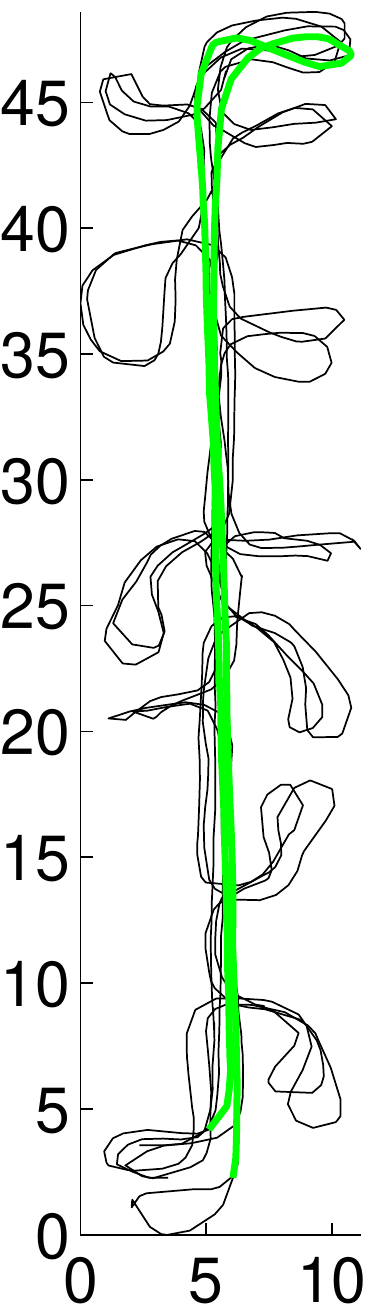}}
		&
		\subfigure[Segment B.]{\includegraphics[width=2.1cm,height=6cm]{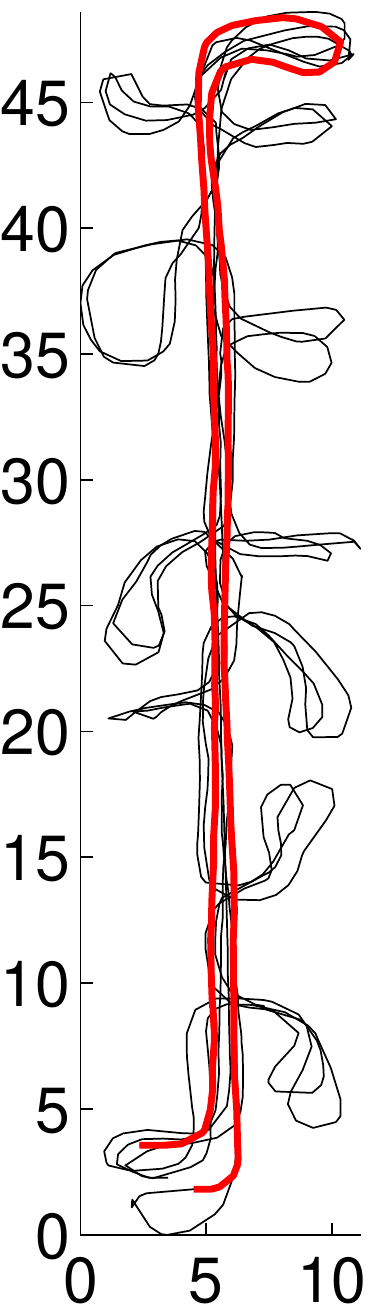}}
		\\
		\hline
	\end{tabular}
	\caption{Examples of \textit{PF1-traj}, segment pair and maximum segment pair}
	\label{fig:sg-examples}
\end{figure*}

\begin{figure*}
	\centering
	\begin{tabular}{c|c|c}
		\hline
		&Maximum Segment Pair 1&Maximum Segment Pair 2\\
		\hline
		\multirow{2}*{\rotatebox{90}{Sequence of Magnetic Magnitude\hspace{0.3cm}}}&&\\
		&\subfigure[Segment A. \label{fig:magnitude-a-sg-2}]{\includegraphics[height=1.6cm,width=7.5cm]{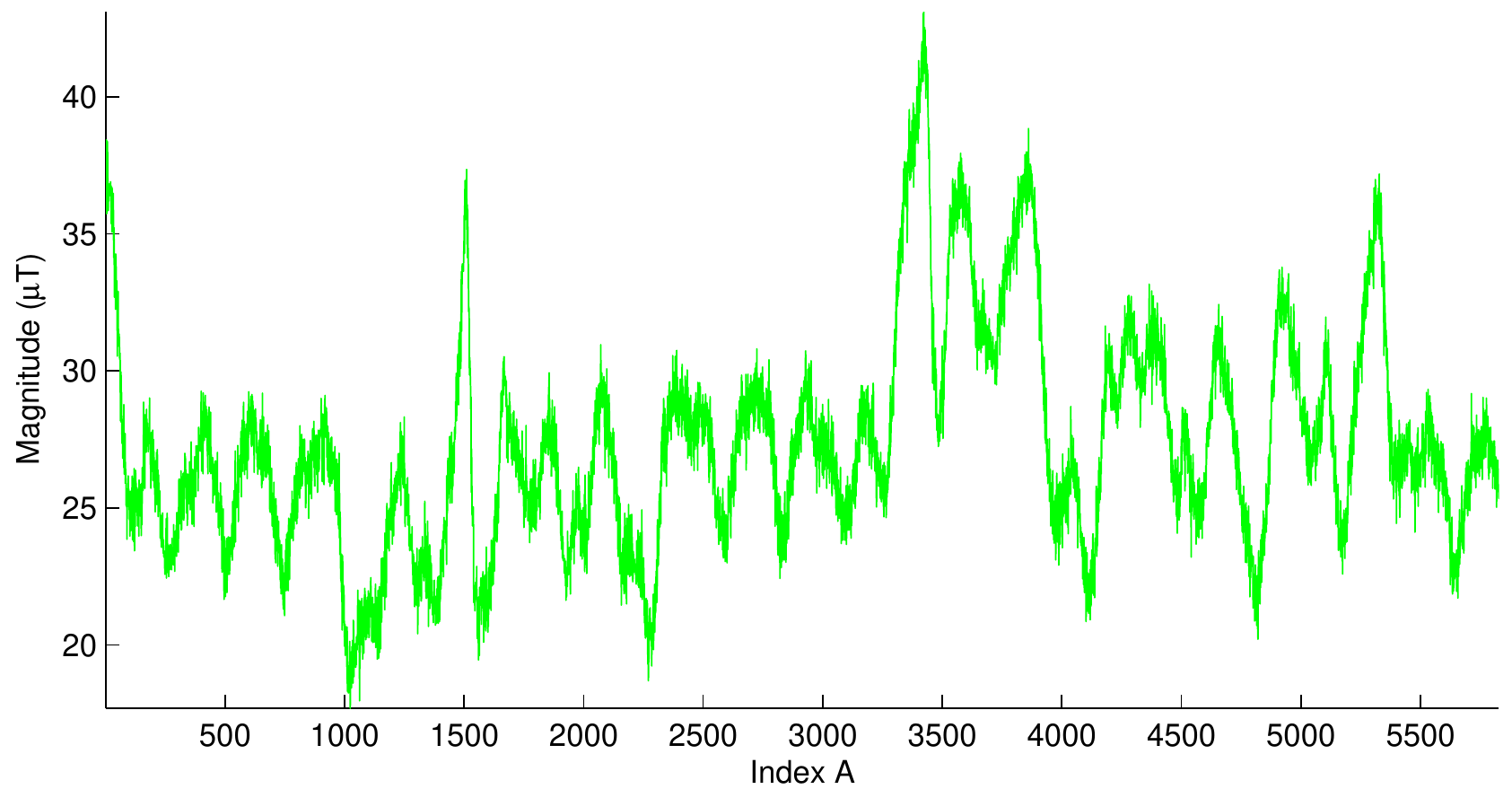}}&
		\subfigure[Segment A. \label{fig:magnitude-a-sg-37}]{\includegraphics[height=1.6cm,width=6.43cm]{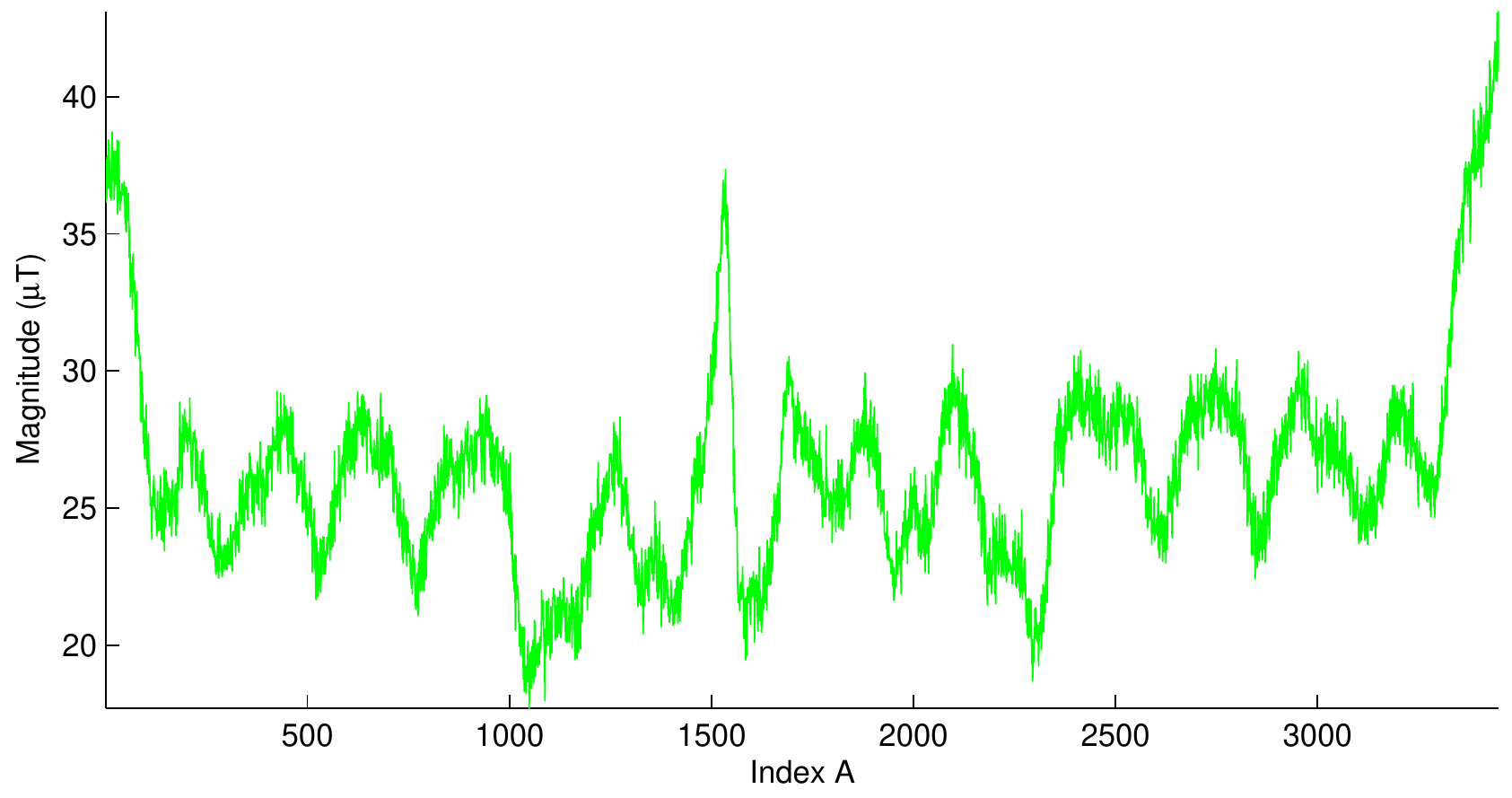}} \\ 	
		&\subfigure[Segment B. \label{fig:magnitude-b-sg-2}]{\includegraphics[height=1.6cm,width=6.14cm]{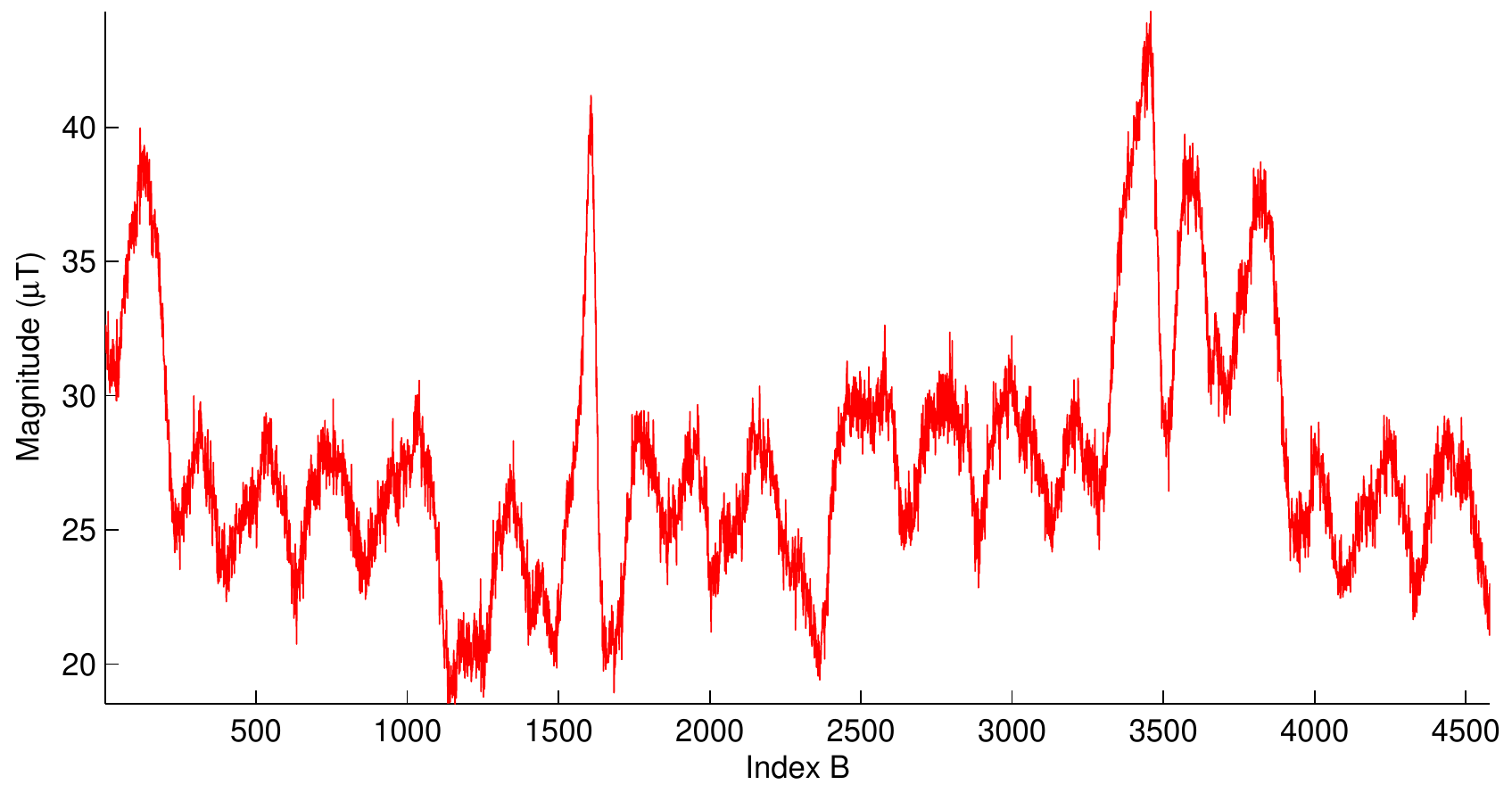}}&
		\subfigure[Segment B. \label{fig:magnitude-b-sg-37}]{\includegraphics[height=1.6cm,width=7.5cm]{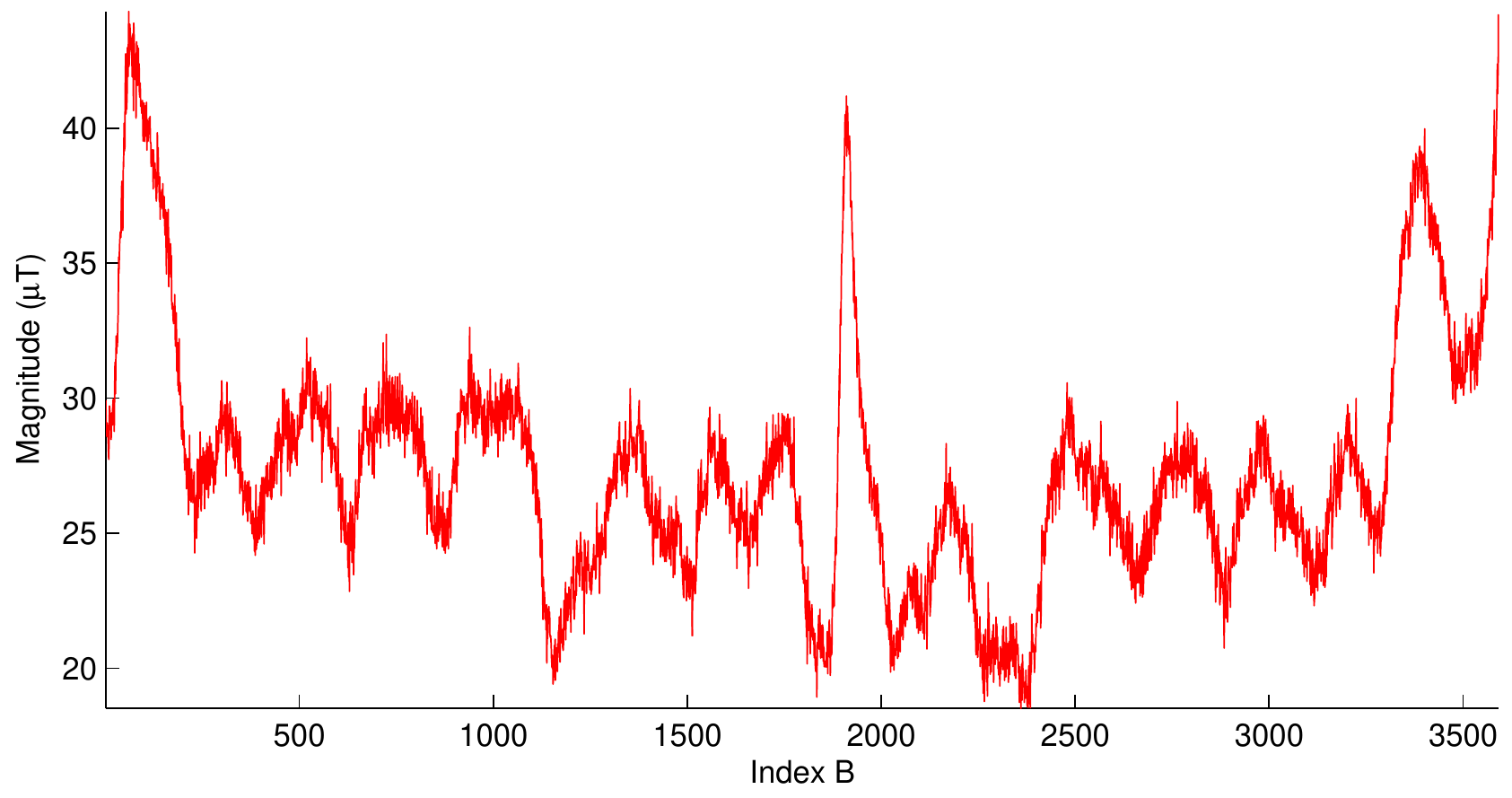}}\\
		\hline
		\rotatebox{90}{Warping Path}&	
		\subfigure[\label{fig:warp-path-sg-2}]{\includegraphics[height=3.5cm,width=6cm]{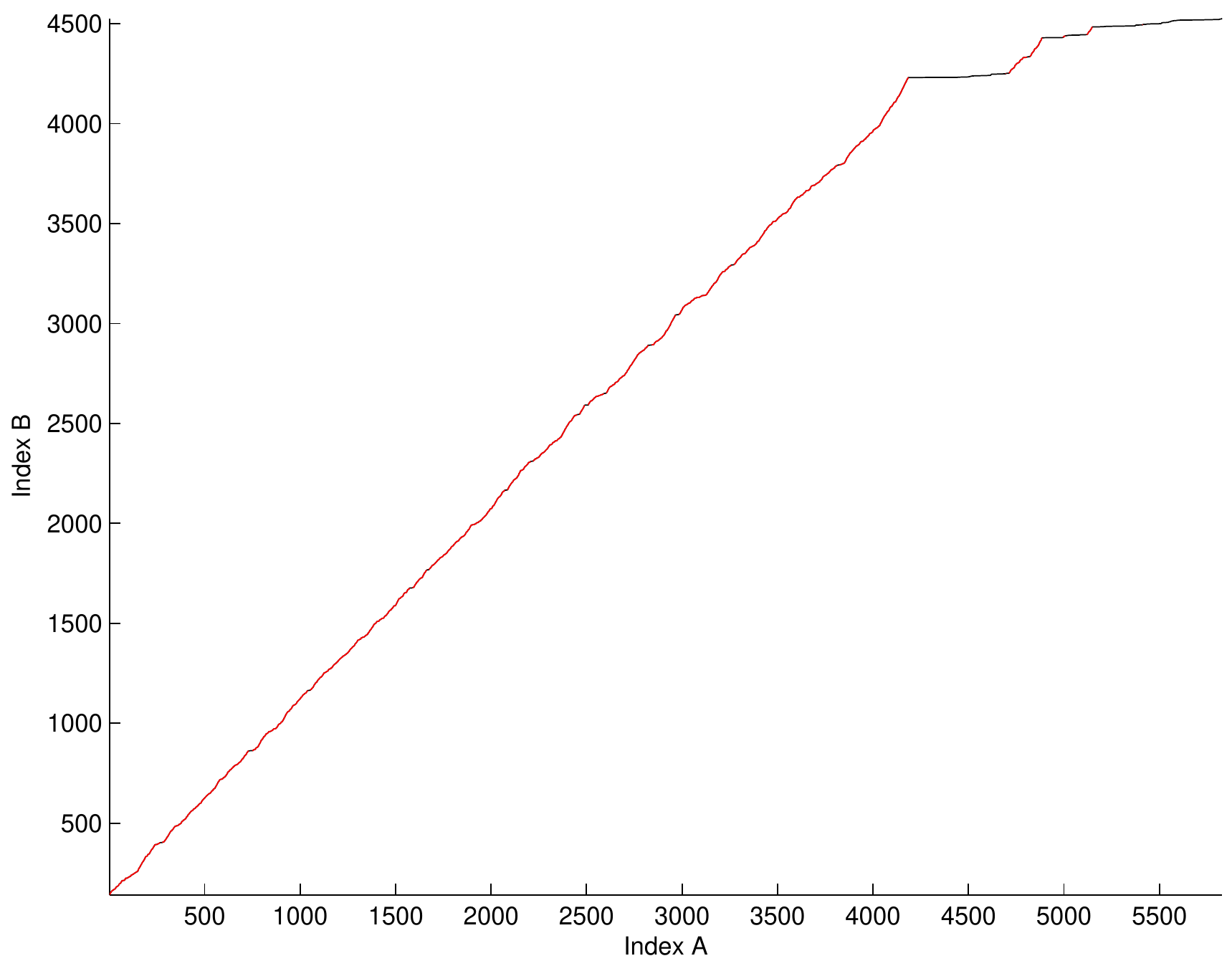}}&
		\subfigure[\label{fig:warp-path-sg-37}]{\includegraphics[height=3.5cm,width=6cm]{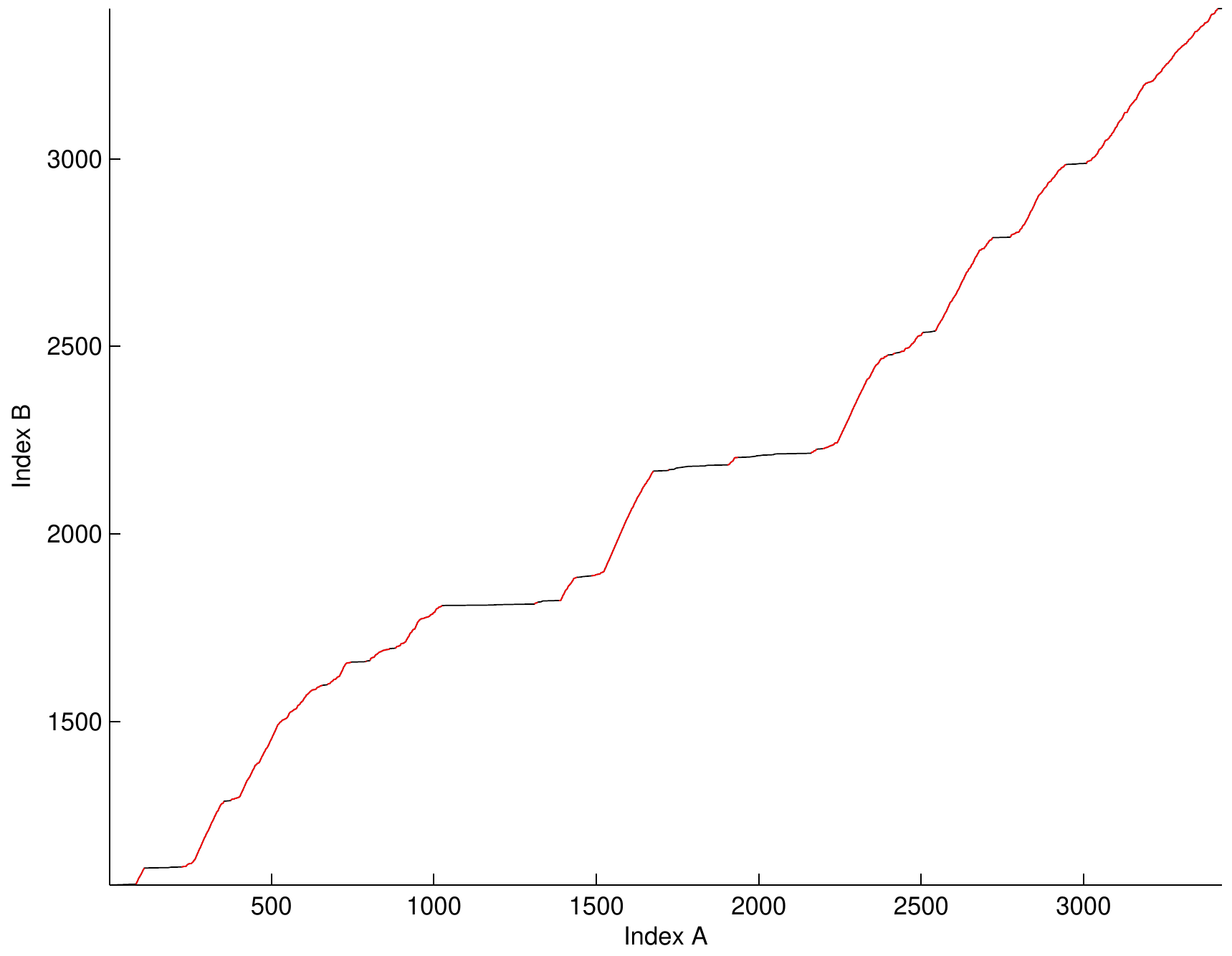}}\\
		\hline
		\rotatebox{90}{Magnetic Loop Closures\hspace{0.3cm}}&
		\subfigure[\label{fig:lc-mag-sg-2}]{\includegraphics[width=3cm,height=7cm,angle=90]{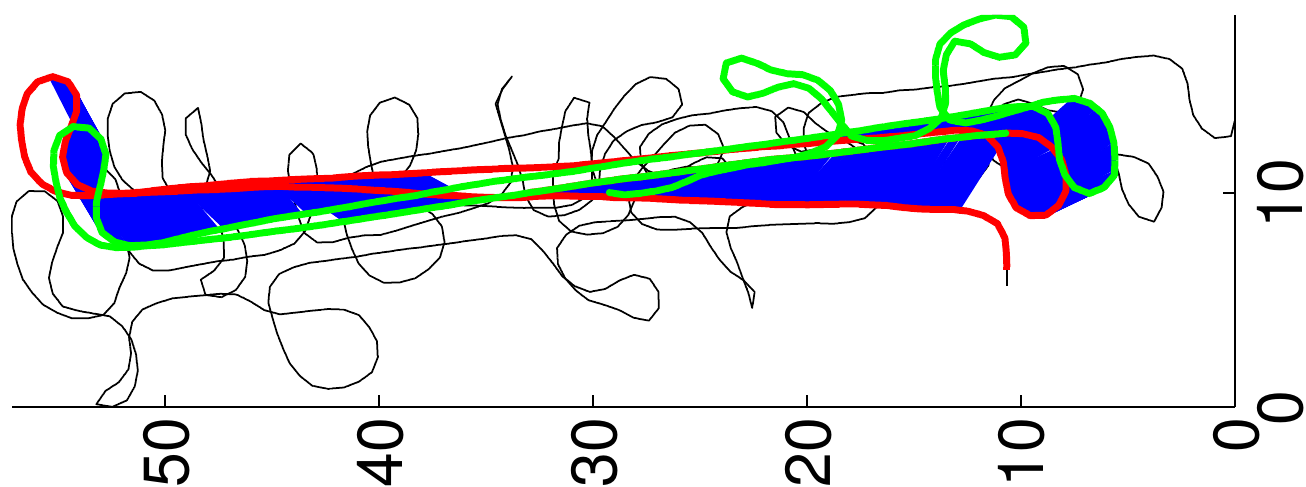}}&
		\subfigure[\label{fig:lc-mag-sg-37}]{\includegraphics[width=3cm,height=7cm,angle=90]{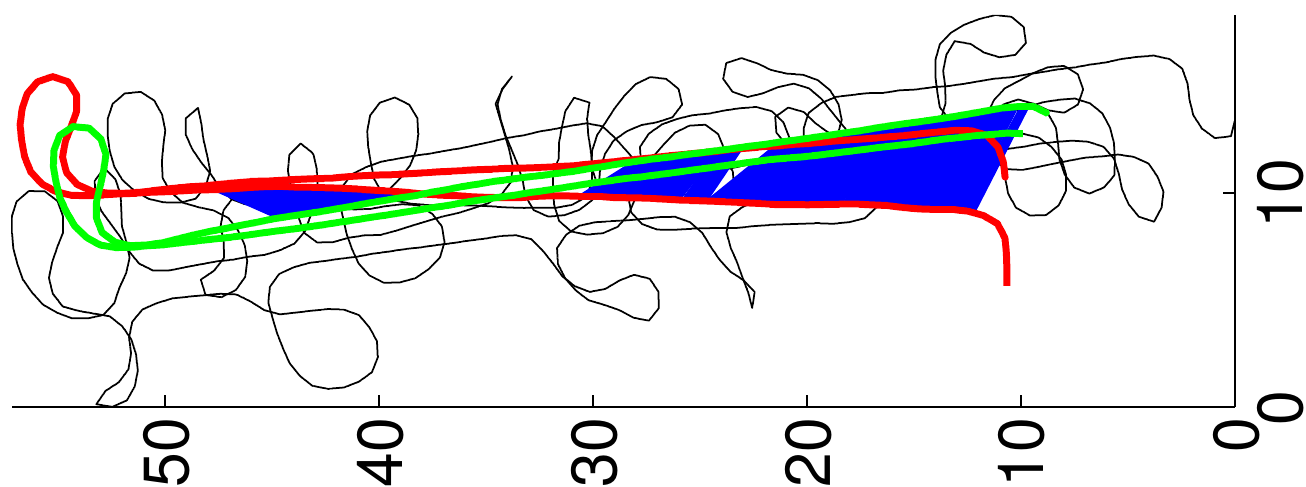}}\\
		\hline
		\rotatebox{90}{PDR Loop Closures\hspace{0.3cm}}&
		\subfigure[\label{fig:lc-hs-sg-2}]{\includegraphics[width=3cm,height=7cm,angle=90]{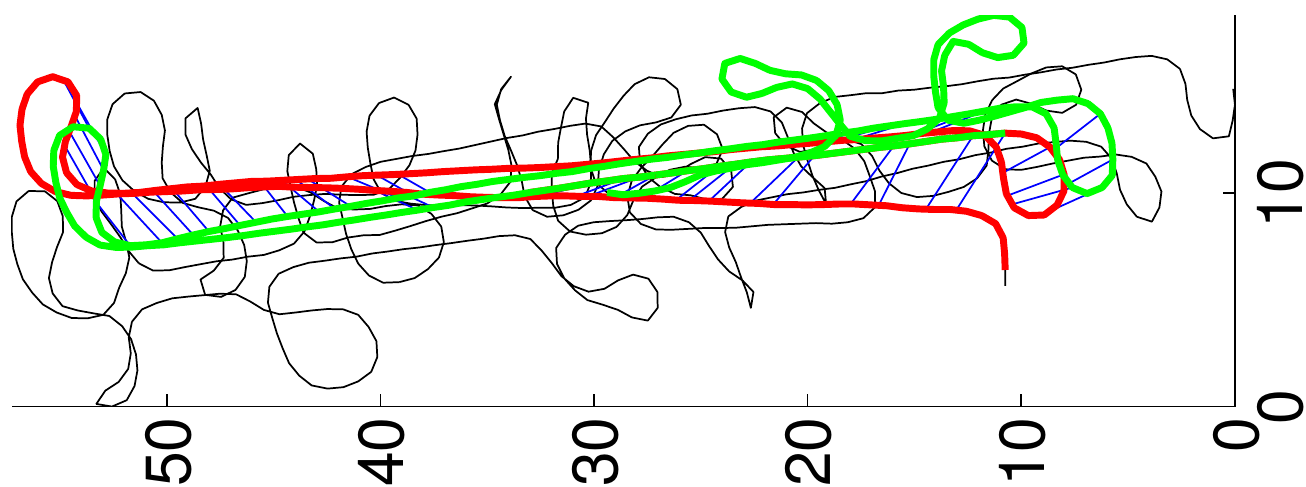}}&
		\subfigure[\label{fig:lc-hs-sg-37}]{\includegraphics[width=3cm,height=7cm,angle=90]{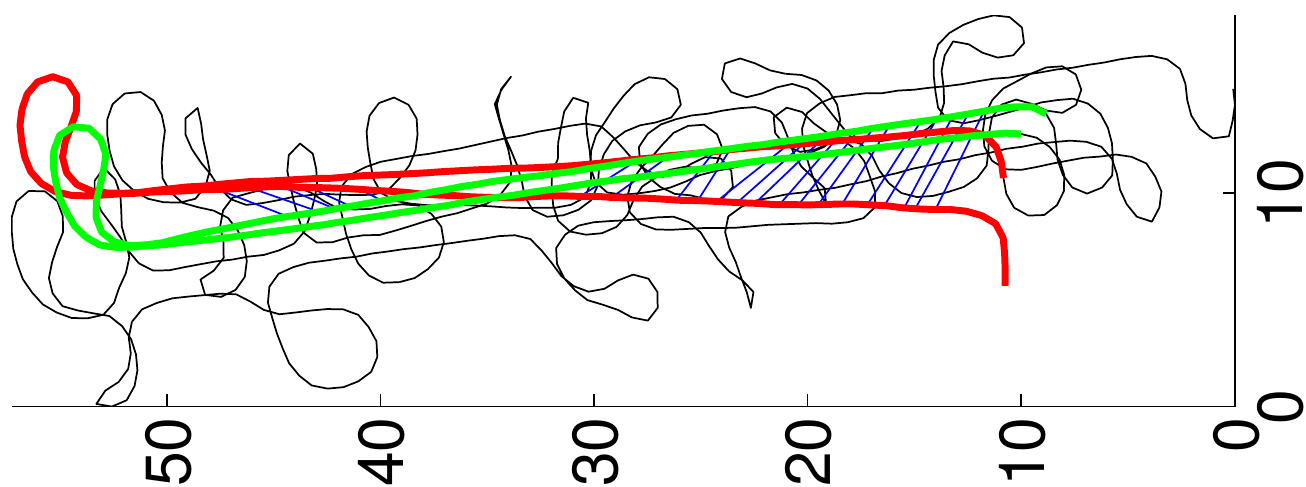}} \\
		\hline
	\end{tabular}
	\caption[Loop closure detection/validation examples for two maximum segment pairs.]{Loop closure detection/validation examples for two maximum segment pairs in Figure \ref{fig:sg-examples}.}
\end{figure*}

\begin{enumerate}
	\item \textbf{Generate PF1-traj} as in Section \ref{sec:pf1}.
	
	\item \textbf{Maximum Segment Pair (MSP) search \label{sec:msp}}. A segment is
          any \textit{consecutive} part of the position sequence,
          $s=\{i_s:i_e\}$, where $i_{s}$ and $i_{e}$ are the start and
          end indices in \textit{PF1-traj}. We process
          \textit{PF1-traj} to find \emph{segment pairs}, $(s_1,s_2)$,
          which are spatially close and hence candidates for being
          loop closures (we verify this latter property in the next
          step).

          Clearly an arbitrary segment could be contained within
          another (e.g $s_j=\{i_4,i_6\}$ is contained by the longer
          $s_k=\{i_4,i_9\}$). To avoid duplicating effort in subsequent
          steps we wish to find the Maximum Segment Pairs (MSPs),
          which are simply the segment pairs containing the longest
          segments possible without their elements violating the
          spatial proximity rule.

          Examples are given in Figure~\ref{fig:sg-examples}. The
          black trajectory is the output of PF1 when input with the
          PDR trajectory shown in Figure~\ref{fig:pdr-w1}. Three pairs
          are shown: one an arbitrary segment pair; and the other two
          maximum segment pairs. In each case one part of the pair is
          shown in green, the other in red. The arbitrary segment pair
          is not a maximum segment pair because it is contained by
          maximum segment pair 1.

          Our algorithm for finding the MSPs is described briefly here
          for the situation illustrated in Figure~\ref{fig:MSPs},
          which shows a short 9-step trajectory. We first visit each
          index of \textit{PF1-traj}, linking it with all other
          indices that lie within a distance, $R$ (shown as green
          circles for the first few steps). We then make a sequential
          pass over \textit{PF1-traj}, checking the linked indices to
          see if they are increasing (or decreasing) in sequence,
          indicating they run in parallel. In the example, we move
          from 0 to 3 and simultaneously observe 8,7,6,6,5 (we observe
          6 twice due to variance in the step length causing it to be
          in range of multiple earlier steps). Thus we find an MSP
          $\{\{0:3\},\{8:5\}\}$.

          \begin{figure}[!htb]
          \begin{center}
            \includegraphics[width=5cm]{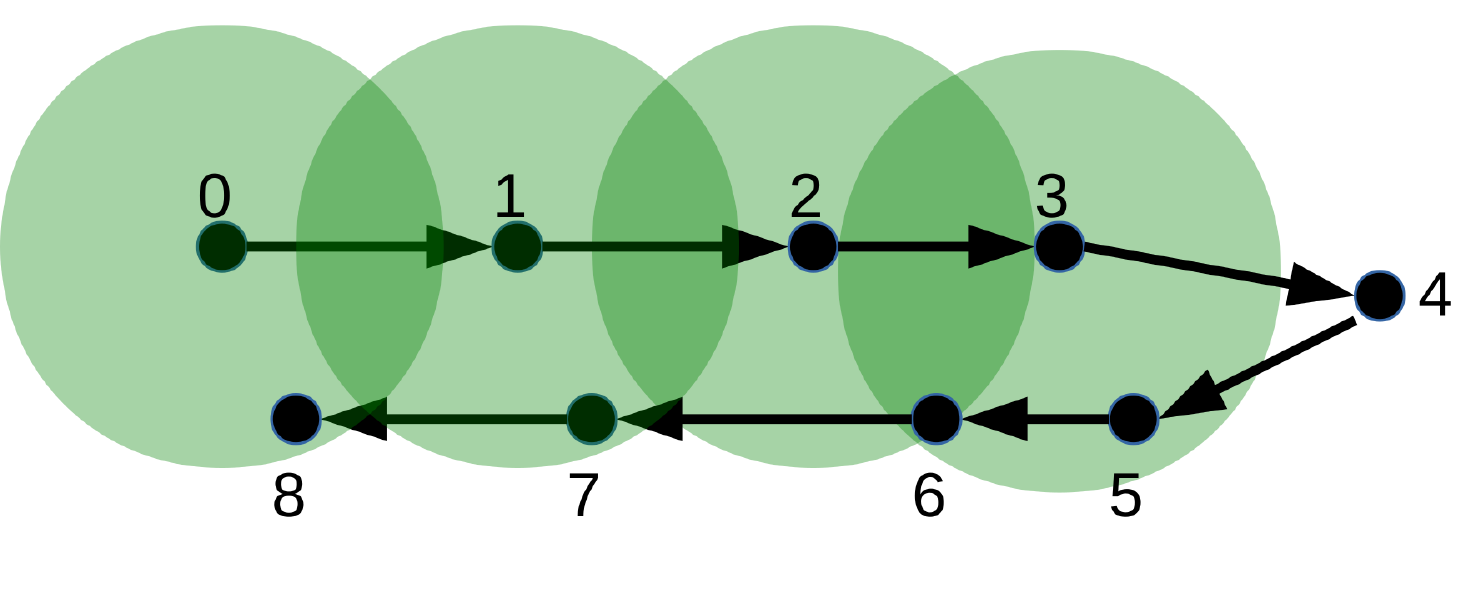}
            \end{center}
          \caption{Determining MSPs}
          \label{fig:MSPs}
          \end{figure}
          
          To set the value for $R$, please note that the drifts in
          \textit{PF1-traj} have been bounded by the floor
          plan. Because the average width of a room/corridor on the
          floor plan used here is about $2\sim3$ metres, the drifts in
          \textit{PF1-traj} are bounded within this range. Therefore
          we conservatively set $R$ to 3 metres. This value was found
          to work well for all the cases tested.
	
	\item \textbf{Sequence-based magnetic loop closure detection}. Having found
	an MSP, we must then determine the loop closures it contains. 
	We
	do so using the magnetic signal strength. We associate each
	magnetic measurement with a PF1-traj position using time
	interpolation.
	
	Figure \ref{fig:magnitude-a-sg-2} and
        \ref{fig:magnitude-b-sg-2}, Figure \ref{fig:magnitude-a-sg-37}
        and \ref{fig:magnitude-b-sg-37} are typical examples of
        magnetic signals in MSPs. The similarities between the
        magnetic sequences (waveforms) can be seen clearly from these
        figures. The intuition is that if we can match parts of
        Segment A to Segment B, we can assert loop closures between
        physical points on these two path segments. We use Open-Begin-End
        Dynamic Time Warping (OBE-DTW) with an `asymmetric' step
        pattern~\cite{giorgino2009computing} to get point-to-point
        correspondences (loop closures) between segments. OBE-DTW compresses or stretches the time
        series to create a `warping path' between the segments. A
        point $(i,j)$ on the warping path means the $i^{th}$ element
        of $M_1$ matches to the $j^{th}$ element of $M_2$ (Figure
        \ref{fig:warp-path-sg-2} and
        Figure~\ref{fig:warp-path-sg-37}). \footnote{Given two time
          sequences, OBE-DTW would return the best alignment and the
          \emph{DTW-distance} which measures how good this matching is
          even if the two sequences do not contain any true loop
          closures~\cite{giorgino2009computing} . A possible
          validation method is to impose a threshold of this
          DTW-distance to reject false positive alignments. But we
          found that it is hard to find a good global threshold for
          the DTW-distance given the noise of the magnetometer on
          modern smartphone. Therefore we do not use a threshold on
          the DTW-distance, preferring instead to filter false
          closures at the next stage.} A horizontal segment in the
        warping path means DTW has either stretched one of the signals
        to fit (accounting for a speed difference) or mapped a chunk
        of the segment to one value on the the other segment since
        that chunk does not match anything. The latter situation leads
        to false positive closures.
	
	We therefore filter the warping path, splitting the segments into
	sub-segments at each horizontal part of the warping path. For
	example, several sub-segments are created from Figure
	\ref{fig:warp-path-sg-2} and
	Figure~\ref{fig:warp-path-sg-37} respectively (all shown in red). The
	sub-segments are carried forward as possible matchings (sequences
	of loop closures).
	
	\item \textbf{Closure validation}. The closure detection algorithm
	produces a large number of potential closures based solely on the
	magnetic observations. We apply a series of spatial criteria to
	reject matched subsequences, $N$ and $M$, that we are not
	confident in. The criteria are based on empirical constants chosen
	to be aggressive in culling closures---we would rather have a few
	true positive closures than a lot of true positives mixed with
	false positives. As such the constant values are not particularly
	sensitive.
	
	\begin{itemize}
		\item
		\textit{Either $length(N)$ or $length(M)$ must be larger than
			2.5}~m (about 3$\sim$4 human steps). We expect that at
		least 3$\sim$4 steps are contained in a valid loop closure
		sequence.
		\item
		Assume $length(N) > length(M)$, then \textit{the
			$\frac{length(N)}{length(M)}$ must be less than 2.0}. This
		is to ensure the magnetic time series are not over compressed
		(stretched) because speed changes are not expected to be
		great.
		\item
		\textit{$M$ and $N$ must be physically close}. Because PF1 has corrected the trajectory to within
		some scaling errors we do not trust any instances where
		$M$ and $N$ are not physically close. To assess this we
		compute the mean spatial distance between the matched points
		of $M$ and $N$ on PF1-traj, $D_{mean}$. We require this
		value empirically to within $3.0$~m, which is the upper limit of the drifts in \textit{PF1-traj} as described earlier in this section.
		
		\item		
		Additionally we expect the shapes of the two segments to be
		similar. To capture this we use the variance of the distances between
		the matched points, $D_{var}$. We set this value empirically to
		lie within 1.0~m$^2$. 
		
	\end{itemize}
	
      \item \textbf{Closure step mapping}. We now have loop closures
        between magnetic measurement points on the survey path
        (Figures \ref{fig:lc-mag-sg-2} and
        \ref{fig:lc-mag-sg-37}). The closures are very dense due to
        the high sampling rate of the magnetic signal. However, we
        only need closures at matching points in the gait cycle. This
        thinning can be done using time interpolation for the part of
        path segments that covered by the dense magnetic loop
        closures---e.g. Figure \ref{fig:lc-hs-sg-2} and
        \ref{fig:lc-hs-sg-37}. Now the loop closures only connect the
        end points of human steps to corresponding points on the
        survey path (we call them the \emph{PDR loop closures}), which
        are far sparser than the original magnetic loop closures.
		
\end{enumerate}

\subsection{PF2}

The previous stages can be seen as pre-processing for this major pass
with a particle filter. The filter is an extended version of PF1 where
we seek higher accuracy output. To that end we adapt the KLD
resampling parameters of PF1 such that $\Delta_x = \Delta_y = 0.5~m$,
and $\Delta_\theta = 1^\circ$. The new value of $n_{min}$ is then
16433. PF2 must incorporate the straight line and loop closure constraints at
the particle reweighting stage. The full reweighting procedure for
particle $p_{i,t}$ incorporating step $s_t$ is:

\begin{enumerate}
	\item
	Initialise the particle weight to 1: $w_{i,t}=1$
	\item
	If the step crosses a wall, set $w_{i,t}=0$ and return.
	
	\item
	If $s_t$ is marked as a straight-line step, find the
	room/area wall that is closest to parallel to the step
	direction. We model the acute angle $\alpha$ between this
	wall and the step vector as a random variable drawn from a
	folded normal distribution with mean $0$ and variance
	$\sigma_{\alpha}^2$. We then multiply the particle weight by the
	probability of the measured $\alpha$:
	
	$$
	w_{i,t}=w_{i,t}\cdot Trunc\mathcal{N}(\alpha~|~0, \sigma_{\alpha}^2) = w_{i,t}\cdot \frac{\sqrt{2}}{\sigma_{\alpha} \sqrt{{\pi}}}e^{-\frac{\alpha^2}{2\sigma_{\alpha}^{2}}}
	$$
	
	where $\sigma_a$ is set to 2.5$^{\circ}$, which is half of the turning angle threshold for straight line detection.

	\item
	If $s_t$ is associated with a loop closure we compute the
	Euclidean distance $d$ between the new particle position and
	the other point specified by this loop closure. We model
	this distance using a folded normal distribution with
	mean $0$ and variance $\sigma_{d}^2$. We multiply the
	weight by the probability of the observed distance:
	
	$$ w_{i,t} =w_{i,t}\cdot Trunc\mathcal{N}(d~|~0,
	\sigma_{d}^2) = w_{i,t}\cdot\frac{\sqrt{2}}{\sigma_{d}
		\sqrt{{\pi}}}e^{-\frac{d^2}{2\sigma_{d}^{2}}}
	$$
	
	Here, a very large $\sigma_{d}$ causes the weight distribution
        to be very flat over particles and a very small $\sigma_{d}$
        gives most particles a weight of almost 0. We found 1.0~m (and
        nearby values) gives a smooth weight distribution over
        particles.
	
\end{enumerate}

Since the loop closures remove the room ambiguities, the PF2 output
does not require a complicated smoother. We use the simple
pruning approach.  More advanced smoothers can be applied,
although at significant additional cost for marginal or no gain in our
experience. With this approach PF2 typically finishes within 2.5
minutes for a 10-minute survey walk (about 950 steps).

\section{Evaluation}

The data used to demonstrate and evaluate this work were collected in
the William Gates Building, a three-storey office building at the
University of Cambridge, UK. Data collection was done using a consumer
Android smartphone that logged WiFi scan results, along with
the accelerometer, gyroscope and magnetometer sensor values. A variety
of path surveys were carried out in the building. Because all the survey paths were taken following the best-practice guidelines described in Section~\ref{sec:dedicated-surveyor}, the randomness in human behaviour was lowered, i.e., survey paths meeting these guideline are very similar. Therefore we only selected three typical ones and used
them throughout this work: \emph{Path-1} covered a single corridor
where high accuracy ground truth location (Figure
\ref{fig:example-path-survey}) obtained using the Bat positioning system (capable of 3D positioning to an accuracy of 3~cm 95\% of the time with a 10--15Hz update rate)~\cite{Addlesee01} was available while
\emph{Path-2} and \emph{Path-3} cover the entire floor but with only
coarse ground truth available (we used PDR and manual checkpoints to achieve nearly meter-accuracy, the sequence of rooms visited and the
actions performed in each were also recorded).

\subsection{Straight Line Detection \label{sec:sline-detect}}

\begin{figure*}
	\centering
	\begin{tabular}{ccc}
		\subfigure[\emph{Path-1}]{\includegraphics[height=4cm]{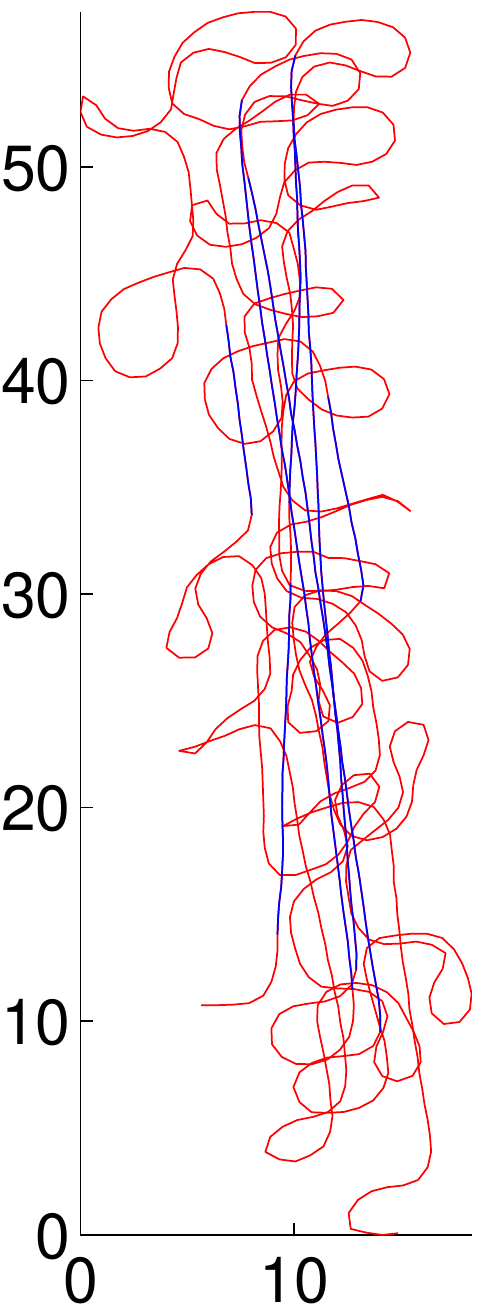}}
		&
		\subfigure[\emph{Path-2}]{\includegraphics[height=4cm]{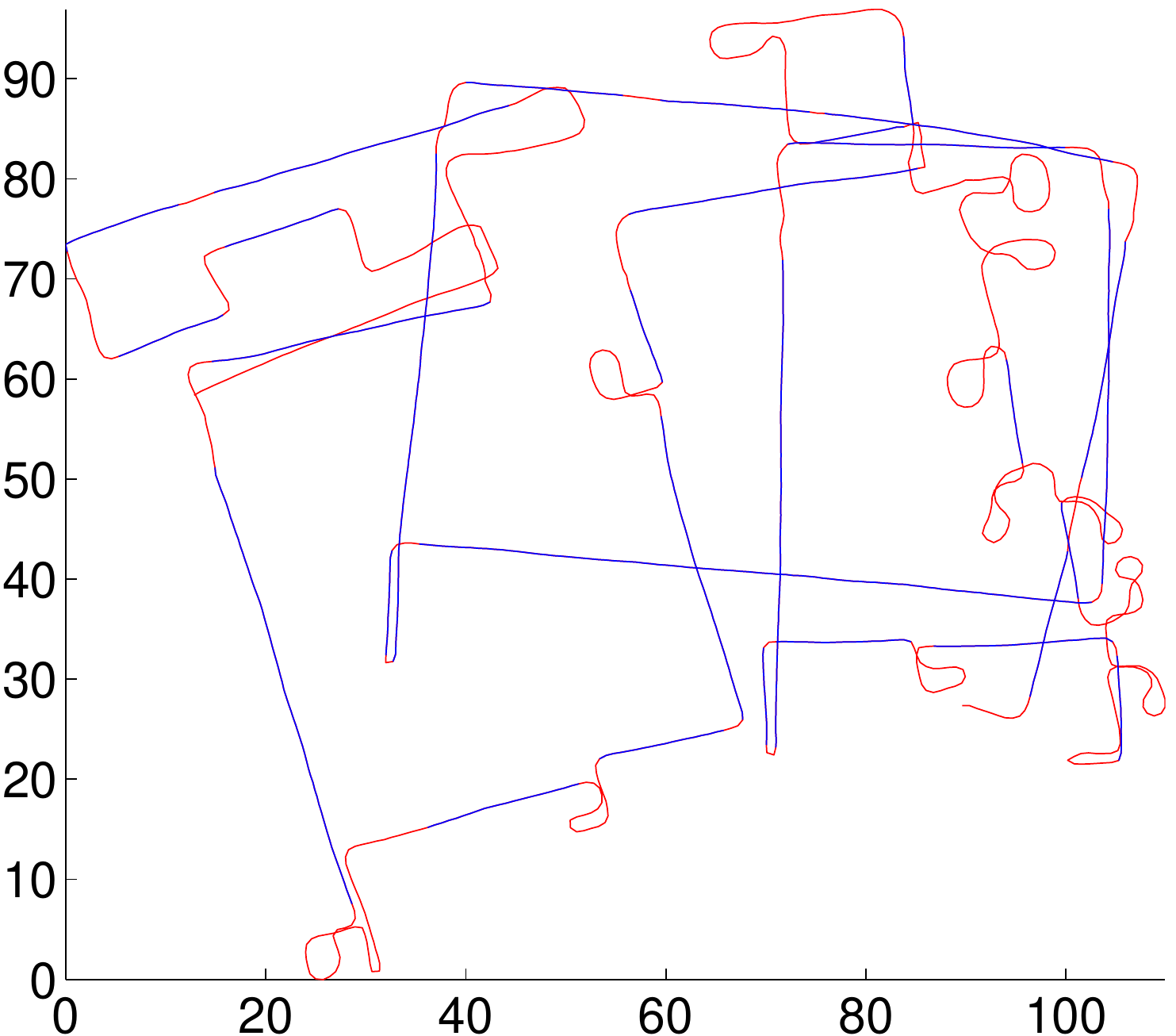}}
		&
		\subfigure[\emph{Path-3}]{\includegraphics[height=4cm]{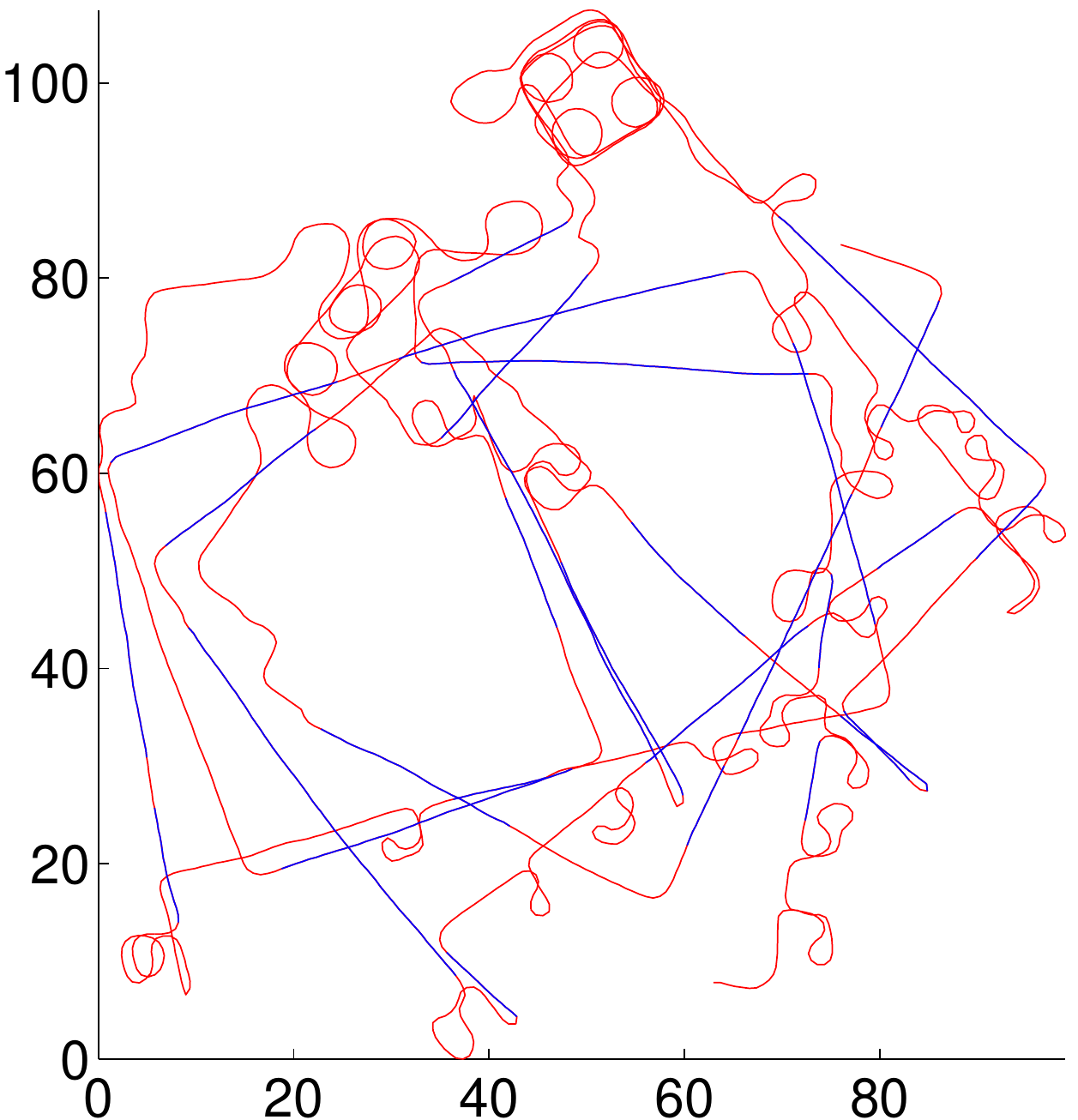}}\\

	\end{tabular}
	\caption{Straight line detection. The red lines indicate the PDR result; blue lines indicate straight-line segments identified by the straight line filter.}
	\label{fig:straight-line-results}
\end{figure*}

We show the straight line detection results in
Figure~\ref{fig:straight-line-results}. It shows that \emph{Path-2} and \emph{Path-3} have more
straight lines detected (30 and 36 respectively) than \emph{Path-1} (only seven). This is expected because \emph{Path-2} and \emph{Path-3} went through long straight corridor areas for much more times than \emph{Path-1} did. Note that the straight line
filter is not applicable for walking in office rooms and large open spaces. However, it is simply one (relatively minor) constraint
used in our system to boost accuracy (Figure~\ref{fig:accuracy-compare}), and is more useful in reducing the ambiguity in the particle cloud (by distributing fair weights to the particles).

\subsection{Loop Closure Detection and Validation\label{sec:lcd}}

\begin{figure*}
	\centering
	\begin{tabular}{l|c|c|c}
		\hline
		&\emph{Path-1}&\emph{Path-2}&\emph{Path-3}\\
		\hline
		\rotatebox{90}{\textit{Unvalidated loop closures}}&
		\subfigure[]{\includegraphics[height=4cm]{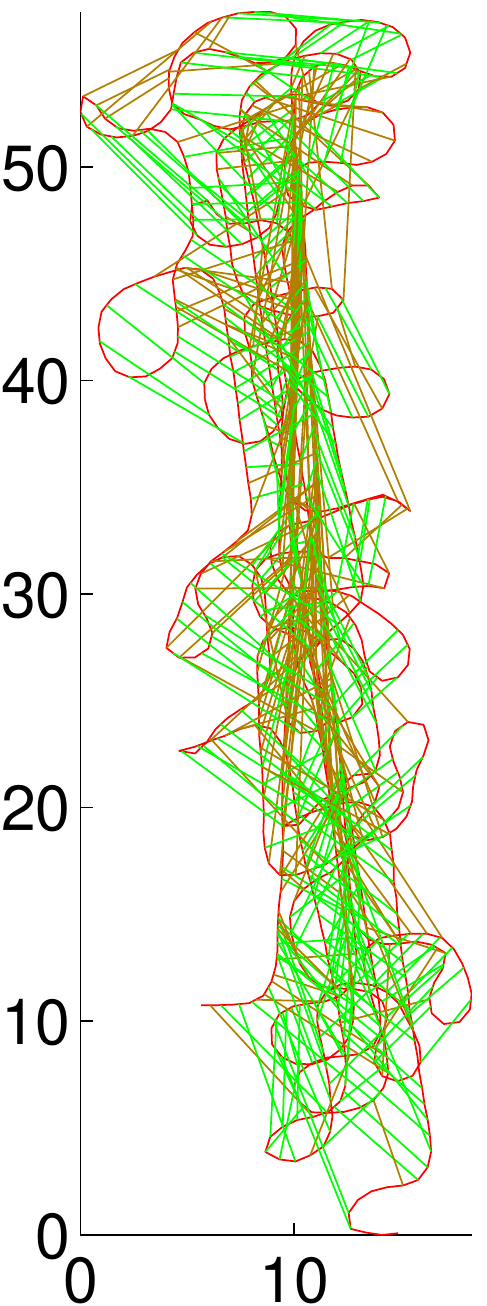}}
		&
		\subfigure[]{\includegraphics[height=4cm]{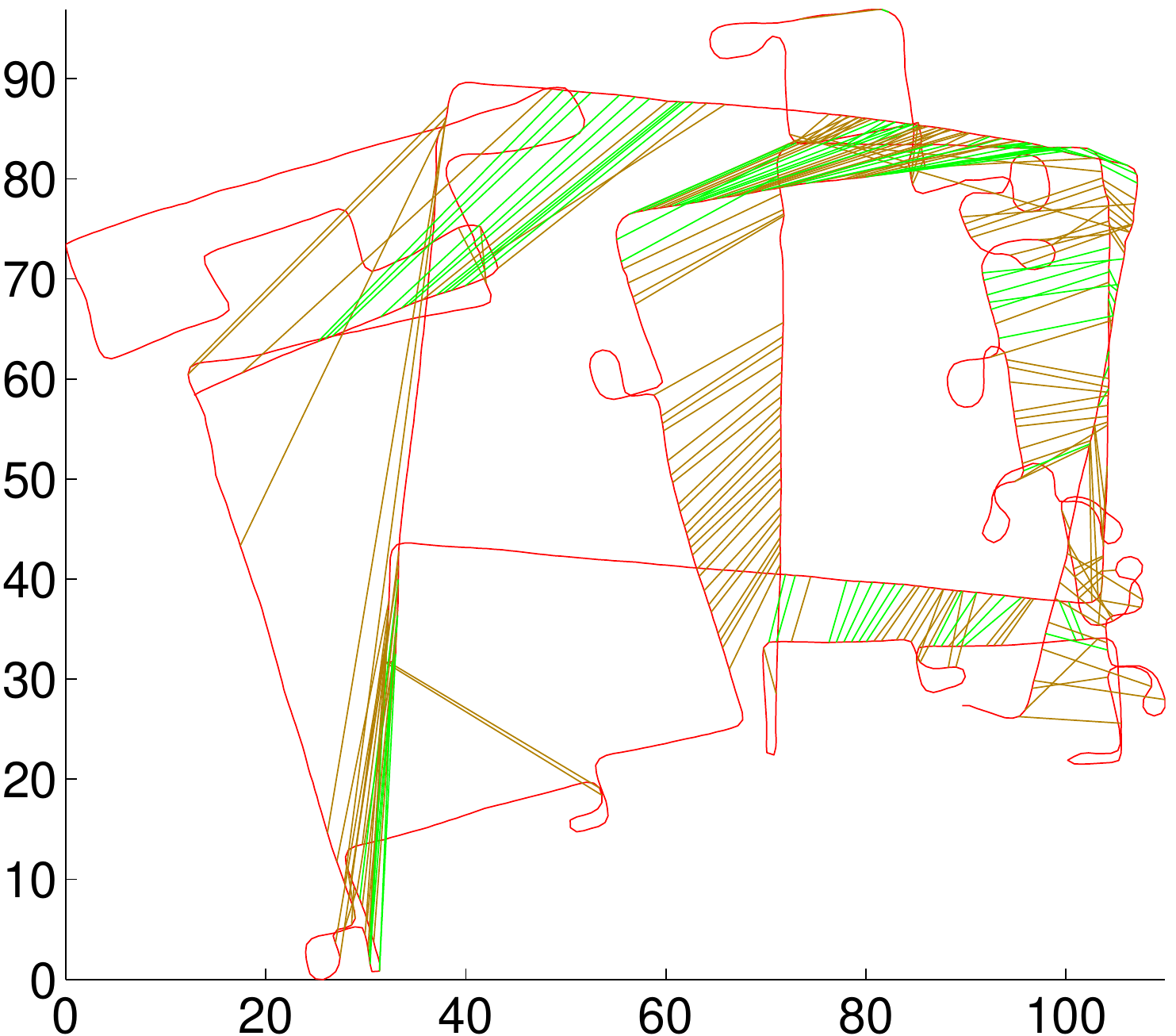}}
		&
		\subfigure[]{\includegraphics[height=4cm]{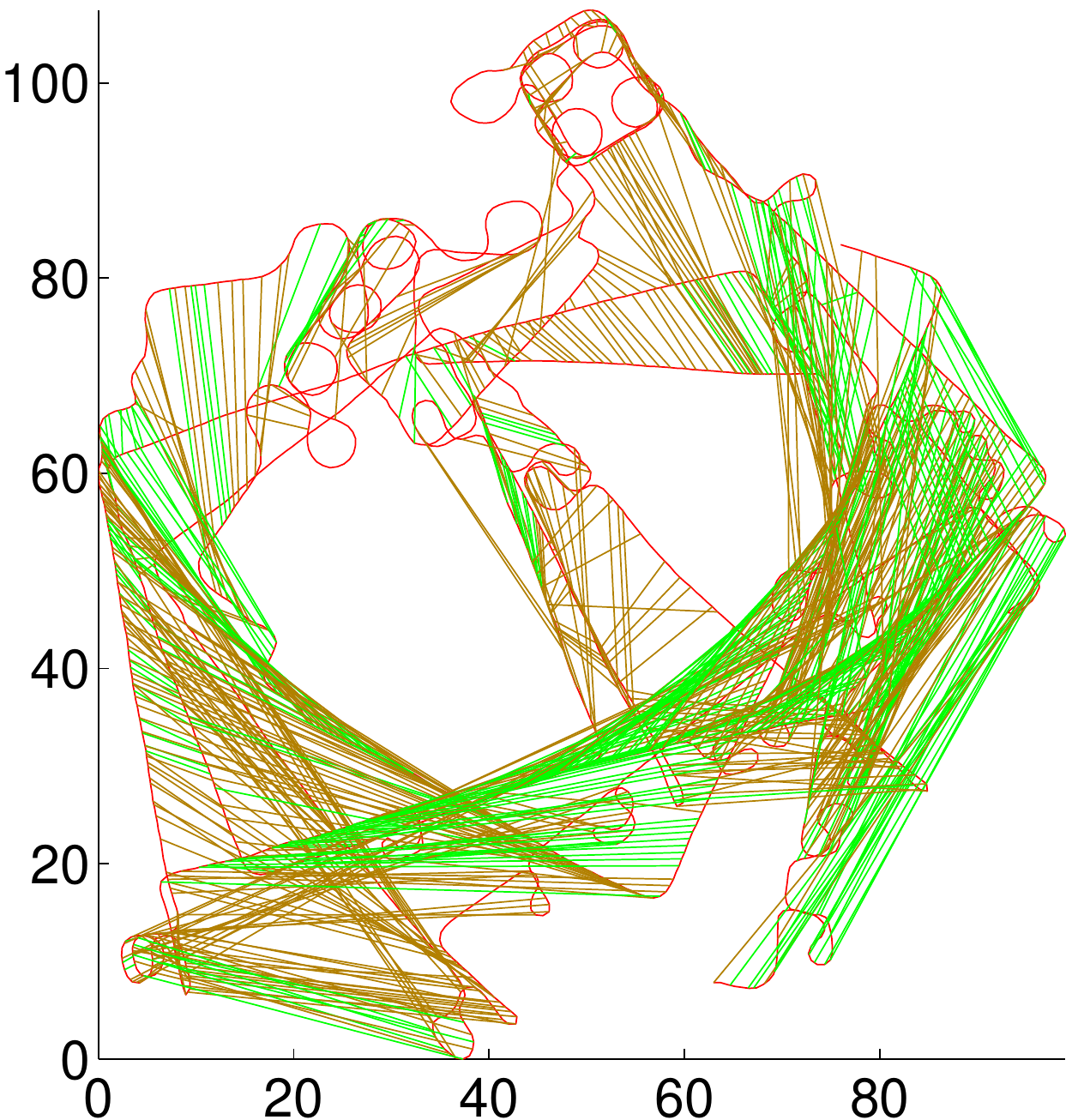}}
		\\
		\hline
		\rotatebox{90}{\textit{Validated loop closures}}&
		\subfigure[]{\includegraphics[height=4cm]{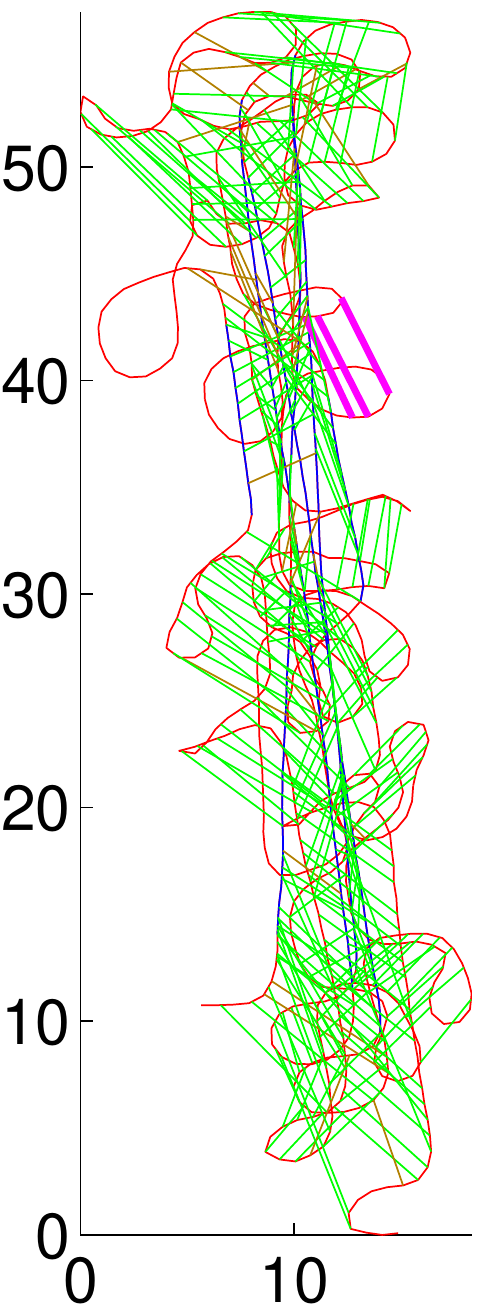}}
		&
		\subfigure[]{\includegraphics[height=4cm]{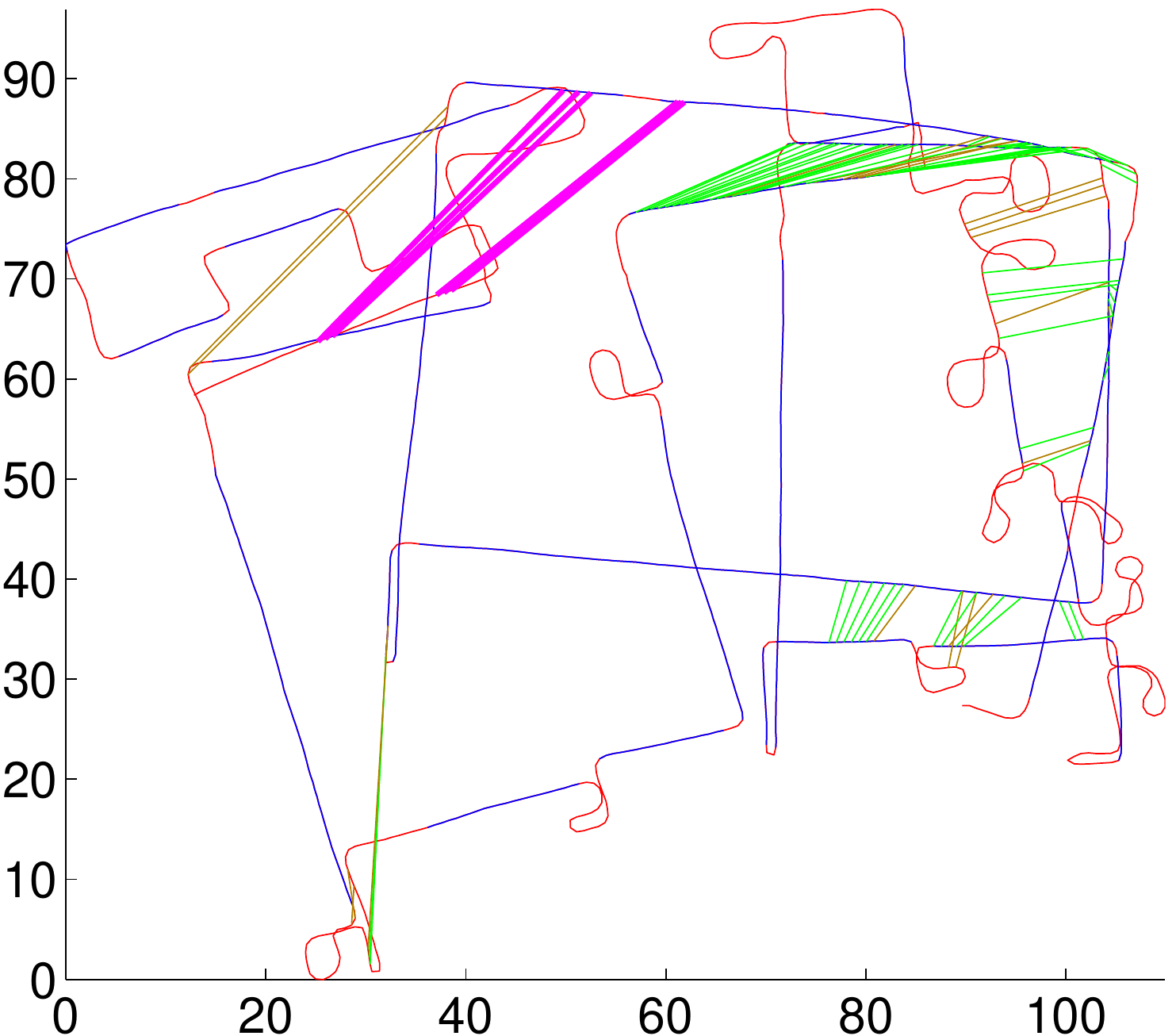}}
		&
		\subfigure[]{\includegraphics[height=4cm]{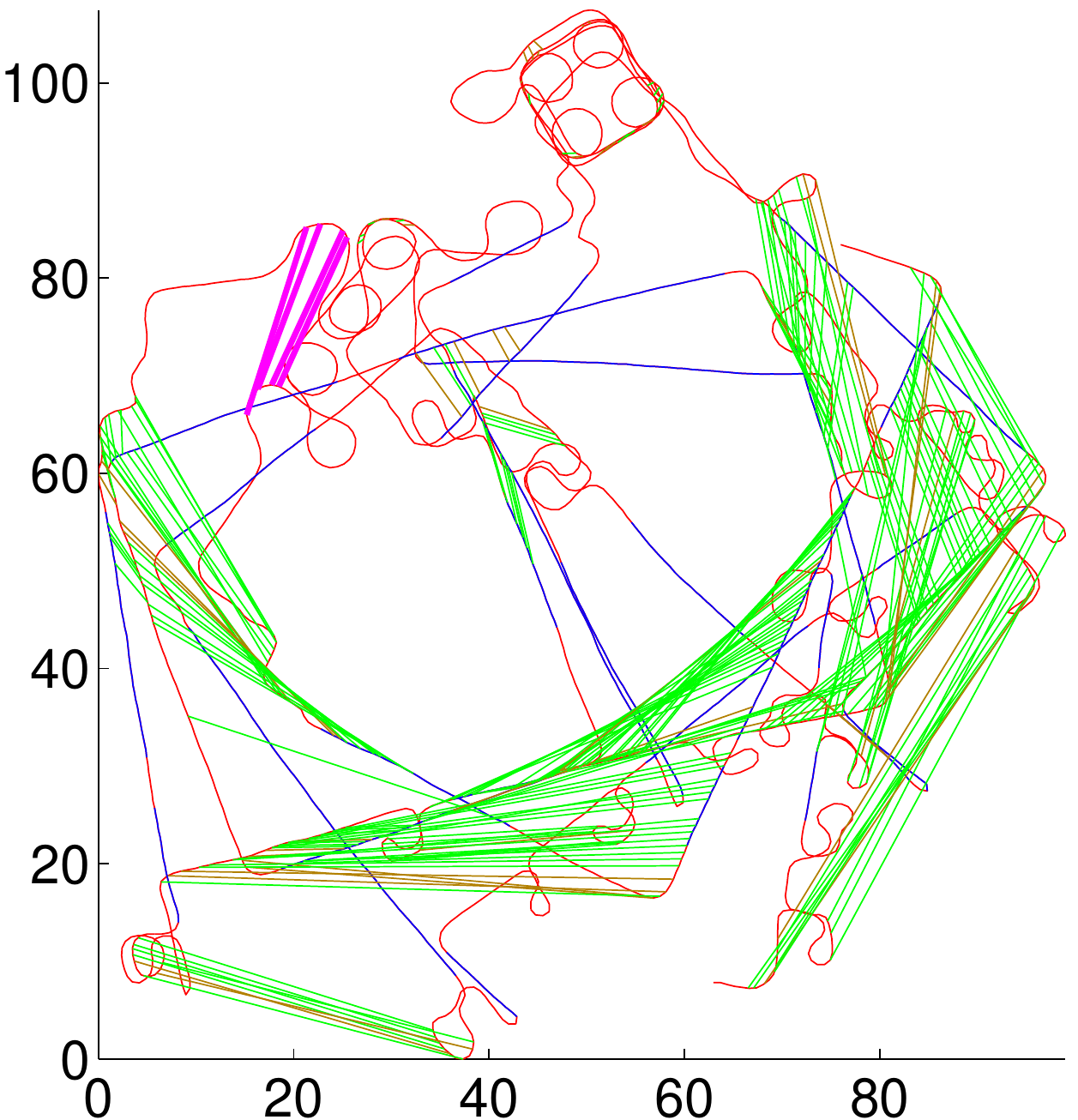}}\\

		\hline
	\end{tabular}
	\caption{Loop closure validation. The green and brown lines indicate true-positive and false positive loop closures respectively.}
	\label{fig:lc-results}
\end{figure*}

Figure~\ref{fig:lc-results} illustrates the loop closures detected
from the \emph{PF1-traj}, drawn onto the \emph{PDR-traj} (since this
represents the input to PF2). In these figures green lines indicate
true-positive loop closures; brown lines indicate false positive loop
closures. Note that the classification of a loop closure was done
post-hoc using the final trajectory---it would not be known at this
stage in a live system. The bottom row shows the loop closures
\emph{after} validation. Very few false positives remain. The
magenta-highlighted lines are true-positive closures that match to the
magenta highlighted regions in Figure~\ref{fig:pfs-results}.

\begin{table}[ht]
	
	\centering
	\begin{tabular}{|c|c|c|c|c|c|c|}
		\hline
		&\multicolumn{2}{c|}{\textit{Path-1}} & \multicolumn{2}{c|}{\textit{Path-2}}&\multicolumn{2}{c|}{\textit{Path-3}} \\
		\hline
		&Before&After&Before&After&Before&After \\
		\hline
		True&304&284&84&68&280&196 \\
		False&217&37&158&21&565&53 \\
		Ratio&0.58&0.88&0.35&0.76&0.33&0.79 \\
		\hline
	\end{tabular}
	\caption{Statistics on loop closures before and after validation.\label{tab:loop-closure-filtering-stats}}
\end{table}

Table \ref{tab:loop-closure-filtering-stats} summarises the statistics
on loop closures for the three paths. We observe that the validation
algorithm was highly conservative as intended: it rejected a large
number of closures. In all cases it significantly boosted the percentage of
true loop closures above 0.7 as intended. False
positives were thus moved to a minority and could not adversely impact
the results (which are shown in the next Section~\ref{sec:results}).

\subsection{Trajectory Outputs, Particle Clouds and Room Ambiguities\label{sec:results}}

\begin{figure*}
	\centering
	\begin{tabular}{c|c|c|c}
		\hline
		&\emph{Path-1}&\emph{Path-2}&\emph{Path-3}\\
		\hline
		\rotatebox{90}{\textit{Trajectory}}
		&
		\subfigure[]{\includegraphics[height=4cm]{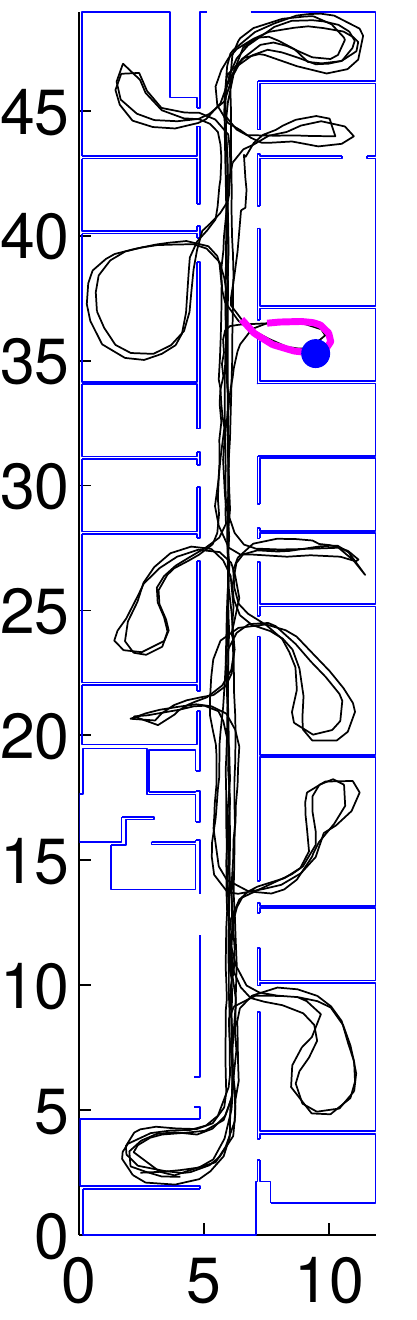}}
		&
		\subfigure[]{\includegraphics[height=4cm]{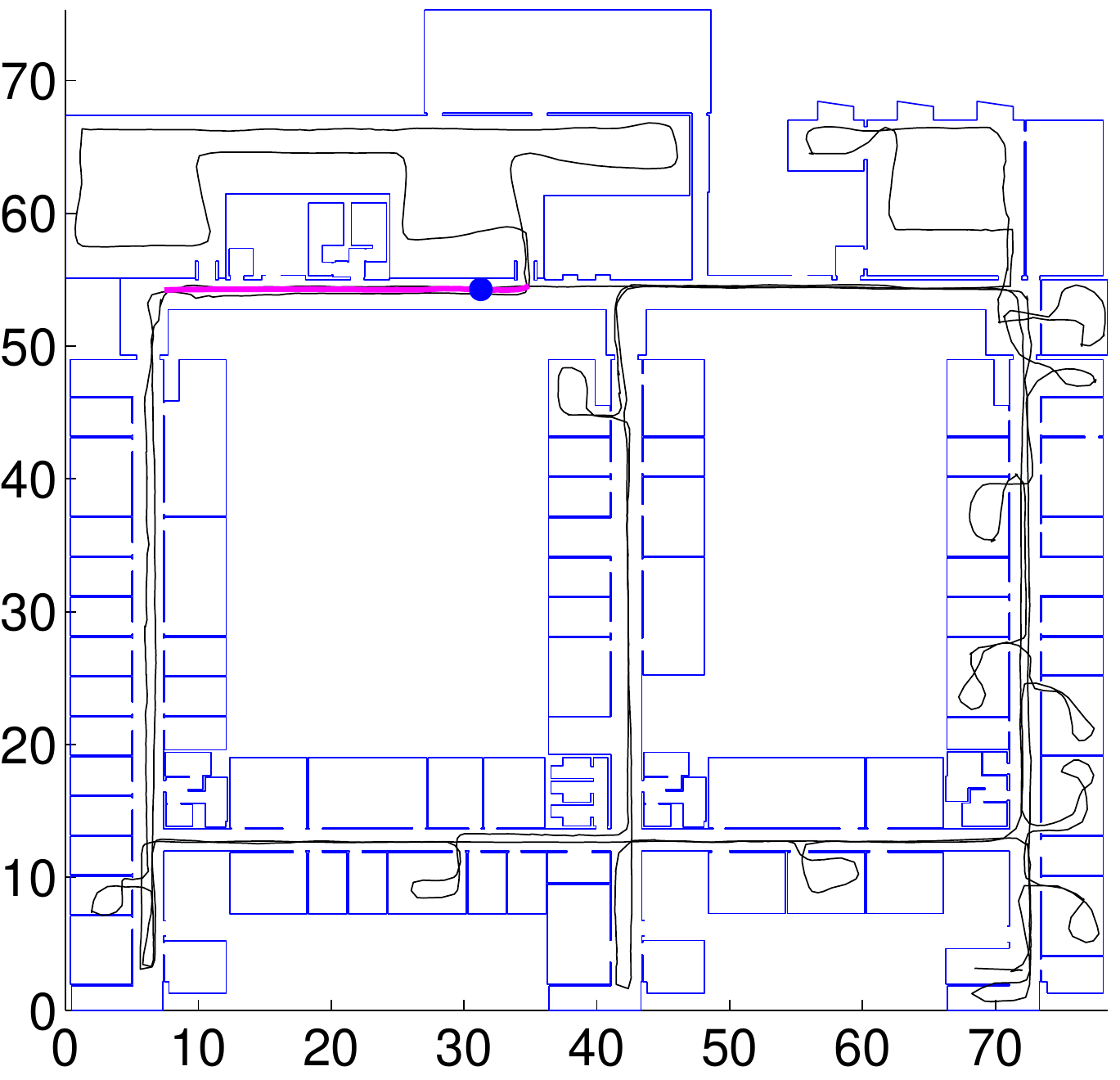}}
		&
		\subfigure[\label{fig:pfs-results-c}]{\includegraphics[height=4cm]{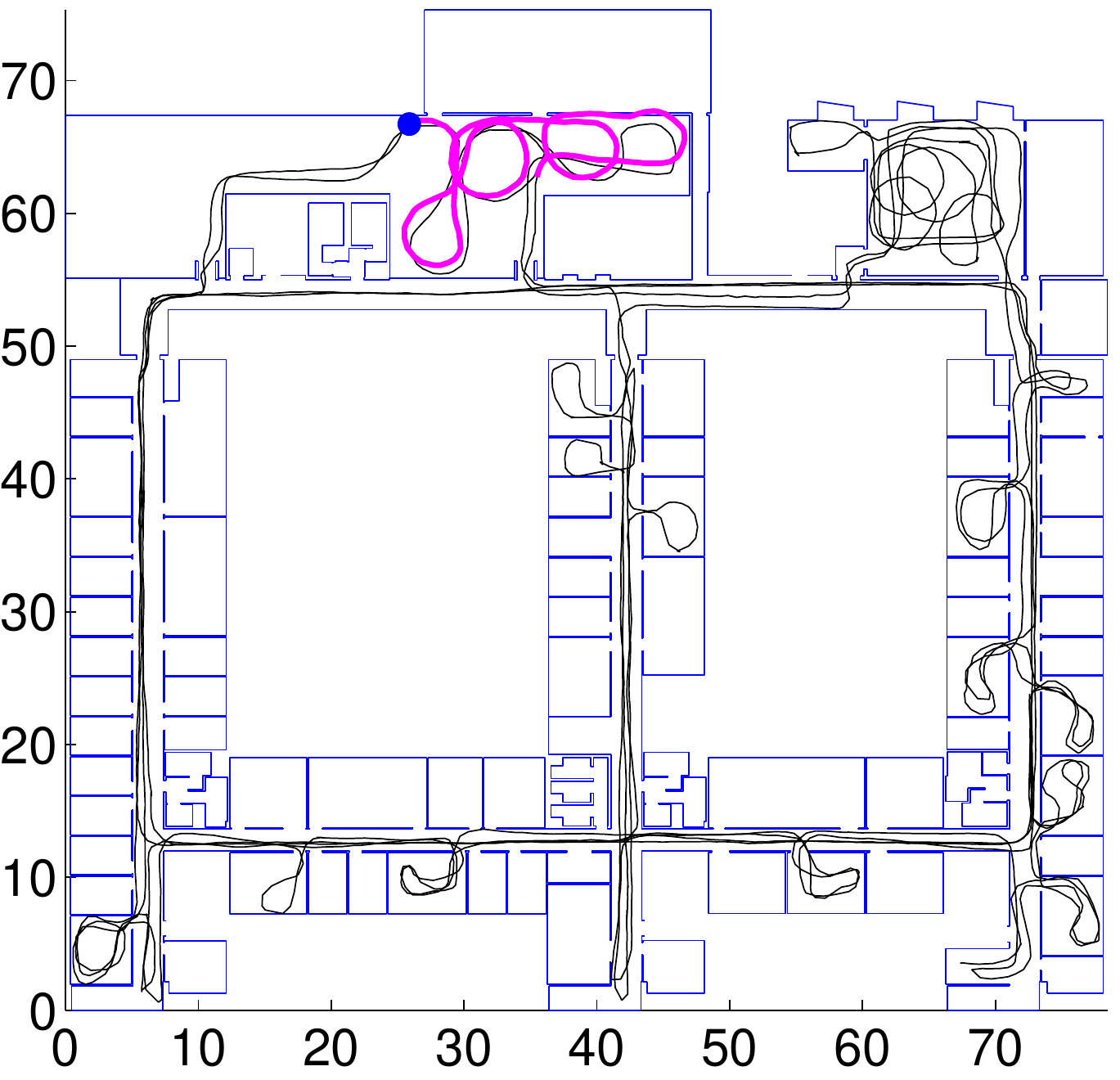}}
		
		\\
		
		\hline 
		\rotatebox{90}{\textit{Particle cloud}}&
		\subfigure[\label{fig:pcloud-w1-pfs-pdr}]{\includegraphics[height=4cm]{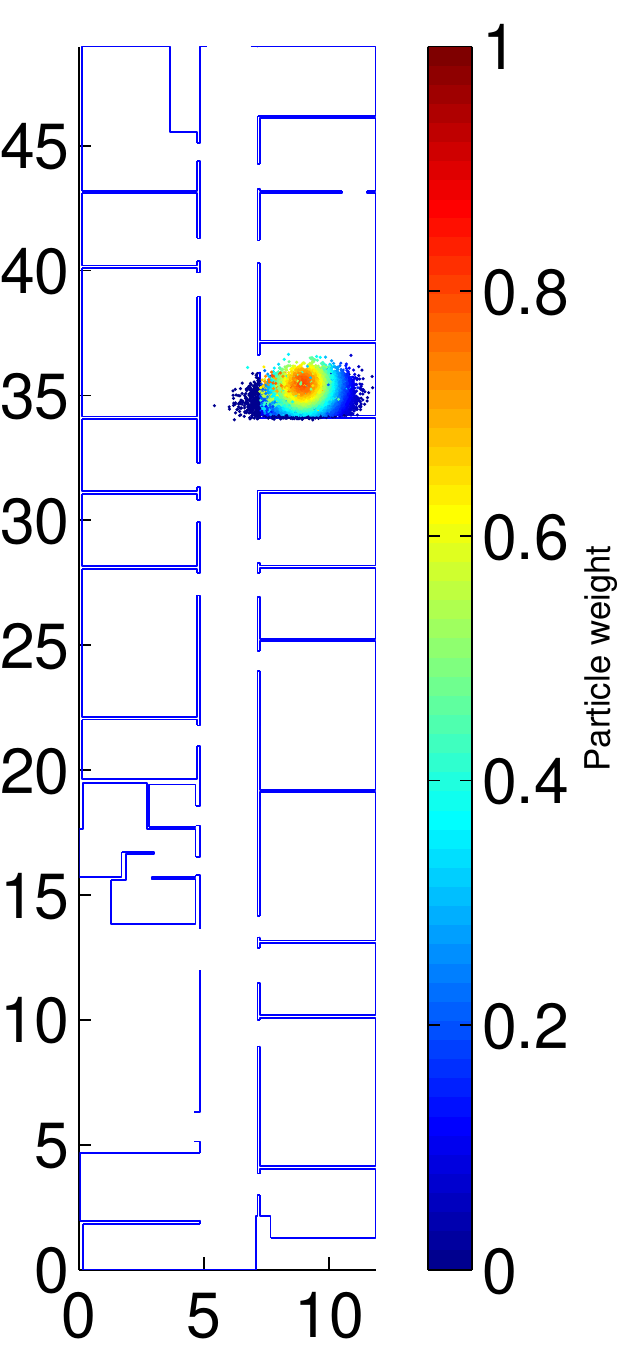}}
		&
		\subfigure[]{\includegraphics[height=4cm]{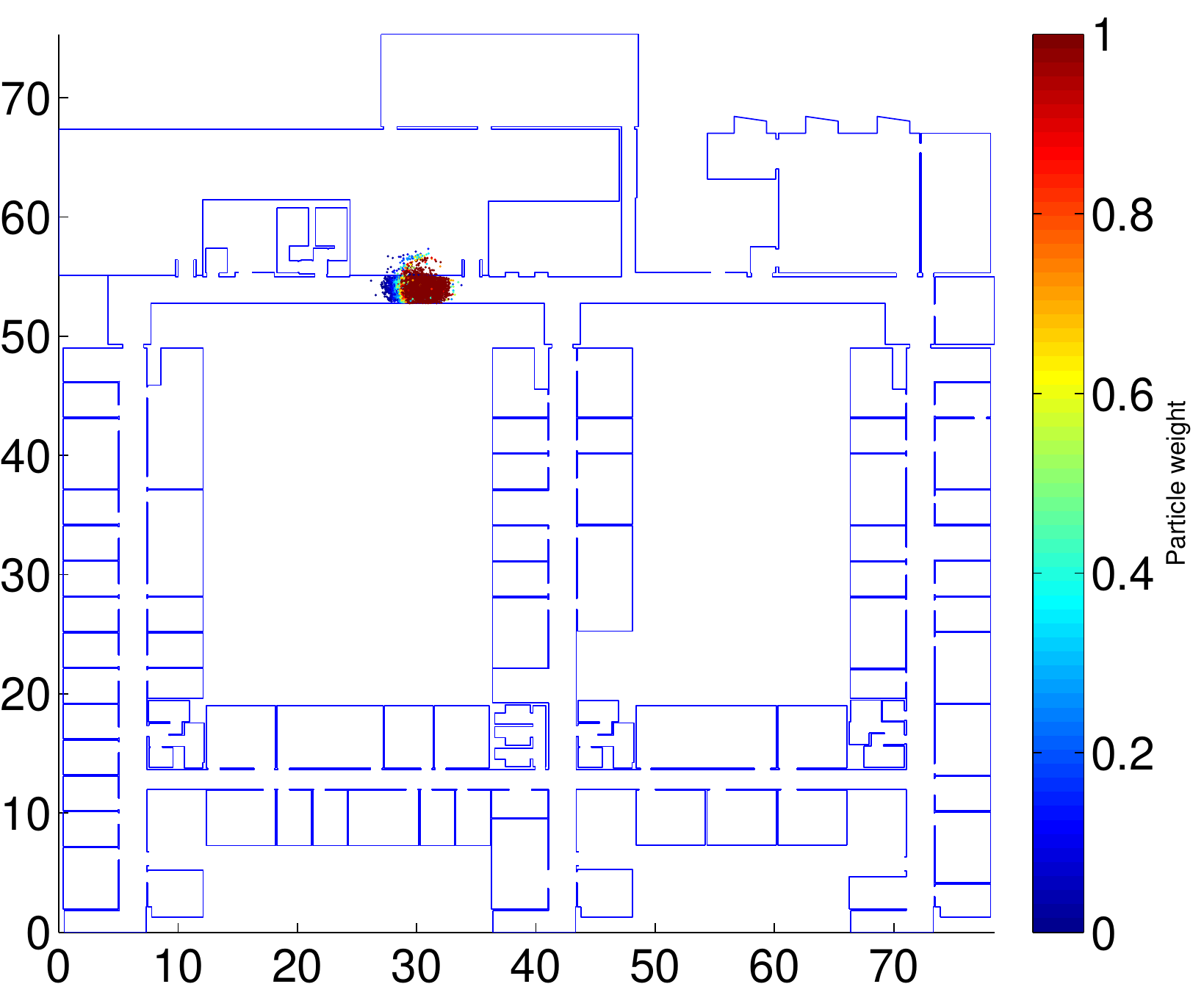}}
		&
		\subfigure[]{\includegraphics[height=4cm]{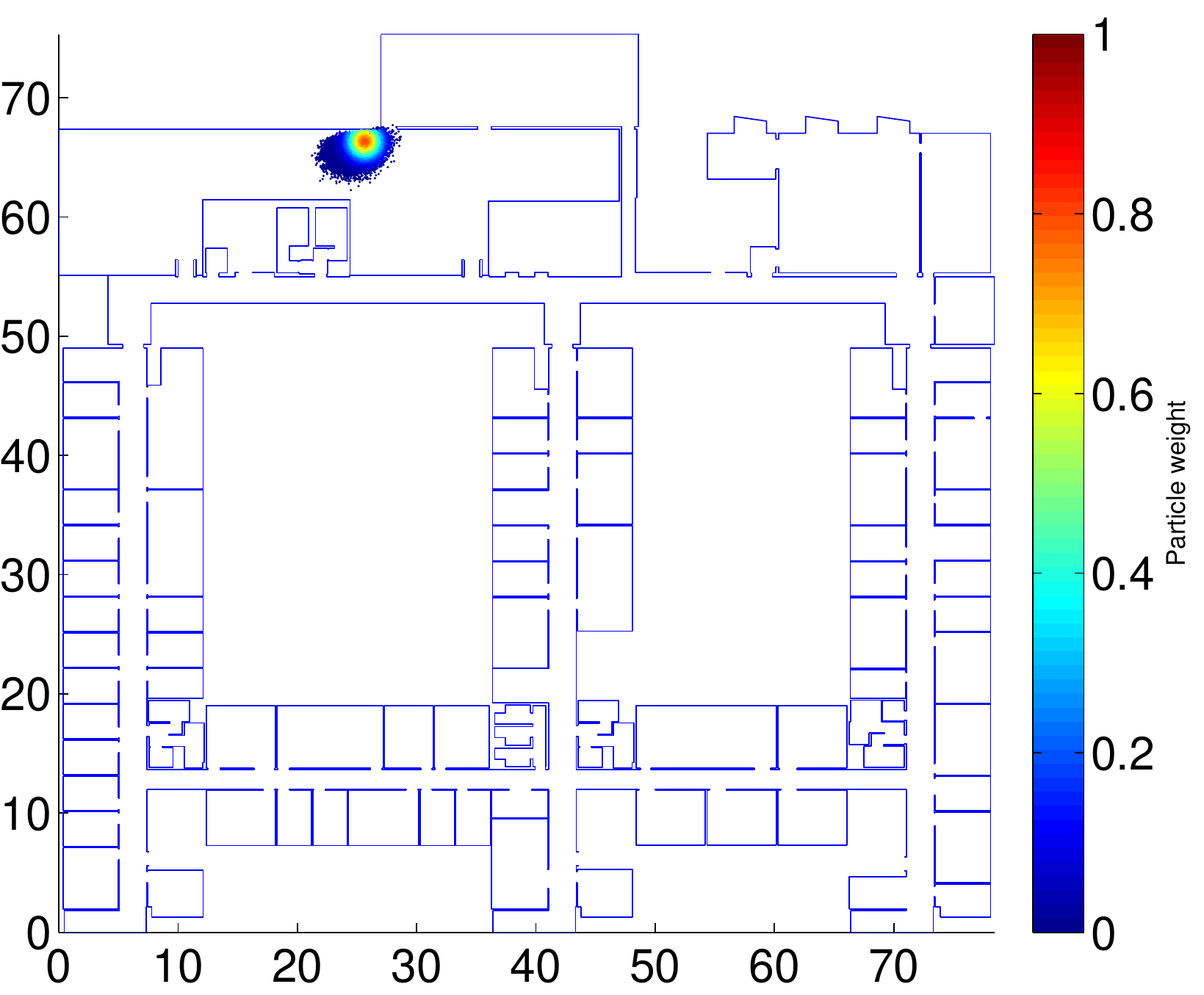}}
		\\
		\hline
	\end{tabular}
	\caption{The results of PFSurvey on the path \textit{Path-1}, \textit{Path-2} and \textit{Path-3}. The magenta parts of the resultant trajectories correspond to the magenta loop closures in Figure~\ref{fig:lc-results}.}
	\label{fig:pfs-results}
\end{figure*}

Figure~\ref{fig:pfs-results} shows some sample outputs from PFSurvey
for the three paths. The top row shows the estimated trajectory in
black, with a magenta highlight corresponding to the magenta loop
closures in Figure~\ref{fig:lc-results}. The bottom row shows the
probability distribution corresponding to the positions marked with a
blue dot in the top row. Note these positions match those used to
generate Figure \ref{fig:pcloud-w1-normal-pdr},
\ref{fig:pcloud-w1-normal-slam}, \ref{fig:pcloud-path-2-normal-slam}
and \ref{fig:pcloud-path-3-normal-slam}, which gave multi-modal
distributions spanning multiple rooms.

\begin{figure}[htbp]
	\begin{center}
		\includegraphics[width=7cm]{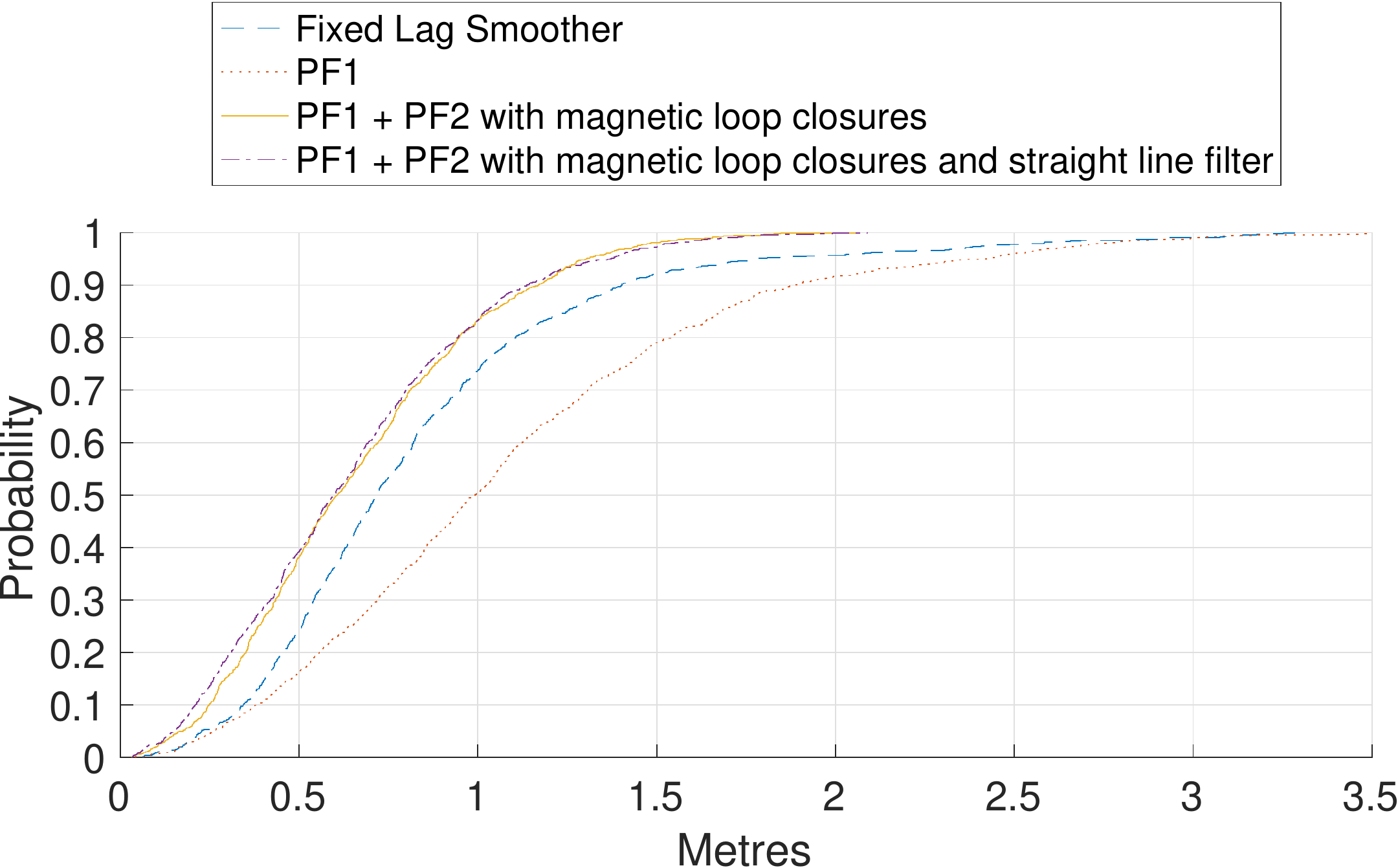} 
		\caption[The error CDFs when applying a FL smoother
		and our proposed system on \textit{Path-1}
		data.]{The error CDFs when applying a FL smoother
			and our proposed system on \textit{Path-1}
			data. }
		\label{fig:accuracy-compare}
	\end{center}
\end{figure}

We first consider \emph{Path-1}. Figure~\ref{fig:pcloud-w1-pfs-pdr}
illustrates that our system has resulted in a uni-modal distribution
where it was previously multi-modal, removing the room
ambiguity. Figure~\ref{fig:accuracy-compare} shows the CDF of the
errors (made possible from the availability of high-accuracy ground
truth for \emph{Path-1}~\cite{Addlesee01}). Results are also shown for a typical run of
the conventional filter plus smoother and the system with different
components disabled. We observe that the PFSurvey result is more
accurate in general: 1.1~m rather than 1.4~m 90\% of the time.  Note
that the CDF is in some sense misleading since it does not capture the
room ambiguity clearly: an error of 1~m is much more significant if it
results in an erroneous room assignment than if it does not.

\begin{figure}[htbp]
	\begin{center}
		\includegraphics[width=5cm]{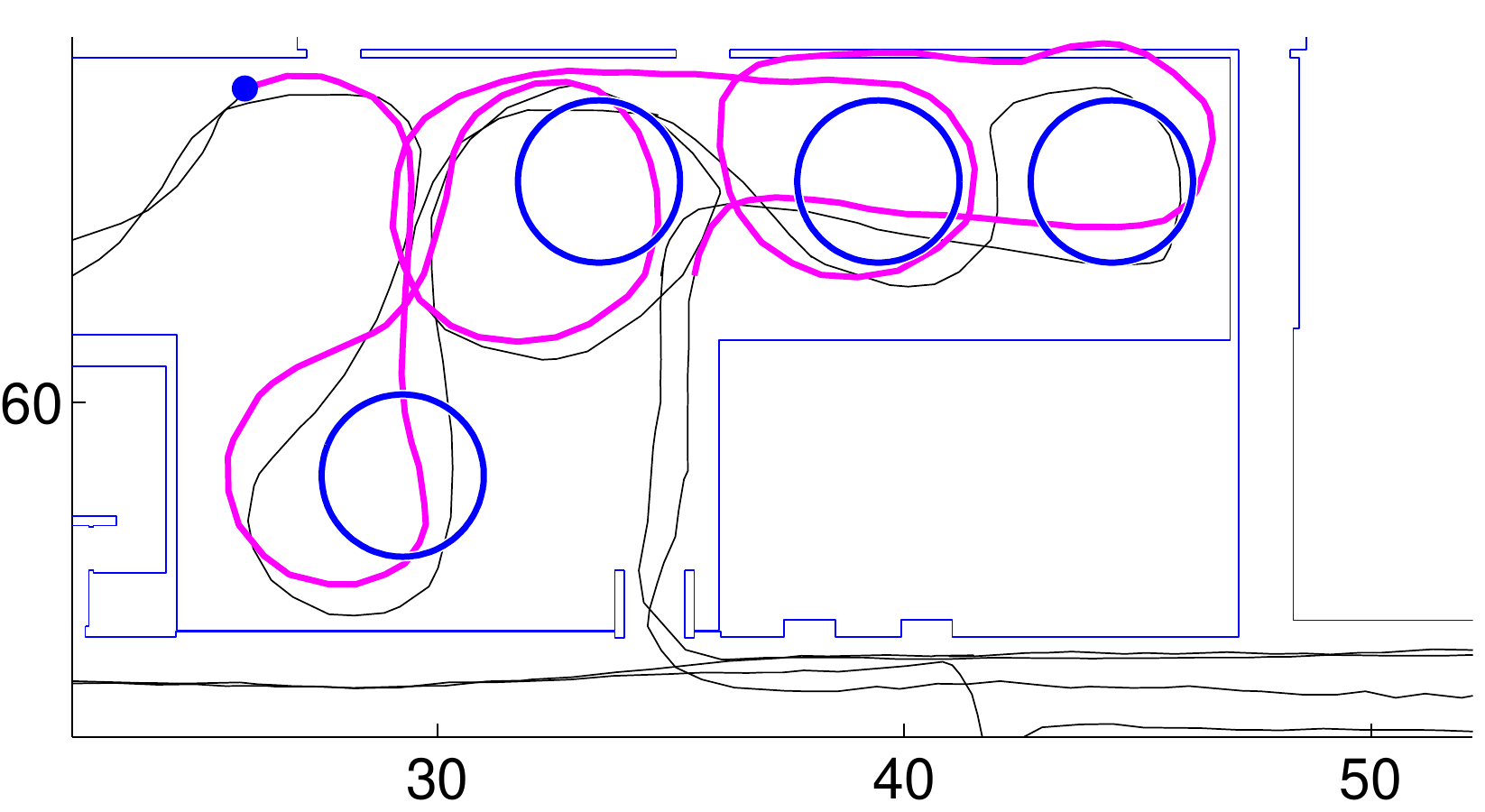} 
		\caption{The estimated trajectory of \textit{Path-3} in the table area. Four large round tables (not in the floorplan) have been added.}
		\label{fig:path-3-table-area}
	\end{center}
\end{figure}

The outputs of PFSurvey for the larger scale \emph{Path-2} and \emph{Path-3} are
also shown in Figure~\ref{fig:pfs-results}. Although these survey
walks did not have accurate ground truth available, the estimated paths
are visually indistinguishable from the paths taken. More specifically, \emph{Path-2} and \emph{Path-3} passed 15 and 16 rooms/spaces in total respectively. The results of conventional method (Figure~\ref{fig:normal-pfs-on-second-floor}) enters five ($33.3\%$) and seven ($43.8\%$) erroneous rooms/spaces respectively; while PFSurvey result (Figure~\ref{fig:pfs-results}) achieves $100\%$ room accuracy.

The magenta-highlighted
loops in Figure~\ref{fig:pfs-results-c} result from walking around
some large tables that were \emph{not} in the floorplan. We
subsequently measured the positions of the tables and we show them
overlaid with the estimated trajectory in
Figure~\ref{fig:path-3-table-area} to illustrate the quality of the
result from PFSurvey. Please note that a digital floor plan usually does not contain the obstacle information like furniture in the indoor environments. A particle filter using this floor plan then has only the wall information to constrain the walking path. In this case, the recovered trajectory may not be well constrained and it may go through furniture. Because many furniture can affect radio propagation, this could cause erroneous signal maps being generated. So here we have demonstrated that by incorporating the loop closure constraints in the filtering process, our system achieves high trajectory recovery accuracy even without the knowledge of furniture information. Because accurate measurement of all the furniture in indoor environments is not an easy task, and furniture sometimes can be added, moved or removed, our system is a more practical and low-cost solution for accurate path signal survey.

\subsection{Signal Map and Positioning Results}

\begin{figure}
	\centering
	\begin{tabular}{ccc}
		\subfigure[PFSurvey output\label{fig:pos-results-a}]{\includegraphics[height=4.5cm]{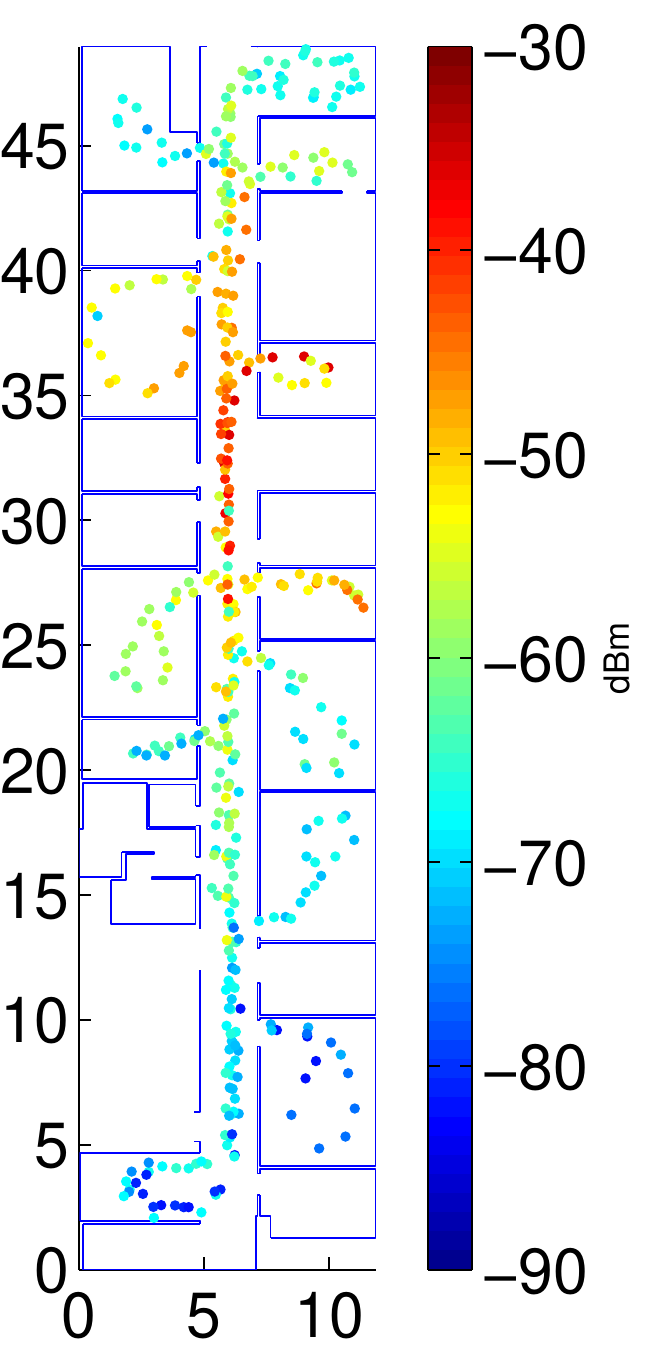}} &
		\subfigure[GP map ($\mu$)\label{fig:pos-results-b}]{\includegraphics[height=4.5cm]{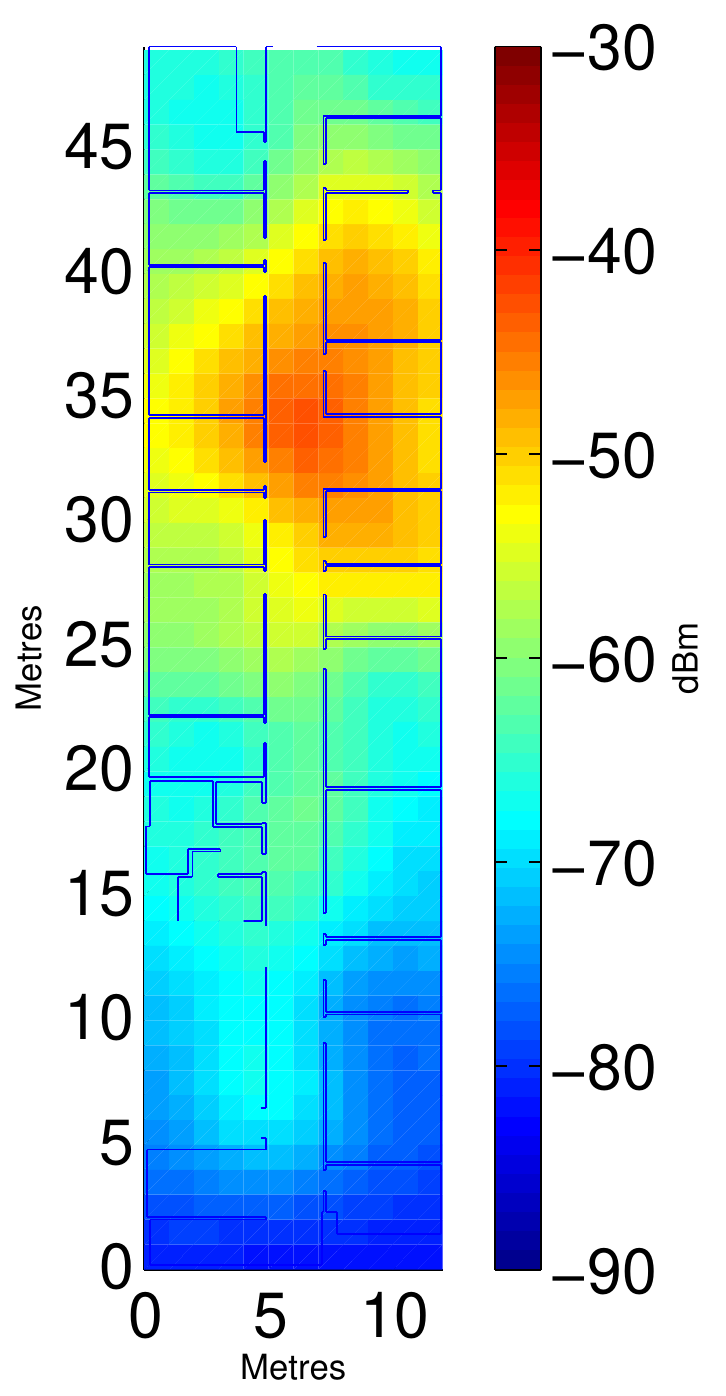}} &
		\subfigure[GP var ($\sigma^2$)\label{fig:pos-results-c}]{\includegraphics[height=4.5cm]{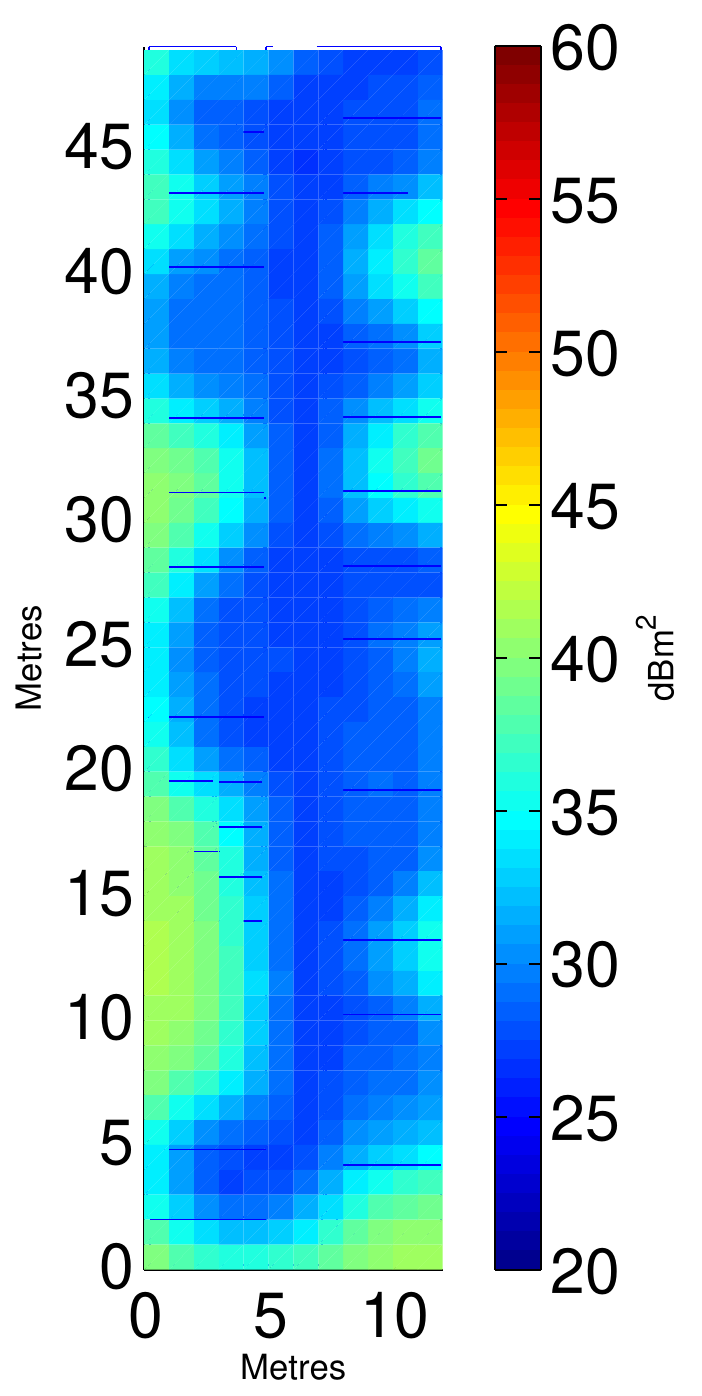}}
		
	\end{tabular}
	\caption{Sample signal map.}
	\label{fig:pos-results}
\end{figure}

\begin{figure}
	\centering
	\begin{tabular}{cc}
		\subfigure[RSS$_{90}$ map]{\includegraphics[height=4.5cm]{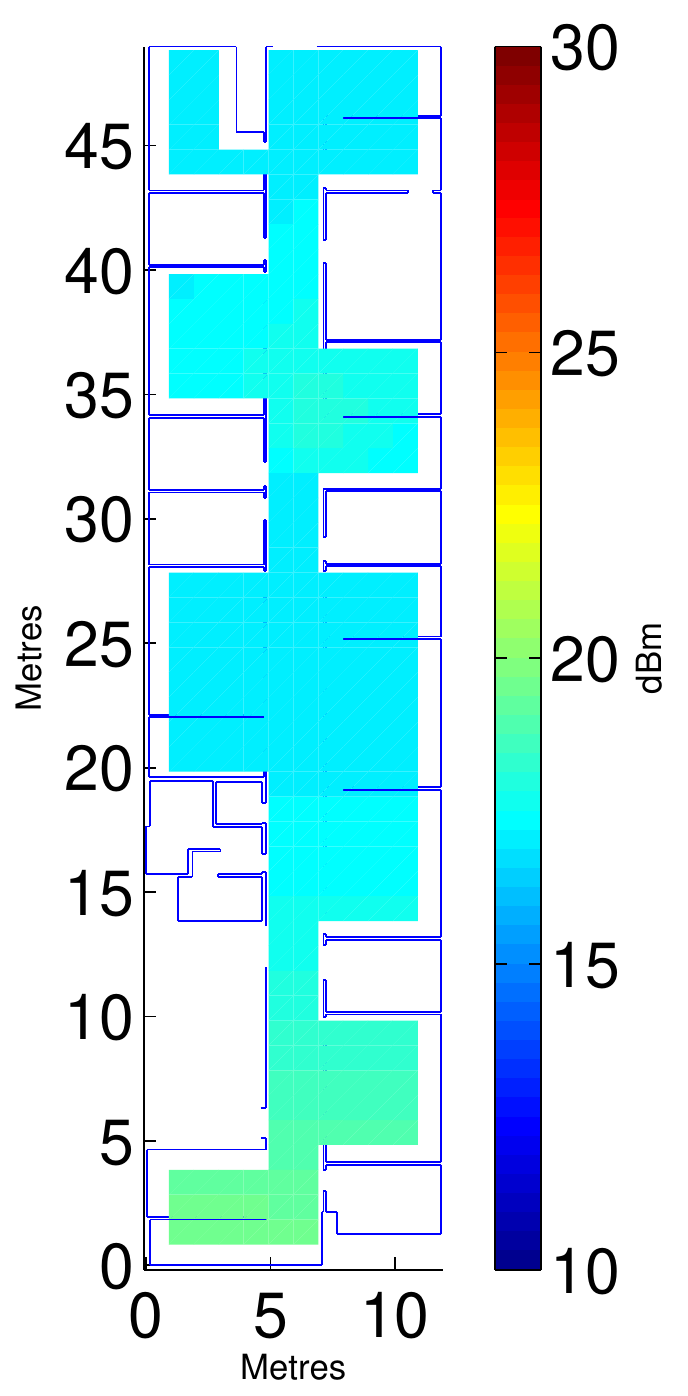}} &

		\subfigure[CDF]{\includegraphics[height=4.5cm]{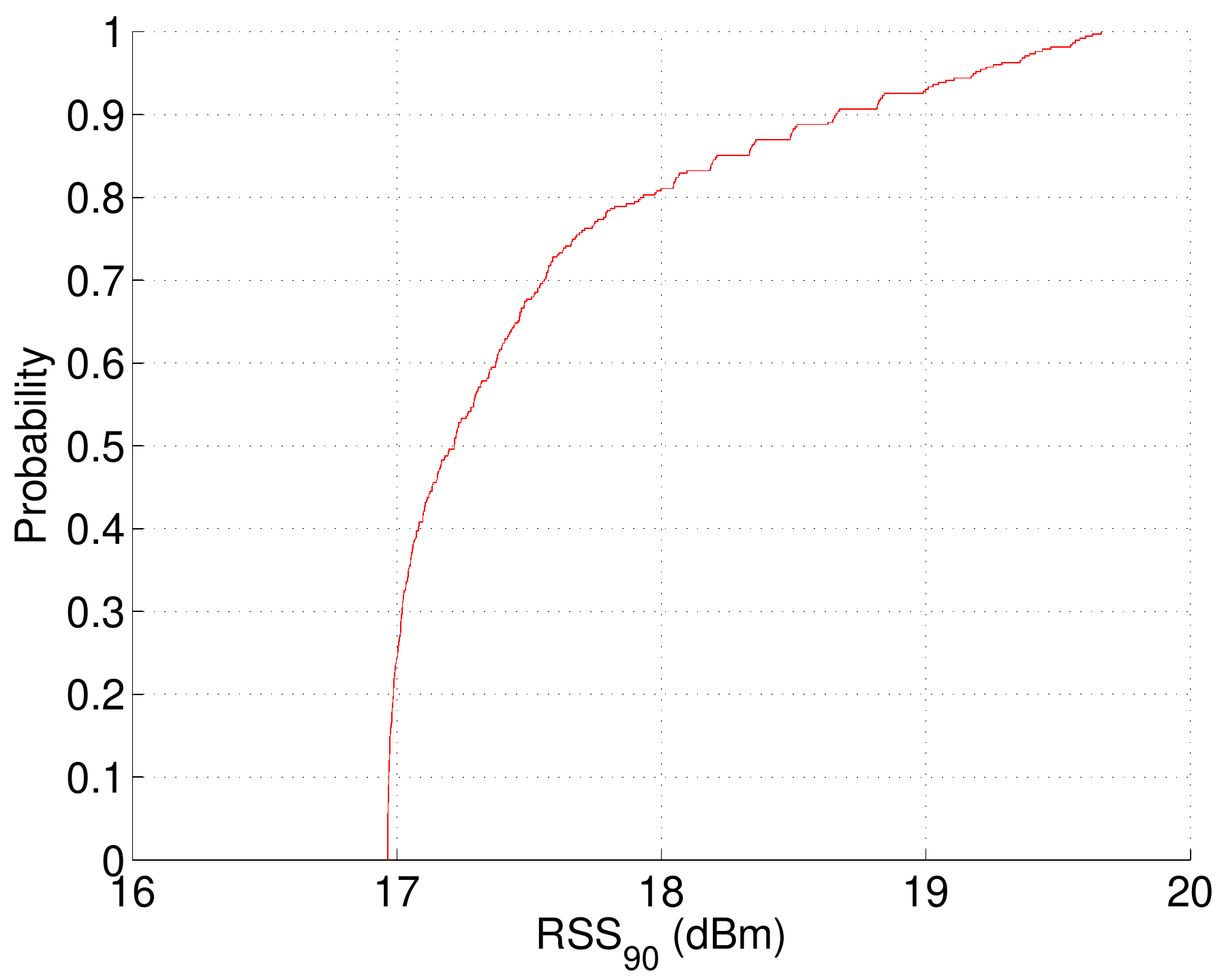}} 
	\end{tabular}
	\caption{RSS$_{90}$ map and CDF}
	\label{fig:gp-pc-map}
\end{figure}

\begin{figure}
  \centering
  \begin{tabular}{c}
    
    \subfigure[Error CDF \label{fig:pos-results-cdf}]{
      \includegraphics[height=4.5cm]{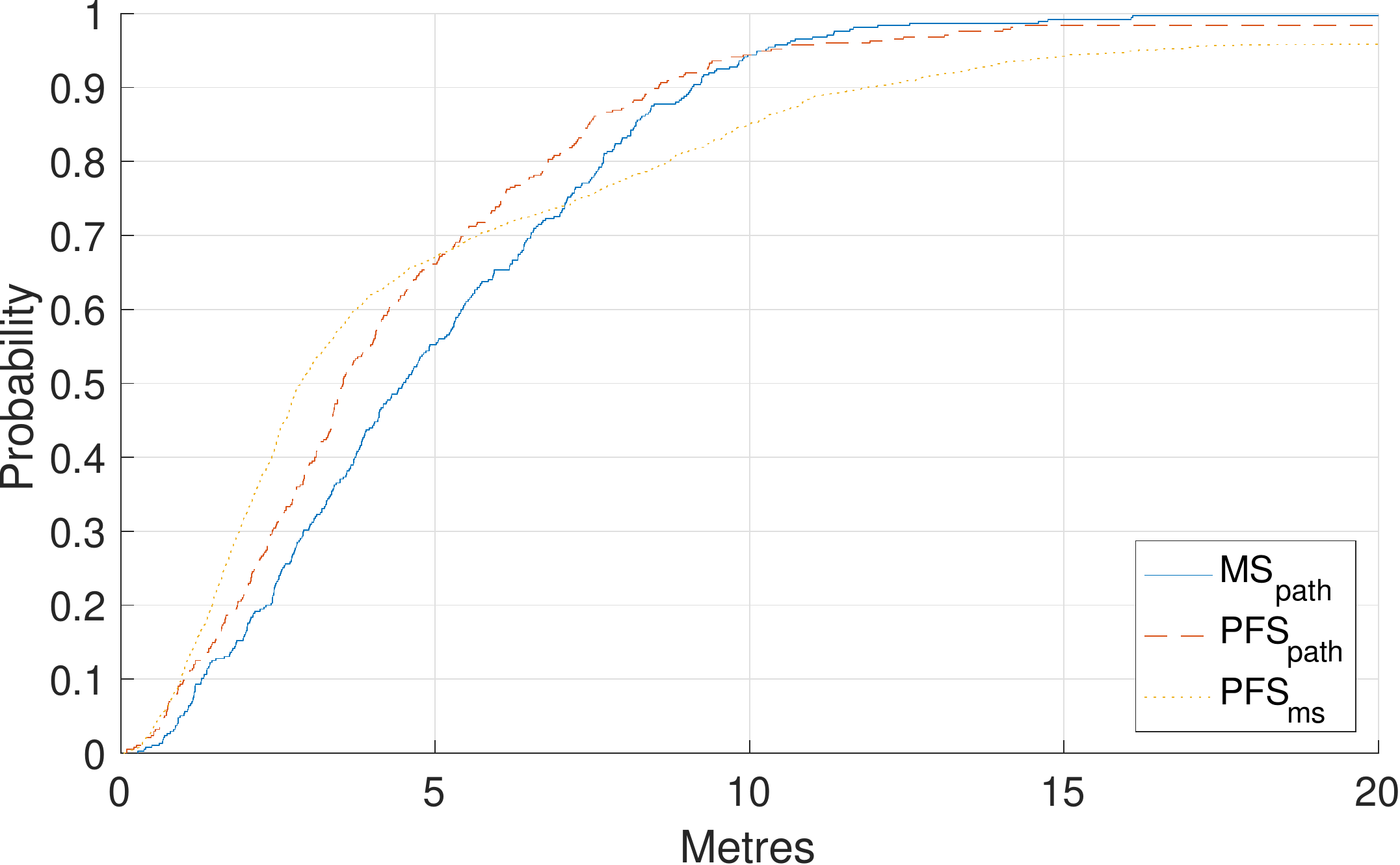}
    }
      \\
    \subfigure[Positioning CDF for PFS$_{ms}$ broken down by distance from survey path\label{fig:pfs-err2dis}] {
	  \includegraphics[height=4.5cm]{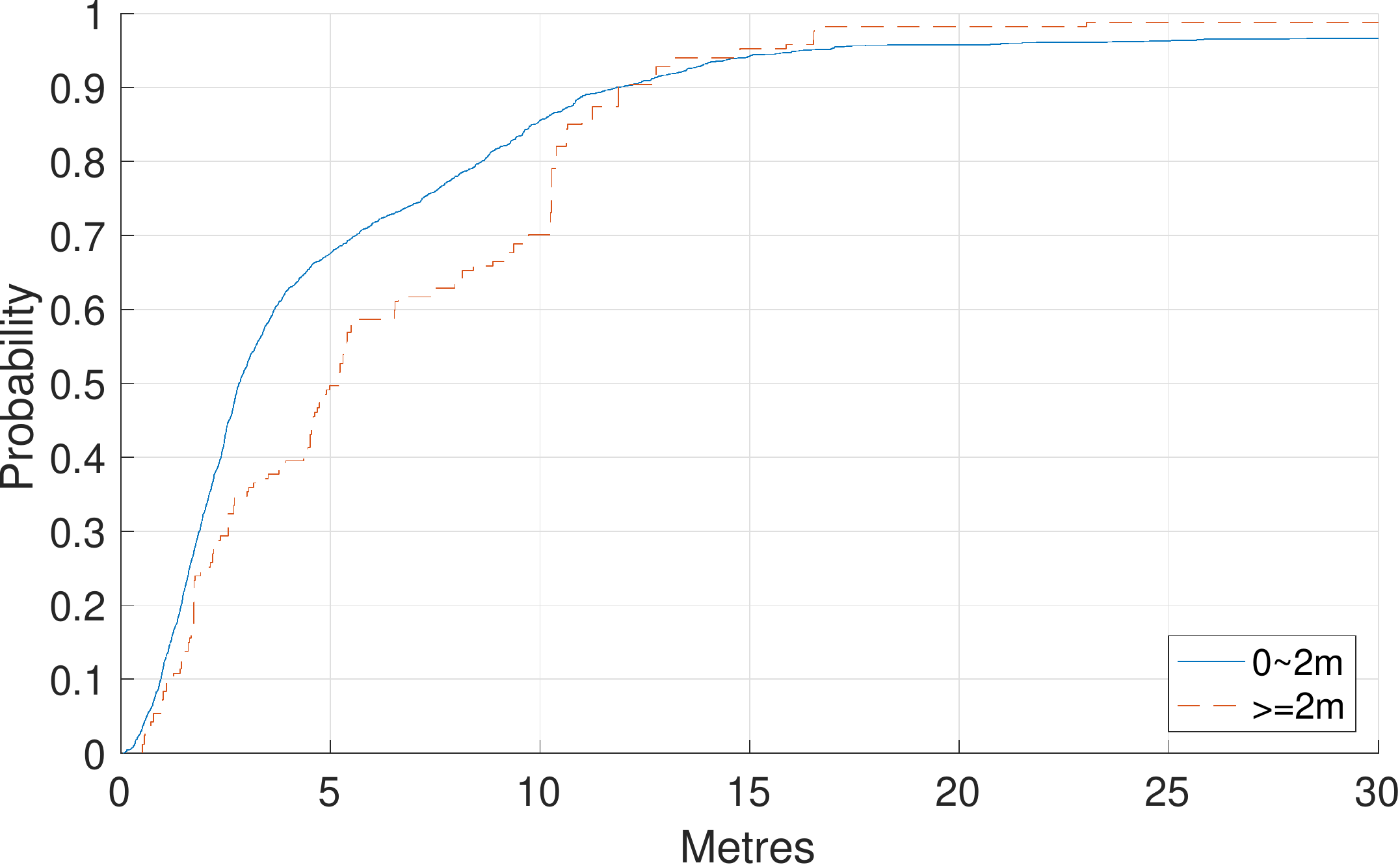}
          } 
        \end{tabular}
	\caption{Positioning Results.}
\end{figure}

The primary focus of this work is the accurate estimation of a user
trajectory from which to generate signal maps using path survey
techniques. For completeness we also consider the subsequent stages of
signal map generation and positioning. We summarise our methodology
here: full details are available in~\cite{gao2016easing,
  gao2015sequence}. We used Gaussian Processes (GP) regression to
generate signal maps from survey points as
per~\cite{ferris2006gaussian}. Figure~\ref{fig:pos-results} shows the
\emph{Path-1} trajectory result. Each point along the trajectory is
coloured to indicate the measured signal strength of an arbitrarily
chosen WiFi access point. The corresponding GP map is illustrated in Figures
\ref{fig:pos-results-b} and \ref{fig:pos-results-c}). We assess its
quality by comparison with the GP map generated from a manual survey
of the same area (the manual survey is shown in Figure
\ref{fig:example-manual-survey}). However, each position on a GP map
is associated with a normal distribution rather than a scalar value so
it is not trivial to compare two GP maps directly. We adopt the
RSS$_{90}$ metric: given two Gaussian distributions, RSS$_{90}$ measures how much they would agree to each other. Please refer to~\cite{gao2016easing} for detailed explanation.

Given two GP maps (the path survey-derived GP map and the
manual survey-derived GP map), we evaluate RSS$_{90}$ at each grid
point position in the area covered by both surveys. We visualise these
RSS$_{90}$s by heatmap and CDF in
Figure~\ref{fig:gp-pc-map}. They show that the agreement between the
path survey map and the manual survey is good.

To evaluate the positioning performance we generated signal maps from
our \emph{Path-1} dataset and used those to perform one-shot
positioning based on two distinct test inputs. The first was an
explicit test walk where ground truth was available through an
external hi-accuracy positioning system~\cite{Addlesee01}. The second
was the manually-surveyed data points. Figure
\ref{fig:pos-results-cdf} shows the CDF of the positioning errors~\footnote{For each test point, the positioning error is defined as the Euclidean distance between its groundtruth position and the position computed by the RSSI positioning algorithm. A point ($n$\%, $m$ metres) on a CDF line means $n$ percent of errors are within the range of $m$ metres.} for
three distinct situations: using the PFSurvey maps to position with
the test walk (PFS$_{path}$); using the PFSurvey maps to position at
the manual survey points (PFS$_{MS}$); and using the manual-survey
maps to position with the test walk (MS$_{path}$).

We observe that the best positioning results were achieved using the
path survey when tested using the explicit test path. This is a direct
consequence of that test path not straying far from the path survey
input. Figure~\ref{fig:pfs-err2dis} emphasises this point: it shows
the positioning errors for the PFS$_{MS}$ broken down by distance of
the test point from the survey path. We see that points close to the
path (within 2~m) gave better accuracy. From this, it might be argued
that the PFS$_{path}$ line in Figure~\ref{fig:pos-results-cdf} is
misleading since points far from the survey path are implicitly
excluded. However, we would expect a dedicated surveyor to walk
through all accessible areas following the most likely paths
(e.g. centre line of the corridor). As such the most common
positioning requests \emph{will} be close to the path survey
trajectory and the PFS$_{path}$ line is then a more realistic
evaluation.

We further note that the maps generated by PFSurvey and those
generated from the manual survey achieve a similar accuracy, despite
the cost of gathering the data being significantly lower for
PFSurvey. In our experience PFSurvey replaces a laborious manual
survey that takes many hours with a simple walk lasting a few minutes.

\section{Conclusions}

In this paper we have described and evaluated PFSurvey, a system
designed to allow a dedicated surveyor to build a signal survey for a
space in a matter of minutes using a commodity smartphone, assuming a
floorplan of the space is available. The system uses a series of
pre-processing steps to generate reliable loop closures between points
on a noisy dead-reckoned trajectory. These are then fused with the
floorplan to provide a robust, accurate trajectory that can be used to
generate maps of any quantity measured during the survey.

We have evaluated PFSurvey in a large building and shown that it can
successfully solve the room ambiguity problem that typically results
from using solely the floorplan to constrain the dead-reckoning
drift. We have demonstrated that the PFSurvey trajectory can be used
to build detailed signal maps that allow positioning accuracy on a par
with conventional, laborious manual surveying.


%

%



\ifCLASSOPTIONcaptionsoff
  \newpage
\fi



%
%
%
\bibliographystyle{abbrv}
\bibliography{refs} 

%

\vspace{-1cm}

\begin{IEEEbiography}[{\includegraphics[width=1in]{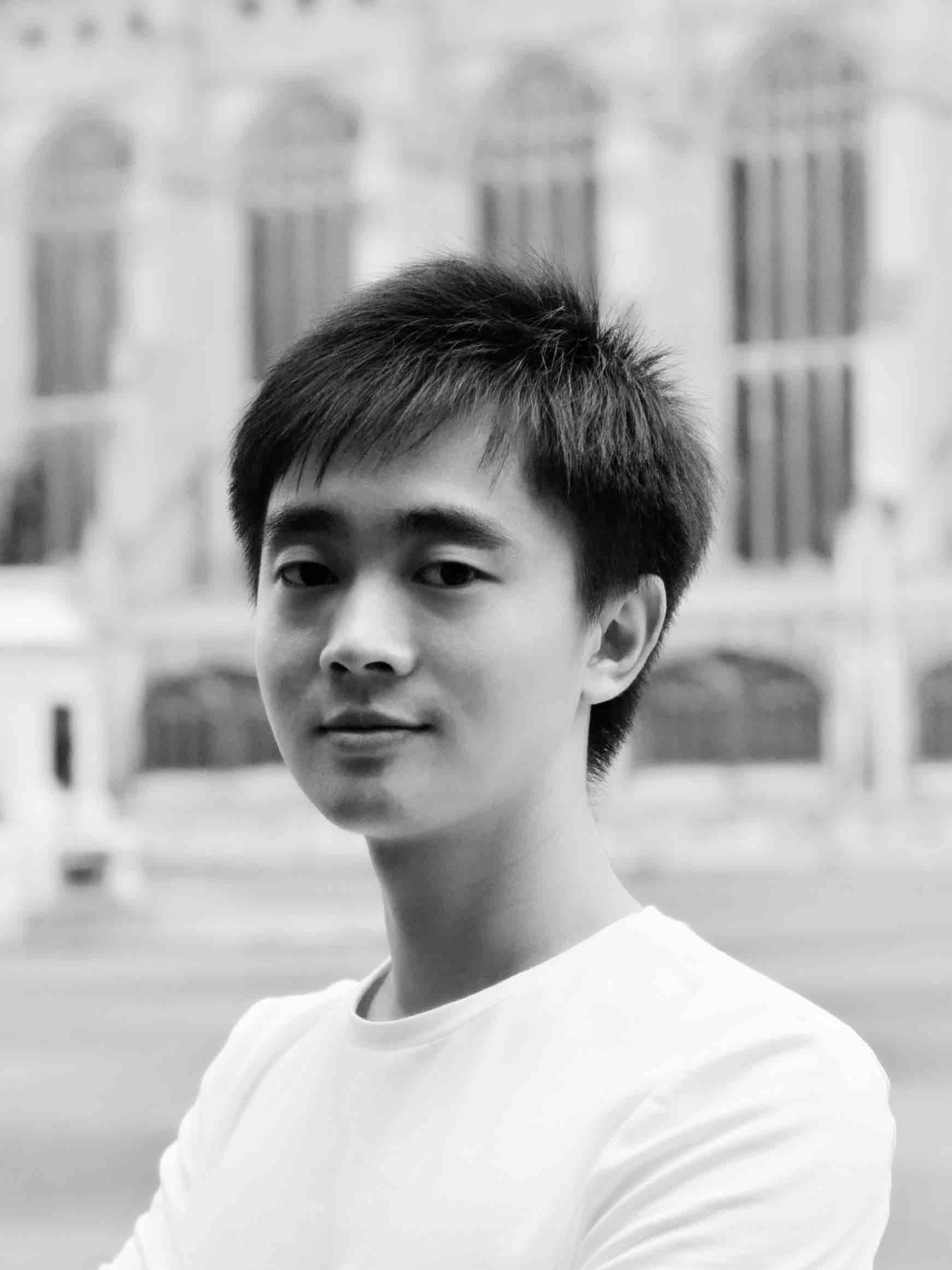}}]{Chao
    Gao} received the BS degree in 2011 from Harbin Institute of
  Technology, China and the M.Phil. and Ph.D. degrees in 2013 and 2017
  from the University of Cambridge, UK.  His research interests
  include indoor positioning, mapping and robotics.
\end{IEEEbiography}

\vfill

\begin{IEEEbiography}[{\includegraphics[width=1in]{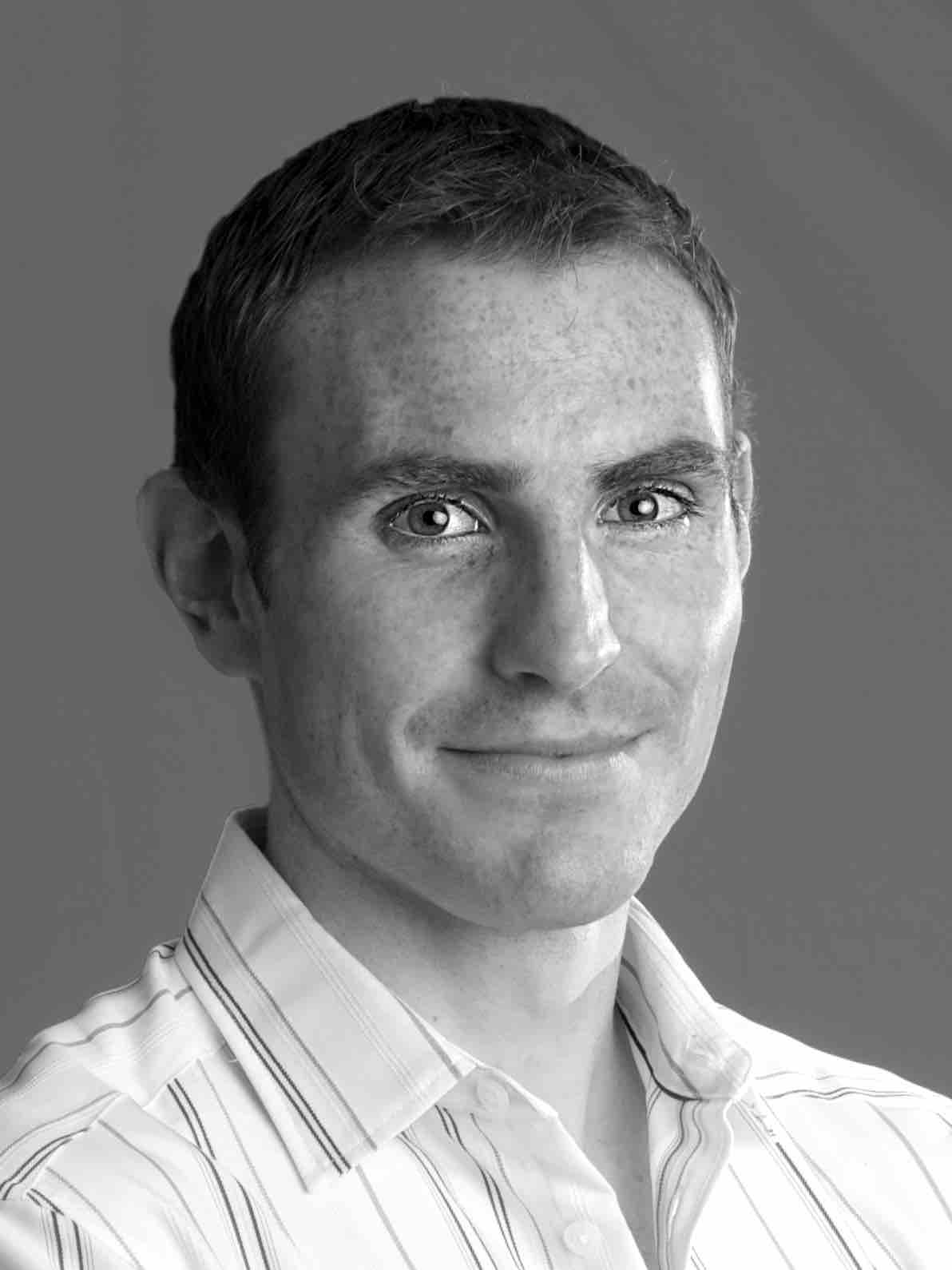}}]{Robert Harle} received the Ph.D. degree from the University of Cambridge, UK, in 2005. He is currently a Senior Lecturer at the Cambridge University Computer Laboratory. His current research interests include mobile, wireless and ubiquitous computing, including mobile health, positioning and the Internet of Things.	
\end{IEEEbiography}

%
%


\vfill


\end{document}